\documentclass[lettersize,journal]{IEEEtran}

\usepackage[utf8]{inputenc}
\usepackage[T1]{fontenc}
\usepackage{graphicx}
\usepackage{amsmath,amssymb}
\usepackage{array}
\usepackage{booktabs}
\usepackage{tabularx}
\usepackage{multirow}
\usepackage{enumitem}
\usepackage{subcaption}
\usepackage{textcomp}
\usepackage{stfloats}
\usepackage{url}
\usepackage{cite}
\usepackage{xcolor}
\usepackage{listings}
\usepackage{tikz}
\usetikzlibrary{shapes.geometric, arrows.meta, positioning, fit, backgrounds, calc}
\usepackage{balance}
\usepackage{hyperref}
\usepackage{eso-pic}
\usepackage{algorithm}
\usepackage{algpseudocode}   %

\hypersetup{
  colorlinks=true,
  linkcolor=blue!50!black,
  citecolor=blue!50!black,
  urlcolor=blue!60!black
}

\usepackage{wrapfig}

\newcommand{\kw}[1]{{\small\texttt{#1}}}
\definecolor{atwl-primary}{RGB}{0,0,153}          %
\definecolor{atwl-control}{RGB}{160,30,30}        %
\definecolor{atwl-type}{RGB}{0,120,120}           %
\definecolor{atwl-intent}{RGB}{120,0,120}         %
\definecolor{atwl-field}{RGB}{154,80,0}           %
\definecolor{atwl-actor}{RGB}{0,100,0}            %
\definecolor{atwl-string}{RGB}{100,100,100}       %
\definecolor{atwl-comment}{RGB}{106,115,125}      %
\definecolor{atwl-operator}{RGB}{180,0,0}         %

\lstdefinelanguage{ATWL}{
  alsoletter={-},
  sensitive=true,
  morekeywords=[1]{artifact, transform},
  morekeywords=[2]{workflow, loop, end, assign, if, then, else, exit},
  morekeywords=[3]{entities, feature, arrangement, visualisation,
    pattern, model, knowledge, specification},
  morekeywords=[4]{define-unit, characterise, contextualise, visualise,
    abstract, build-model, generate-knowledge, assess},
  morekeywords=[5]{template, description, origin, intent, manner,
    input, output, actor, purpose, until, body, features, id,
    layout, form, encoding, context, principle, embedment,
    internal, structure, value, type, representation},
  morekeywords=[6]{human, machine, hybrid},
  morecomment=[l]{\#},
  morestring=[b]",
}

\lstdefinestyle{atwl}{
  language=ATWL,
  basicstyle=\small\ttfamily,
  xleftmargin=1em,
  framexleftmargin=0.5em,
  frame=l,
  framerule=0.6pt,
  breaklines=true,
  columns=flexible,
  keepspaces=true,
  aboveskip=0.8em,
  belowskip=0.8em,
  keywordstyle=[1]\bfseries\color{atwl-primary},   %
  keywordstyle=[2]\bfseries\color{atwl-control},   %
  keywordstyle=[3]\bfseries\color{atwl-type},      %
  keywordstyle=[4]\bfseries\color{atwl-intent},    %
  keywordstyle=[5]\color{atwl-field},              %
  keywordstyle=[6]\bfseries\itshape\color{atwl-actor}, %
  commentstyle=\itshape\color{atwl-comment},
  stringstyle=\color{atwl-string},
  literate={->}{{\textcolor{atwl-operator}{\textrightarrow}}}2
           {:=}{{\textcolor{atwl-operator}{:=}}}2
           {\ }{{\space}}1
}

\tikzset{
    artifact/.style={
        rectangle, draw=black, thick,
        fill=blue!10,
        minimum width=2.5cm, minimum height=1cm,
        text centered, font=\small
    },
    transform/.style={
        ellipse, draw=black, thick,
        fill=green!10,
        minimum width=2cm, minimum height=0.8cm,
        text centered, font=\small
    },
    human/.style={
        artifact, fill=orange!10, line width=1.5pt
    },
    loop/.style={
        draw, dashed, inner sep=0.5cm,
        label={[anchor=north west]north west:\textbf{#1}}
    },
    arrow/.style={-Stealth, thick}
}

\tikzset{
    atwl_icon/.style={scale=0.5}
}

\definecolor{atwlEntities}      {HTML}{4682B4}  %
\definecolor{atwlFeature}       {HTML}{5B9BD5}  %
\definecolor{atwlArrangement}   {HTML}{D4A520}  %
\definecolor{atwlVisualisation} {HTML}{2E8B57}  %
\definecolor{atwlPattern}       {HTML}{E07050}  %
\definecolor{atwlModel}         {HTML}{C0392B}  %
\definecolor{atwlKnowledge}     {HTML}{7B68AE}  %
\definecolor{atwlSpecification} {HTML}{9B8EC4}  %

\tikzset{atwl_icon/.style={line width=0.6pt}}

\newcommand{\iconEntities}{%
  \scalebox{0.8}{%
  \begin{tikzpicture}[atwl_icon, baseline=-0.5ex]
    \foreach \x/\y in {0.05/0.02,  0.32/0.24,  0.65/0.10,
                       0.18/-0.20, 0.52/-0.28}{
      \fill[atwlEntities] (\x,\y) circle (2.2pt);
    }
  \end{tikzpicture}}%
}

\newcommand{\iconFeature}{%
  \scalebox{0.8}{%
  \begin{tikzpicture}[atwl_icon, baseline=-0.5ex]
    \foreach \x/\y in {0.05/0.07,  0.32/0.29,
                       0.65/0.15,  0.18/-0.15}{
      \fill[atwlFeature] (\x,\y) circle (2.2pt);
    }
    \draw[atwlFeature, thick, fill=atwlFeature!15,
          rounded corners=1pt]
      (0.54,-0.38) rectangle (0.92,-0.06);
    \node[atwlFeature, font=\tiny\bfseries]
      at (0.73,-0.22) {f};
  \end{tikzpicture}}%
}

\newcommand{\iconArrangement}{%
  \scalebox{0.8}{%
  \begin{tikzpicture}[atwl_icon, baseline=-0.5ex]
    \foreach \x/\y in {0/0.25, 0.35/0.25, 0.70/0.25,
                       0/0,    0.35/0,    0.70/0}{
      \fill[atwlArrangement] (\x,\y) circle (2.2pt);
    }
  \end{tikzpicture}}%
}

\newcommand{\iconVisualisation}{%
  \scalebox{0.8}{%
  \begin{tikzpicture}[atwl_icon, baseline=-0.5ex]
    \draw[atwlVisualisation, thick]
      (0,0.58) -- (0,0) -- (0.88,0);
    \draw[atwlVisualisation, thick]
      (0.08,0.12) -- (0.25,0.42) -- (0.42,0.22)
      -- (0.60,0.50) -- (0.80,0.34);
  \end{tikzpicture}}%
}

\newcommand{\iconPattern}{%
  \scalebox{0.8}{%
  \begin{tikzpicture}[atwl_icon, baseline=-0.5ex]
    \draw[atwlPattern, thick]
      (0,0.15)  sin  (0.2,0.35)
                cos  (0.4,0.15)
                sin  (0.6,-0.05)
                cos  (0.8,0.15);
  \end{tikzpicture}}%
}

\newcommand{\iconModel}{%
  \scalebox{0.8}{%
  \begin{tikzpicture}[atwl_icon, baseline=-0.5ex]
    \draw[atwlModel, thick, ->]
      (0,0.05) -- (0.85,0.50);
    \draw[atwlModel, thick]
      (0,0.45) .. controls (0.35,-0.15) .. (0.85,0.25);
  \end{tikzpicture}}%
}

\newcommand{\iconKnowledge}{%
  \scalebox{0.8}{%
  \begin{tikzpicture}[atwl_icon, baseline=1.0ex]
    \draw[atwlKnowledge, thick, fill=atwlKnowledge!15]
      (0.40,0.42) circle (0.19);
    \draw[atwlKnowledge, thick]
      (0.28,0.24) -- (0.28,0.10)
      -- (0.52,0.10) -- (0.52,0.24);
    \begin{scope}[shift={(0.40,0.42)}]
      \foreach \a in {45,90,135}{
        \draw[atwlKnowledge, thick] (\a:0.22) -- (\a:0.32);
      }
    \end{scope}
  \end{tikzpicture}}%
}

\newcommand{\iconSpecification}{%
  \scalebox{0.8}{%
  \begin{tikzpicture}[atwl_icon, baseline=-0.5ex]
    \foreach \y in {0, 0.20, 0.40}{
      \draw[atwlSpecification] (0,\y) -- (0.8,\y);
    }
    \fill[atwlSpecification] (0.50, 0.40) circle (2.5pt);
    \fill[atwlSpecification] (0.65, 0.20) circle (2.5pt);
    \fill[atwlSpecification] (0.25, 0.00) circle (2.5pt);
  \end{tikzpicture}}%
}

\definecolor{atwlTransform}{HTML}{2C3E50}   %

\newcommand{\iconDefineUnit}{%
  \scalebox{0.6}{%
  \begin{tikzpicture}[atwl_icon, baseline=-0.5ex]
    \draw[atwlTransform, thick]
      (0.15,0.32) -- (0.55,0.08);
    \draw[atwlTransform, thick]
      (0.15,-0.22) -- (0.55,0.02);
    \fill[atwlTransform] (0.48,0.05) circle (1.4pt);
    \draw[atwlTransform, thick]
      (0.08,0.36) circle (0.1);
    \draw[atwlTransform, thick]
      (0.08,-0.26) circle (0.1);
    \draw[atwlTransform, thick]
      (0.55,0.08) -- (0.88,0.22);
    \draw[atwlTransform, thick]
      (0.55,0.02) -- (0.88,-0.12);
  \end{tikzpicture}}%
}

\newcommand{\iconCharacterise}{%
  \scalebox{0.7}{%
  \begin{tikzpicture}[atwl_icon, baseline=-0.5ex]
    \draw[atwlTransform, thick]
      (0,0.25) -- (0.85,0.25);
    \draw[atwlTransform, thick]
      (0.05,0.25) -- (0.05,-0.15);
    \draw[atwlTransform, thick]
      (0.05,-0.15) -- (0.15,-0.15);
    \draw[atwlTransform, thick]
      (0.55,0.25) -- (0.55,-0.15);
    \draw[atwlTransform, thick]
      (0.55,-0.15) -- (0.45,-0.15);
    \draw[atwlTransform, thin]
      (0.85,0.28) -- (0.85,0.22);
    \draw[atwlTransform, thin, <->]
      (0.15,-0.05) -- (0.45,-0.05);
  \end{tikzpicture}}%
}

\newcommand{\iconContextualise}{%
  \scalebox{0.7}{%
  \begin{tikzpicture}[atwl_icon, baseline=-0.5ex]
    \draw[atwlTransform, thick]
      (0,0.55) -- (0,0) -- (0.75,0);
    \draw[atwlTransform, thin, opacity=0.4]
      (0.25,0) -- (0.25,0.50);
    \draw[atwlTransform, thin, opacity=0.4]
      (0.50,0) -- (0.50,0.50);
    \draw[atwlTransform, thin, opacity=0.4]
      (0,0.25) -- (0.70,0.25);
    \fill[atwlTransform] (0.50,0.38) circle (2.2pt);
    \draw[atwlTransform, thin]
      (0.38,0.38) -- (0.62,0.38);
    \draw[atwlTransform, thin]
      (0.50,0.26) -- (0.50,0.50);
  \end{tikzpicture}}%
}

\newcommand{\iconVisualise}{%
  \scalebox{0.7}{%
  \begin{tikzpicture}[atwl_icon, baseline=-0.5ex]
    \draw[atwlTransform, thick]
      (0,0.15) .. controls (0.25,0.50) and (0.60,0.50)
      .. (0.85,0.15);
    \draw[atwlTransform, thick]
      (0,0.15) .. controls (0.25,-0.15) and (0.60,-0.15)
      .. (0.85,0.15);
    \draw[atwlTransform, thick]
      (0.425,0.15) circle (0.13);
    \fill[atwlTransform]
      (0.425,0.15) circle (0.06);
  \end{tikzpicture}}%
}

\newcommand{\iconAbstract}{%
  \scalebox{0.8}{%
  \begin{tikzpicture}[atwl_icon, baseline=-0.5ex]
    \fill[atwlTransform] (0.10,-0.10) circle (1.8pt);
    \fill[atwlTransform] (0.35,-0.10) circle (1.8pt);
    \fill[atwlTransform] (0.60,-0.10) circle (1.8pt);
    \draw[atwlTransform, thick, ->]
      (0.15,0.00) -- (0.32,0.28);
    \draw[atwlTransform, thick, ->]
      (0.35,0.02) -- (0.35,0.28);
    \draw[atwlTransform, thick, ->]
      (0.55,0.00) -- (0.38,0.28);
    \fill[atwlTransform] (0.35,0.38) circle (2.2pt);
  \end{tikzpicture}}%
}

\newcommand{\iconBuildModel}{%
  \scalebox{0.8}{%
  \begin{tikzpicture}[atwl_icon, baseline=-0.5ex]
    \begin{scope}[shift={(0.40,0.25)}]
      \foreach \a in {0,60,120,180,240,300}{
        \draw[atwlTransform, thick, fill=atwlTransform!12,
              rotate around={\a:(0,0)}]
          (-0.055,0.15) rectangle (0.055,0.27);
      }
      \draw[atwlTransform, thick, fill=atwlTransform!12]
        (0,0) circle (0.16);
      \fill[white] (0,0) circle (0.05);
      \draw[atwlTransform, thick] (0,0) circle (0.05);
    \end{scope}
  \end{tikzpicture}}%
}

\newcommand{\iconGenerateKnowledge}{%
  \scalebox{0.8}{%
  \begin{tikzpicture}[atwl_icon, baseline=-0.5ex]
    \draw[atwlTransform, thick, fill=atwlTransform!8,
          rounded corners=2.5pt]
      (0,0.15) rectangle (0.80,0.58);
    \fill[atwlTransform!8]
      (0.12,0.15) -- (0.08,-0.02) -- (0.28,0.15) -- cycle;
    \draw[atwlTransform, thick]
      (0.12,0.15) -- (0.08,-0.02) -- (0.28,0.15);
    \begin{scope}[shift={(0.40,0.36)}]
      \draw[atwlTransform, thick]
        (0,-0.12) -- (0,0.12);
      \draw[atwlTransform, thick]
        (-0.12,0) -- (0.12,0);
      \draw[atwlTransform, thick]
        (-0.07,-0.07) -- (0.07,0.07);
      \draw[atwlTransform, thick]
        (-0.07,0.07) -- (0.07,-0.07);
      \fill[atwlTransform] (0,0) circle (1.2pt);
    \end{scope}
  \end{tikzpicture}}%
}

\newcommand{\iconAssess}{%
  \scalebox{0.8}{%
  \begin{tikzpicture}[atwl_icon, baseline=-0.5ex]
    \draw[atwlTransform, thick]
      (0.40,0.22) circle (0.28);
    \draw[atwlTransform, very thick, line cap=round, line join=round]
      (0.22,0.22) -- (0.35,0.08) -- (0.58,0.40);
  \end{tikzpicture}}%
}

\tikzset{
    artifact node/.style={
        artifact,
        inner sep=4pt
    },
    transform node/.style={
        transform,
        inner sep=3pt
    }
}

\renewcommand{\iconDefineUnit}{%
  \scalebox{0.8}{%
  \begin{tikzpicture}[atwl_icon, baseline=-0.5ex]
    \draw[atwlTransform, thin] (0.10,0.30) circle (1.5pt);
    \draw[atwlTransform, thin] (0.10,-0.28) circle (1.5pt);
    \draw[atwlTransform, thin] (0.94,0.24) circle (1.5pt);
    \draw[atwlTransform, thick, dash pattern=on 2pt off 1.5pt, rounded corners=3pt]
      (0.26,-0.32) rectangle (0.80,0.38);
    \fill[atwlTransform] (0.40,0.22) circle (1.9pt);
    \fill[atwlTransform] (0.62,0.12) circle (1.9pt);
    \fill[atwlTransform] (0.44,-0.10) circle (1.9pt);
    \fill[atwlTransform] (0.64,-0.18) circle (1.9pt);
  \end{tikzpicture}}%
}

\renewcommand{\iconAbstract}{%
  \scalebox{0.8}{%
  \begin{tikzpicture}[atwl_icon, baseline=-0.5ex]
    \fill[atwlTransform] (0.06,-0.30) circle (1.6pt);
    \fill[atwlTransform] (0.28,-0.22) circle (1.6pt);
    \fill[atwlTransform] (0.50,-0.32) circle (1.6pt);
    \fill[atwlTransform] (0.72,-0.20) circle (1.6pt);
    \fill[atwlTransform] (0.94,-0.30) circle (1.6pt);
    \draw[atwlTransform, thick, -{Stealth[length=3pt]}] (0.50,-0.12) -- (0.50,0.08);
    \draw[atwlTransform, thick]
      (0.04,0.26) .. controls (0.20,0.46) and (0.34,0.46) .. (0.50,0.26)
                  .. controls (0.66,0.06) and (0.80,0.06) .. (0.96,0.26);
  \end{tikzpicture}}%
}

\title{ATWL: A Formal Language for Representing, Comparing, and
Reusing Visual Analytics Workflows}

\author{Natalia~Andrienko,
        Gennady~Andrienko,
        J\"urgen~Bernard,
        Kuno Kurzhals,
        and~Michael~Sedlmair%
\IEEEcompsocitemizethanks{%
\IEEEcompsocthanksitem N.\ Andrienko and G.\ Andrienko are with
Fraunhofer Institute IAIS, the Lamarr Institute for Machine Learning
and Artificial Intelligence, Sankt Augustin, Germany, and City
St~George's, University of London, UK.\protect\\
E-mail: \{natalia, gennady\}.andrienko@iais.fraunhofer.de
\IEEEcompsocthanksitem J.\ Bernard is with the University of Zurich,
Switzerland.\protect\\
E-mail: juergen.bernard@uzh.ch
\IEEEcompsocthanksitem K.\ Kurzhals is with the University of
Stuttgart, Germany.\protect\\
E-mail: Kuno.Kurzhals@visus.uni-stuttgart.de%
\IEEEcompsocthanksitem M.\ Sedlmair is with the University of
Stuttgart, Germany.\protect\\
E-mail: michael.sedlmair@visus.uni-stuttgart.de}%
\thanks{Manuscript received xx xxx.\ 202x; accepted xx xxx.\ 202x.
Digital Object Identifier: xx.xxxx/TVCG.202x.xxxxxxx}}

\IEEEtitleabstractindextext{%
\begin{abstract}
Visual analytics (VA) workflows combine data transformation, feature computation, visual representation, and human interpretation, but they are usually described in prose. This hinders comparison of workflows, reuse of proven strategies, and training of novices. We present Artifact--Transform Workflow Language (ATWL), a declarative language for representing the structure and intent of VA workflows. A workflow consists of typed artifacts and of transforms, each with an analytical intent and a human, machine, or hybrid actor. We built a library of 54 workflows extracted from published VA papers with the help of LLMs under human supervision. An exploratory computational comparison of the library workflows shows regularities, such as a regular division of labour between humans and machines and cycles of refinement. We ran experiments in which LLMs designed workflows for new tasks. They show the effects of using the library, the ATWL representation, and both together, and they reveal a limitation in expressing branching that requires an extension of the ATWL syntax. The library is small and partly drawn from our own work, so the findings are preliminary.
\end{abstract}

}

\begin{document}

\maketitle
\IEEEdisplaynontitleabstractindextext
\IEEEpeerreviewmaketitle

\thispagestyle{empty}
\AddToShipoutPictureFG*{%
  \put(\LenToUnit{0.5\paperwidth},\LenToUnit{\paperheight-0.55in}){%
    \makebox(0,0)[c]{\footnotesize\rmfamily Submitted to IEEE Transactions on Visualization and Computer Graphics, 25 May 2026; revised 16 October 2026}}}

\setlength{\belowcaptionskip}{-5mm}

\section{Introduction} \label{sec:introduction}

Visual analytics (VA) integrates computational methods and human cognition to derive insights from complex data. Joint activities of computers and humans form analytical workflows, which include data transformation, feature computation, visual representation, interpretation, and iterative refinement. These workflows are communicated mainly in natural language, in research papers and technical reports such as design studies~\cite{sedlmair2012design}.

This reliance on prose has three consequences. First, workflows are difficult to compare across data types and application domains. For example, it is hard to tell whether a trajectory clustering approach and a topic modelling pipeline share an underlying structure. Second, implementation details (methods, parameters, encodings, etc.) are conflated with analytical intent, which limits reproducibility. Third, novice practitioners have no collection of proven analytical strategies to consult or adapt. More generally, without a shared formal language, workflows cannot be decomposed, compared, and composed on principled grounds, and the study of human--computer analytical processes remains narrative.

We present the \emph{Artifact--Transform Workflow Language} (ATWL), a declarative language for representing the structure and intent of VA workflows. A workflow in ATWL consists of typed artifacts, such as entities, features, and arrangements, and of transforms that produce artifacts from artifacts. A transform has an analytical intent (e.g., \kw{define-unit}, \kw{characterise}, \kw{contextualise}, \kw{abstract}) and an actor, which is a human, a machine, or both. Loops and conditionals express iteration and conditional steps. The language was developed by progressively formalising existing VA workflows from diverse application areas.

ATWL describes a workflow as it is reported or designed, not as an execution trace. ATWL is meant to be produced with the help of LLMs (rather than written by hand) and checked by the syntax checker and by the authors of the workflow. We consider three uses: (1)~documentation of workflows in publications, (2)~comparative study and teaching with a library of formalised workflows, and (3)~support of workflow design by LLMs. The typed and linked artifacts and transforms of ATWL additionally allow automatic checking of a workflow and computational comparison of workflows, which we demonstrate in this paper. 

ATWL is specific to VA in three respects. It includes visualisation as one of its artifact types. It explicitly represents the division of labour between humans and machines through the actors of the transforms. Loops and specification artifacts represent how analysts redefine the analysis after inspecting results. The other constructs are not tied to visual representations, but we have not tested whether ATWL suits other analytical workflows, such as data science pipelines.

To test the language, we built an open library of 54 workflows extracted from published VA papers. LLMs extracted the workflows under human supervision, and the ATWL syntax checker verified their structure. The library is not a representative sample, and 26 of its 54 papers have a co-author of this article among their authors, so our analyses are exploratory. We invite the VA research community to take part in its further development. To investigate what ATWL makes possible, we analysed the workflows computationally and tested how the library and the representation affect workflow designs produced by LLMs for two tasks. We also received valuable feedback from the authors of ten library papers, none of them authors of this article, who assessed the representations of their own workflows.

The contributions of this work are as follows. First, \textbf{a formal language for representing VA workflows (conceptual contribution).} ATWL is, to our knowledge, the first formal representation of VA workflows that combines independence of the application domain, a focus on the analysis process, a machine-readable syntax, and a restriction to observable, externalised products. Its ontology types artifacts by analytical role, classifies transforms by intent and not by method, and treats externalised human knowledge and specifications as artifacts.

Second, \textbf{an open library of 54 formalised workflows (resource contribution)} \url{https://geoanalytics.net/VAworkflows/library}. The workflows come from papers published between 1999 and 2026. Each has a summary, an ATWL representation, and graphical and modular views. Apart from the library, an interactive viewer, a syntax checker, and the LLM sessions used for extraction are publicly available.

Third, \textbf{a supervised procedure for extracting workflows with LLMs, with checks of the results (methodological contribution).} An extractor and an evaluator LLM instance produce and revise the representation under human supervision. Authors of ten library papers rated the accuracy of the structure of their workflows with $\text{m}=4.80$ on a 5-point scale. Extracting five workflows again with two models reproduces the backbone of a workflow but not the details, so the authors of a workflow should check its representation.

Fourth, \textbf{an exploratory cross-workflow analysis (analytical contribution).} To explore and demonstrate the possibilities enabled by ATWL, we conducted analytical comparison of the library workflows by transitions between intents, closed cycles, and 25 computationally derived features. The regularities we found are preliminary observations, since the library is small and partly drawn from our own work.

Fifth, \textbf{an experiment on the effect of the library and of ATWL on LLM-designed workflows (practical contribution).} Two LLMs designed workflows for two tasks under conditions that vary whether the ATWL specification and the library, in prose or in ATWL form, are provided. Designs made with the library refer to library workflows. ATWL designs declare an exit for every loop, and none of the 17 loops in the prose designs states one. Translating prose designs into ATWL loses conditional paths beyond loops. The results are descriptive and do not measure design quality.

Section~\ref{sec:related_work} discusses related work, and Section~\ref{sec:language} defines ATWL. Section~\ref{sec:library} describes the library, Section~\ref{sec:analysis} analyses it, Section~\ref{sec:recommendation} reports the experiments, and Section~\ref{sec:expert_feedback} reports the feedback from experts. Section~\ref{sec:discussion} discusses the limitations and further work.

\section{Related Work}
\label{sec:related_work}

ATWL draws on and complements a substantial body of research on conceptual models, ontologies, and design frameworks for visual analytics. Rather than reviewing each contribution in isolation, we organise the discussion around five questions that together define ATWL's position in the landscape: (1)~what are the conceptual origins of the present work? (2)~how have researchers modelled the \emph{cognitive} side of VA processes? (3)~how have they modelled the \emph{design} of VA systems? (4)~how have formal languages been used to represent \emph{computational workflows} and their provenance in adjacent fields, and how is analytic provenance characterised in visual analytics? and (5)~what formal representations have been proposed for VA \emph{workflows} themselves?

\begin{table*}[ht]
\centering\footnotesize
\begin{tabular}{@{}l c c c >{\raggedright\arraybackslash}p{6.3cm}@{}}
\toprule
\textbf{Framework} & \textbf{Scope} & \textbf{Cognition} & \textbf{Formalisation} & \textbf{Primary purpose} \\
\midrule
Van Wijk~\cite{vanWijk2005value}
  & Analysis & Modelled & Semi-formal
  & Formalise visualisation value via specification feedback \\
VA as model building~\cite{VAasModelBuilding}
  & Analysis & Observable only & Conceptual
  & Conceptualise VA as goal-oriented model building \\
Sacha et al.\ KG model~\cite{sacha2014knowledge}
  & Analysis & Modelled & Conceptual
  & Explain analyst reasoning \\
He et al.\ SoK~\cite{SoK2026}
  & Analysis & Modelled & Conceptual
  & Insight lifecycle taxonomy \\
NBGM~\cite{meyer2013nested}
  & Design & --- & Semi-formal
  & Compare design decisions \\
Design Activity~\cite{McKenna2014}
  & Design & --- & Descriptive
  & Guide design process \\
IVAS~\cite{ChenEbert2019}
  & Design & --- & Semi-formal
  & Optimise system components \\
Wu et al.~\cite{WuDesignPatterns2025}
  & Design & --- & Semi-formal
  & Extract design patterns \\
CWL, BPMN, VisTrails, Kepler,\\ KNIME~\cite{CWL2022,BPMN2011,VisTrails2006,Kepler2006,KNIME2009}
  & Execution & --- & Formal (executable)
  & Computational reproducibility and portability \\
VIS4ML~\cite{VIS4ML2019}
  & Analysis (ML) & Partial & Formal (OWL)
  & Map VA support in ML \\
Beaucamp et al.~\cite{Beaucamp2025}
  & Analysis & --- & Quantitative
  & Measure process quality \\
W3C PROV~\cite{prov-dm2013}
  & Any activity & --- & Formal (data model)
  & Interchange of provenance information \\
Ragan et al.~\cite{Ragan2016}
  & Analysis (history) & Insight, rationale & Conceptual
  & Organise provenance types and purposes \\
\addlinespace
\textbf{ATWL}
  & Analysis & Observable only & Formal (declarative)
  & Describe, compare, reuse analytical workflows \\
\bottomrule
\end{tabular}
\caption{Positioning of ATWL in respect to related frameworks.} 
\label{tab:related_positioning}
\end{table*}

\subsection{Conceptual Origins}
\label{sec:rw:origins}

The present work develops ideas first introduced by Andrienko et al.~\cite{VAasModelBuilding}, who proposed viewing visual analytics as a goal-oriented model-building process directed at constructing an appropriate behavioural model of a piece of reality. That framework defined a conceptual vocabulary including subjects and their aspects, structural and behavioural models, focus relationships, and model appropriateness criteria, and represented the analytical process as a directed workflow of data transformation, initial model generation, evaluation, and iterative development. Several core ideas of ATWL originate in that framework: the distinction between entities and their attributes, the role of data transformations in making focus relationships observable, the centrality of model evaluation as a driver of iteration, and the explicit treatment of prior knowledge as an input to the process.

A second conceptual origin is van Wijk's formal model of the visualisation process~\cite{vanWijk2005value} introducing \emph{specification-mediated feedback} as the mechanism by which human analytical judgment enters the computational process. ATWL adopts this concept as a core design element (\kw{specification} is one of its eight artifact types) and generalises van Wijk's formulation beyond visualisation parameters to encompass all forms of analyst-produced directives, including feature selections, model configurations, query constraints, and method choices.

However, both frameworks remained \emph{conceptual accounts} of aspects of visual analytics; neither provided a formal language in which concrete workflows could be specified, compared, or computationally processed. ATWL addresses this gap by developing their conceptual vocabulary into a declarative specification language with typed artifacts, defined transform intents, well-formedness constraints, and control structures for iteration and branching. Where the model-building framework identified the need for systematic cataloguing of analytical approaches and their transfer across application domains (Sections~6.1--6.3 of~\cite{VAasModelBuilding}), ATWL provides the representational machinery to realise that vision.

\subsection{Models of Analytical Cognition}
\label{sec:rw:cognition}

Several influential frameworks model visual analytics as a cognitive process. Sacha et al.~\cite{sacha2014knowledge} propose a knowledge generation model that integrates the computer side of VA (data, models, visualisations) with the human cognitive side, representing the analyst's reasoning as three nested loops, exploration, verification, and knowledge generation, driven by internal constructs such as hypotheses, findings, and insights. He et al.~\cite{SoK2026} complement this macro-level view with a micro-level analysis of the insight lifecycle, tracing how individual insights are discovered through interaction with visualisations, externalised through annotation, and communicated through reports and data stories; their distinction between \emph{insights into the data} (correlations, trends) and \emph{insights into the domain} (contextualised understanding) provides a useful taxonomy of analytical products.

ATWL shares with these frameworks the overarching goal of providing a structured account of VA processes, and certain correspondences are direct: ATWL's \emph{pattern} artifacts map onto data-level insights, while \emph{knowledge} artifacts map onto domain-level insights in He et al.'s taxonomy; the computer-side concepts in Sacha et al.'s model (data, models, visualisations) can be seen as coarse-grained precursors of ATWL's eight artifact types. However, ATWL differs from both frameworks in a fundamental design choice: it deliberately restricts itself to what is explicitly externalised by the analyst (recorded statements, specifications, labels, judgements) and does not attempt to describe cognitive processes that remain internal to the human mind. This sacrifices explanatory reach for formal precision: ATWL representations are machine-readable and verifiable against observable analytical products, whereas cognitive models offer richer explanatory power for understanding \emph{why} analysts act as they do, at the cost of concepts that cannot be directly observed or formalised. The restriction aligns with He et al.'s emphasis on \emph{insight externalisation} as the critical bridge between cognition and shareable products: ATWL can be seen as a formal language for representing exactly those externalised artifacts that their framework identifies as the observable outputs of the insight process. The approaches are thus complementary: cognitive models provide the context within which ATWL-encoded workflows are executed; ATWL captures the larger analytical architecture within which individual insights are produced.

\subsection{Models of VA System Design}
\label{sec:rw:design}

A second family of frameworks addresses how VA systems should be \emph{designed}. Meyer et al.~\cite{meyer2013nested} extend Munzner's four-level nested model~\cite{munzner2009nested} into the Nested Blocks and Guidelines Model (NBGM), introducing finer-grained \emph{blocks} (design outcomes) and \emph{guidelines} (relationships between blocks) within and across design levels. McKenna et al.~\cite{McKenna2014} propose the Design Activity Framework, modelling visualisation design as four overlapping activities - understand, ideate, make, and deploy - linked to the nested model's levels. Chen and Ebert~\cite{ChenEbert2019} take an information-theoretic perspective in their ontological framework IVAS, categorising design shortcomings as \emph{symptoms}, reasoning about their \emph{causes}, and identifying \emph{remedies} using 24 abstract entities derived from crossing four system components with three cost--benefit measures. Wu et al.~\cite{WuDesignPatterns2025} conduct a meta-analysis of 220 VA papers to map relationships between analytical requirements and design solutions, extracting problem-driven design patterns organised in knowledge graphs.

ATWL and these design frameworks share a conviction that formalisation enables cross-domain comparison, and all arrive at structured vocabularies that abstract away domain-specific terminology. The key distinction is that design frameworks focus on \emph{system design}: they reason about which abstractions, encodings, and interactions a system should provide; ATWL focuses on \emph{analysis process}: it captures what happens when an analyst uses available methods and tools to progress from data to knowledge. This shift is reflected in classification vocabulary: a clustering algorithm is a ``Clustering \& Grouping'' manipulation in Wu et al.'s solution typology but a \kw{define-unit} transform in ATWL. It is the difference between \emph{what the system does} and \emph{why the analyst does it}.

The two perspectives are complementary. Design frameworks could inform the selection of tools that populate individual ATWL transforms; conversely, ATWL's workflow structures  could provide design frameworks with the analytical process context they currently lack.

\subsection{Scientific Workflow Languages and Provenance Frameworks}

A substantial body of work addresses formal representation of computational workflows. BPMN \cite{BPMN2011} specifies processes as typed task sequences with execution semantics; scientific workflow systems such as Kepler \cite{Kepler2006}, VisTrails \cite{VisTrails2006}, and KNIME \cite{KNIME2009} compose and execute computational pipelines from typed modules; CWL \cite{CWL2022} provides portable pipeline specifications emphasising cross-platform reproducibility; and W3C PROV \cite{prov-dm2013} is a general model of entities, activities, and agents, with relations such as usage, generation, and derivation. PROV is not limited to computational execution: it can describe the provenance of human activities and of any kind of entity, at different levels of abstraction. Its structure resembles ATWL's artifact–transform–actor triad.

ATWL differs in two respects. First, it operates at the level of \emph{analytical intent} rather than computational execution: a clustering step in CWL specifies an algorithm and its parameters; in ATWL it specifies the intent to \kw{define-unit} by similarity-based grouping, enabling comparison regardless of method. Second, ATWL treats human-dominated transforms as first-class typed components, whereas scientific workflow systems model purely computational pipelines \cite{Provenance2008}. For visual analytics, which enables and demands human interpretation, this distinction is essential. An ATWL workflow can be mapped to PROV, with artifacts as entities, transforms as activities, and actors as agents. ATWL adds typed artifacts, analytical intents, and control structures such as loops and exits, which PROV does not define, and CWL could implement machine-dominated ATWL transforms. ATWL contributes the layer these systems do not address: a vocabulary for the analytical logic of human–computer collaboration.

In visual analytics, analytic provenance is the history of an analysis. Ragan et al.~\cite{Ragan2016} organise it by five types of provenance information (data, visualization, interaction, insight, and rationale) and six purposes (recall, replication, action recovery, collaborative communication, presentation, and meta-analysis). They treat information about workflows, both prospective (the steps of a procedure) and retrospective (execution details), as part of data provenance. ATWL describes an analysis at the level of analytical steps and covers parts of four of the five types. Its transforms and the entities and features that they produce describe the history of the data and of the workflow, its arrangements and visualisations describe the views, its patterns and knowledge artifacts describe insights, and its assessments and specifications, together with the criteria of loops, describe decisions and part of their rationale. ATWL does not record interaction provenance, the traces of single user actions. It describes an analysis as reported or designed and contains no execution details, so it is closer to prospective than to retrospective provenance. Ragan et al.\ note that insight and rationale cannot be captured automatically from system logs. ATWL representations are instead obtained from the descriptions of the analysts, by LLM extraction followed by a check by the authors. %

\subsection{Formal Workflow Representations}
\label{sec:rw:workflows}

The most directly related work concerns formal representations of VA workflows themselves.

Sacha et al.~\cite{VIS4ML2019} propose VIS4ML, a formal ontology for VA-assisted machine learning implemented in OWL. VIS4ML decomposes ML workflows into four phases: Prepare-Data, Prepare-Learning, Model-Learning, Evaluate-Model. It represents individual workflows as pathways through a fixed ontological structure, with ``bus stops'' marking points where visualisation assists the ML process. Its bipartite architecture of \emph{Processes} and \emph{IO-Entities} (Data, Model, Knowledge) corresponds directly to ATWL's transform--artifact structure, making VIS4ML the closest predecessor to our work.

The two frameworks differ in three key respects. First, \emph{scope}: VIS4ML is specialised to VA-assisted ML, whereas ATWL is independent of the application domain, covering temporal analysis, movement analytics, text mining, and statistical modelling with the same vocabulary. Second, \emph{granularity}: VIS4ML's three IO-Entity subclasses are coarse-grained compared to ATWL's eight artifact types, which distinguish computed features from raw entities, organisational arrangements from visual representations, and evaluative knowledge from control specifications. Third, \emph{representational strategy}: VIS4ML traces workflows as pathways through a single fixed ontological template, which is powerful for identifying under-explored phases but constrains every workflow to the same structure; ATWL represents each workflow as a standalone specification with its own dependency structure and control flow, enabling exploratory cross-workflow analysis, such as the closed cycles and the feature profiles of Section~\ref{sec:analysis}, that arises from comparing freely structured specifications rather than mapping them onto a predetermined schema.

Beaucamp et al.~\cite{Beaucamp2025} take a complementary approach, translating Chen and Golan's~\cite{ChenGolan2016} information-theoretic cost--benefit framework into a methodology for quantitatively analysing hybrid decision workflows. Their approach decomposes workflows into component processes and measures how well each transforms information, using entropy and divergence to quantify benefit and distortion. While ATWL is a \emph{structural} language capturing what operations are performed, on what objects, in what order, and by whom, Beaucamp et al.\ provide a \emph{quantitative} framework for evaluating how well each operation performs. The two are directly composable: ATWL could supply the structural decomposition that their methodology requires as input, while their information-theoretic measures could enrich ATWL-encoded workflows with quantitative performance annotations.

\subsection{Positioning ATWL}
\label{sec:rw:positioning}

Table~\ref{tab:related_positioning} summarises the positioning of ATWL relative to the reviewed frameworks along four dimensions: temporal scope (design time vs.\ analysis time), treatment of cognition (modelled vs.\ restricted to observables), level of formalisation, and primary purpose.

ATWL occupies a distinctive position in this landscape. To our knowledge, it is the only framework that combines (a)~independence of the application domain, (b)~analysis process focus, (c)~formal, machine-readable syntax, and (d)~restriction to observable, externalised analytical products. The first three properties it shares partially with VIS4ML; the fourth it shares with the model-building framework~\cite{VAasModelBuilding} from which ATWL descends and, in spirit, with He et al.'s emphasis on insight externalisation. Their conjunction into a formally precise language for the observable structure of analytical processes, independent of the application domain, supports the uses examined in this paper: LLM-assisted extraction of workflow representations from research literature, an exploratory cross-workflow analysis of the resulting library, and LLM-assisted workflow design.

\section{Language Design}
\label{sec:language}

This section presents the design of ATWL in three parts: the conceptual model that defines the general structure of workflows, the ontologies that classify the building blocks of analytical processes, and a concise overview of the language syntax. The complete definition of the language is available at the URL \url{https://geoanalytics.net/VAworkflows/ATWL_definition.pdf}, the language design process is described in Section~\ref{sec:language:design:process}.

\subsection{Conceptual Model}
\label{sec:conceptual-model}

The central abstraction in ATWL is the \emph{analytical workflow} consisting of two kinds of building blocks: informational objects, called \emph{artifacts}, and operations, called \emph{transforms}, producing new artifacts from existing ones.  Each transform consumes one or more input artifacts and produces one or more output artifacts; the resulting input--output dependencies capture the logical progression of analysis from raw data to derived insights. Figure~\ref{fig:layer-cake} gives an overview of the conceptual architecture with artifacts as nodes and transforms as directed edges, as described below.

Artifacts represent different kinds of objects with different analytical roles, %
structured across five layers: data, organisation, representation, interpretation, and epistemic. 
Each artifact has a \emph{type} drawn from a fixed ontology (Section~\ref{sec:artifact-ontology}, Table \ref{tab:artifact-types}), which constrains its internal description and determines how it can participate in transforms.

Transforms are classified not by computational method or algorithm but by their analytical \emph{intent}, drawn from a fixed ontology (Section~\ref{sec:transform-ontology}, Table \ref{tab:intents}). 
The intent declares the purpose of the transform (what it achieves analytically), enabling structural comparison of workflows that employ different methods to achieve the same analytical goal.

Every artifact in a workflow is either \emph{exogenous} or \emph{derived}. Exogenous artifacts, such as raw datasets, prior domain knowledge, and initial parameter settings, are supplied as inputs to the workflow and are marked with the designation \kw{origin:~given}. Derived artifacts are produced as outputs of transforms. 
This distinction makes explicit the boundary between exogenous inputs and artifacts derived within the workflow.

The design of ATWL is guided by four principles:

\begin{enumerate}%
\item \textbf{Distinct artifact roles.}
The artifact ontology assigns each artifact to one of five roles in the analytical process: data (entities, features), organisation (arrangements), representation (visualisations), interpretation (patterns, models), and epistemic output (knowledge, specifications). As illustrated by the layered structure in Figure~\ref{fig:layer-cake}, this stratification makes explicit the progressive transitions between data manipulation, visual encoding, and human reasoning that are typically conflated in prose descriptions.  %
The layers are bridged upward by transforms that enact an analytical ascent (e.g., \emph{visualise} maps data and organisational artifacts into representations), and downward by feedback, through which specifications and knowledge at the epistemic layer re-enter earlier processing stages.
    
\item \textbf{Intent-driven transforms.}
Transforms are classified by their analytical purpose rather than their computational implementation. A clustering algorithm and a manual grouping by a domain expert may both realise the same intent (\kw{define-unit}) despite radically different implementations.

\item \textbf{Knowledge as a first-class artifact.}
Explicitly externalised knowledge in the form of recorded statements, judgments, specifications, labels, rules, etc., is given the same formal status as data, features, and visualisations: it is typed, named, and connected to transforms as input or output.  At the same time, ATWL does not attempt to describe cognitive processes that remain internal to the analyst's mind, unlike conceptual frameworks that model hypothesised mental activities (e.g., \cite{sacha2014knowledge}).  Every element of a workflow description thus corresponds to an observable, communicable analytical product.
    
\item \textbf{Domain independence through composable primitives.}
Artifact types and transform intents are defined at a level of abstraction that surpasses specific application domains while remaining sufficiently expressive to capture the essential structure of diverse workflows.
\end{enumerate}

\textbf{Scope.}
An ATWL workflow describes the analytical logic of an analysis as it is reported or designed. It is not a record of a session. Real analyses involve improvisation, mistakes, and abandoned attempts, and analysts can follow different paths through the same design. ATWL expresses such variation only where the design anticipates it: loops describe repeated refinement, and conditionals describe steps that depend on a judgement or on a property of an artifact.

\newsavebox{\boxIconEntities}
\newsavebox{\boxIconFeature}
\newsavebox{\boxIconArrangement}
\newsavebox{\boxIconVisualisation}
\newsavebox{\boxIconPattern}
\newsavebox{\boxIconModel}
\newsavebox{\boxIconKnowledge}
\newsavebox{\boxIconSpecification}

\newsavebox{\boxIconDefineUnit}
\newsavebox{\boxIconCharacterise}
\newsavebox{\boxIconContextualise}
\newsavebox{\boxIconVisualise}
\newsavebox{\boxIconAbstract}
\newsavebox{\boxIconBuildModel}
\newsavebox{\boxIconGenerateKnowledge}
\newsavebox{\boxIconAssess}

\sbox{\boxIconEntities}{\iconEntities}
\sbox{\boxIconFeature}{\iconFeature}
\sbox{\boxIconArrangement}{\iconArrangement}
\sbox{\boxIconVisualisation}{\iconVisualisation}
\sbox{\boxIconPattern}{\iconPattern}
\sbox{\boxIconModel}{\iconModel}
\sbox{\boxIconKnowledge}{\iconKnowledge}
\sbox{\boxIconSpecification}{\iconSpecification}

\sbox{\boxIconDefineUnit}{\iconDefineUnit}
\sbox{\boxIconCharacterise}{\iconCharacterise}
\sbox{\boxIconContextualise}{\iconContextualise}
\sbox{\boxIconVisualise}{\iconVisualise}
\sbox{\boxIconAbstract}{\iconAbstract}
\sbox{\boxIconBuildModel}{\iconBuildModel}
\sbox{\boxIconGenerateKnowledge}{\iconGenerateKnowledge}
\sbox{\boxIconAssess}{\iconAssess}

\begin{figure}[t]
\centering
\begin{tikzpicture}[
  >=Stealth,
  artifact/.style={
    draw=#1, thick,
    fill=#1!10,
    rounded corners=2.5pt,
    minimum width=1.5cm,
    minimum height=0.9cm,
    align=center,
    font=\scriptsize\sffamily,
    inner sep=2pt,
  },
  txlbl/.style={
    font=\tiny\sffamily,
    atwlTransform,
    inner sep=1.5pt,
    fill=white,
    fill opacity=0.6,
    text opacity=1,
    rounded corners=1pt,
  },
  txarrow/.style={
    ->,
    atwlTransform,
    semithick,
    shorten >=2pt,
    shorten <=2pt,
  },
]

\def\W{8.2}
\def\H{1.25}
\def\G{0.5}

\pgfmathsetmacro{\yA}{0}
\pgfmathsetmacro{\yB}{\yA+\H+\G}
\pgfmathsetmacro{\yC}{\yB+\H+\G}
\pgfmathsetmacro{\yD}{\yC+\H+\G}
\pgfmathsetmacro{\yE}{\yD+\H+\G}

\fill[atwlEntities!8, rounded corners=3pt]
  (0,\yA) rectangle (\W,\yA+\H);
\node[anchor=north west, font=\scriptsize\sffamily\bfseries,
      atwlEntities!35, inner sep=3pt]
  at (0.05,\yA+\H) {Data};

\fill[atwlArrangement!8, rounded corners=3pt]
  (0,\yB) rectangle (\W,\yB+\H);
\node[anchor=north west, font=\scriptsize\sffamily\bfseries,
      atwlArrangement!50, inner sep=3pt]
  at (0.05,\yB+\H) {Organisation};

\fill[atwlVisualisation!8, rounded corners=3pt]
  (0,\yC) rectangle (\W,\yC+\H);
\node[anchor=north west, font=\scriptsize\sffamily\bfseries,
      atwlVisualisation!40, inner sep=3pt]
  at (0.05,\yC+\H) {Representation};

\fill[atwlPattern!8, rounded corners=3pt]
  (0,\yD) rectangle (\W,\yD+\H);
\node[anchor=north west, font=\scriptsize\sffamily\bfseries,
      atwlPattern!40, inner sep=3pt]
  at (0.05,\yD+\H) {Interpretation};

\fill[atwlKnowledge!8, rounded corners=3pt]
  (0,\yE) rectangle (\W,\yE+\H);
\node[anchor=north west, font=\scriptsize\sffamily\bfseries,
      atwlKnowledge!40, inner sep=3pt]
  at (0.05,\yE+\H) {Epistemic};

\node[artifact=atwlEntities] (ent)
  at (2.5, \yA+\H/2)
  {\usebox{\boxIconEntities}\\[-1pt]Entities};
\node[artifact=atwlFeature] (fea)
  at (5.9, \yA+\H/2)
  {\usebox{\boxIconFeature}\\[-1pt]Feature};

\node[artifact=atwlArrangement] (arr)
  at (3.5, \yB+\H/2)
  {\usebox{\boxIconArrangement}\\[-1pt]Arrangement};

\node[artifact=atwlVisualisation] (vis)
  at (3.5, \yC+\H/2)
  {\usebox{\boxIconVisualisation}\\[-1pt]Visualisation};

\node[artifact=atwlPattern] (pat)
  at (3.0, \yD+\H/2)
  {\usebox{\boxIconPattern}\\[-1pt]Pattern};
\node[artifact=atwlModel] (mod)
  at (5.8, \yD+\H/2)
  {\usebox{\boxIconModel}\\[-1pt]Model};

\node[artifact=atwlKnowledge] (kno)
  at (3.0, \yE+\H/2)
  {\usebox{\boxIconKnowledge}\\[-0pt]Knowledge};
\node[artifact=atwlSpecification] (spe)
  at (5.8, \yE+\H/2)
  {\usebox{\boxIconSpecification}\\[-1pt]Specification};

\draw[txarrow]
  ([yshift=-0.15cm]ent.west) to[out=210, in=240, looseness=3.5]
  node[txlbl, pos=-0.01, left=3pt]
    {\usebox{\boxIconDefineUnit}\; define-unit}
  ([xshift=-0.25cm]ent.south);
  
\draw[txarrow]
  ([yshift=-0.25cm]ent.east) -- ([yshift=-0.25cm]fea.west)
  node[txlbl, midway, above=2pt]
    {\usebox{\boxIconCharacterise}\; characterise};

\draw[txarrow] (3.4, \H) -- (3.4, \H+\G) 
    node[txlbl, midway, right=3pt] {\usebox{\boxIconContextualise}\; contextualise};    

\draw[txarrow] (2.0, \H) -- (2.0, 2*\H+2*\G) 
    node[txlbl, pos=0.9, right=3pt] {\usebox{\boxIconVisualise}\; visualise};    

\draw[txarrow] (3.75, \H+\G+\H) -- (3.75, 2*\H+2*\G) 
    node[txlbl, midway, right=3pt] {\usebox{\boxIconVisualise}\; visualise};    

\draw[txarrow] (2.8, 2*\H+2*\G+\H) -- (2.8, 3*\H+3*\G) 
    node[txlbl, midway, right=3pt] {\usebox{\boxIconAbstract}\; abstract};

\draw[txarrow] (5.4, \H) -- (5.4, 3*\H+3*\G) 
    node[txlbl, pos=0.94, right=3pt] {\usebox{\boxIconBuildModel}\; build-model};
 
\draw[txarrow] (2.8, 3*\H+3*\G+\H) -- (2.8, 4*\H+4*\G) 
    node[txlbl, midway, right=3pt] {\usebox{\boxIconGenerateKnowledge}\; generate-knowledge};

\draw[txarrow] (5.5, 3*\H+3*\G+\H) -- (5.5, 4*\H+4*\G) 
    node[txlbl, midway, right=3pt] {\usebox{\boxIconAssess}\; assess};

\draw[txarrow, dashed, atwlSpecification!70!black,
      rounded corners=4pt]
  ([xshift=+0.2cm]spe.east)  -| (\W-0.4, \yB+\H/2)
  -- (\W-0.4, 0.5*\H) -- (\W-1.2, 0.5*\H)
  node[txlbl, anchor=south, rotate=90] at (8.1, \yC+\H/2)
    {\small\itshape feedback};

\draw[txarrow, dashed, atwlSpecification!70!black,
      rounded corners=4pt]
  ([xshift=+0.2cm]spe.east)  -| (\W-0.4, \yD+\H/2)
  -- (\W-0.4, 1*\H+0.5*\G) -- (\W-1.2, 1*\H+0.5*\G);

\draw[txarrow, dashed, atwlSpecification!70!black,
      rounded corners=4pt]
  ([xshift=+0.2cm]spe.east)  -| (\W-0.4, \yD+\H/2)
  -- (\W-0.4, 2*\H+1.5*\G) -- (\W-1.2, 2*\H+1.5*\G);
    
\draw[txarrow, dashed, atwlSpecification!70!black,
      rounded corners=4pt]
  ([xshift=+0.2cm]spe.east)  -| (\W-0.4, \yD+\H/2)
  -- (\W-0.4, 3*\H+2.5*\G) -- (\W-1.2, 3*\H+2.5*\G);

\draw[atwlTransform!40, thick, ->,
      line cap=round, shorten >=3pt, shorten <=3pt]
  (8.3, \yA+0.2) -- (8.25, \yE+\H-0.2);
\node[anchor=south, rotate=90,
      font=\small\sffamily\itshape, atwlTransform!50]
  at (8.7, \yC) {analytical ascent};

\end{tikzpicture}
\caption{Conceptual architecture of ATWL. Five analytical-role layers contain eight artifact types (nodes); transform intents (edges) connect them across layers, ascending from raw data to epistemic outcomes. }
\vspace{6pt}
\label{fig:layer-cake}
\end{figure}

\subsection{Artifact Ontology}
\label{sec:artifact-ontology}

ATWL defines eight artifact types that collectively span the full lifecycle of a visual analytics process, structured along five analytical role layers. Table~\ref{tab:artifact-types} provides an overview. 

\begin{table}[t]
\centering\footnotesize
\renewcommand{\arraystretch}{1.2}
\begin{tabular}{@{}l >{\raggedright\arraybackslash}p{0.65\columnwidth}@{}}
\toprule
\textbf{Artifact type} & \textbf{Description} \\
\midrule

\multicolumn{2}{@{}l}{\itshape \underline{Data}} \\ 
\iconEntities\ Entities      
& Collections of identifiable objects, each treated as a unit of analysis \\

\iconFeature\ Feature       
& Explicit descriptor of entity properties or relationships \\

\addlinespace
\multicolumn{2}{@{}l}{\itshape \underline{Organisation}} \\

\iconArrangement\ Arrangement   
& Positioning of entities within a reference context \\

\addlinespace
\multicolumn{2}{@{}l}{\itshape \underline{Representation}} \\

\iconVisualisation\ Visualisation 
& External visual representation for human perception \\

\addlinespace
\multicolumn{2}{@{}l}{\itshape \underline{Interpretation}} \\

\iconPattern\ Pattern       
& Regularity, trend, or structure identified through abstraction \\

\iconModel\ Model         
& Formal/computational representation for prediction or explanation \\

\addlinespace
\multicolumn{2}{@{}l}{\itshape \underline{Epistemic}} \\

\iconKnowledge\ Knowledge     
& Explicitly formulated substantive knowledge: insights, judgments, expertise \\

\iconSpecification\ Specification 
& Control directives: parameters, constraints, or method choices directing transforms \\

\bottomrule
\end{tabular}
\caption{The eight ATWL artifact types, organised by analytical role.}
\vspace{12pt}
\label{tab:artifact-types}
\end{table}

\begin{table}[t]
\centering\footnotesize
\renewcommand{\arraystretch}{1.2}
\begin{tabular}{@{}l >{\raggedright\arraybackslash}p{0.48\columnwidth}@{}}
\toprule
\textbf{Intent} & \textbf{Purpose} \\
\midrule

\iconDefineUnit\ \kw{define-unit}
  & Create or redefine entities as units of analysis \\

\iconCharacterise\ \kw{characterise}
  & Compute or transform features describing entities \\

\iconContextualise\ \kw{contextualise}
  & Position entities within a reference context \\

\iconVisualise\ \kw{visualise}
  & Create visual representations for human perception \\

\iconAbstract\ \kw{abstract}
  & Derive patterns or conceptual structures \\

\iconBuildModel\ \kw{build-model}
  & Construct or refine a formal/computational model \\

\iconGenerateKnowledge\ \kw{generate-knowledge}
  & Formulate explicit knowledge or specifications \\

\iconAssess\  \kw{assess}
  & Evaluate quality or appropriateness of artifacts \\

\bottomrule
\end{tabular}
\caption{The eight ATWL transform intents characterising analytical purpose rather than implementation.}
\label{tab:intents}
\end{table}

\subsubsection{Data Layer: Entities and Features}

\emph{Entities} represent the fundamental units of analysis. Each entities artifact describes a collection of analytical objects characterised along three orthogonal dimensions.

\textbf{Internal structure} describes how components within each entity are organised. Table~\ref{tab:entity-structure} lists the six structure types, which fall into three categories: \emph{atomic} entities are indivisible; \emph{container} entities (groups, episodes, regions) enclose components in an unstructured or implicitly structured manner; and \emph{relational} entities (sequences, formations) organise components through explicit relational structure.

\begin{table}[ht]
\centering\footnotesize
\begin{tabularx}{\columnwidth}{@{}l l >{\raggedright\arraybackslash}X >{\raggedright\arraybackslash}X @{}}
\toprule
\textbf{Category} & \textbf{Type} & \textbf{Description} & \textbf{Examples} \\
\midrule
Atomic & \kw{elementary} & Indivisible object; internal composition not analysed. & Events, measurements, whole images. \\
\midrule
\multirow{5}{*}{Container} & \kw{group} & Unordered collection without internal structure. & Clusters of days, related object groups. \\
\cmidrule(lr){2-4}
& \kw{episode} & Bounded time interval with referenced components. & Time series segments, sports match episodes. \\
\cmidrule(lr){2-4}
& \kw{region} & Bounded spatial extent with spatial components. & City districts, spatial measurement cells. \\
\midrule
\multirow{4}{*}{Relational} & \kw{sequence} & Components in a linear, essential order. & Event sequences, sentence words, task lists. \\
\cmidrule(lr){2-4}
& \kw{formation} & Networks, hierarchies, or spatial neighbourhoods. & Network snapshots, hierarchies, team formations. \\
\bottomrule
\end{tabularx}
\caption{Six types of internal structure for entities, categorized by Atomic, Container, and Relational.}
\vspace{12pt}
\label{tab:entity-structure}
\end{table}

\textbf{Embedment} describes the shared environment in which the entities of a collection reside and relate to one another. We use the term \emph{embedment} (the state of being embedded) on purpose: it avoids confusion with embeddings in machine learning. Table~\ref{tab:entity-embedment} lists the five embedment types. Multiple embedment types may co-occur, for example, entities embedded in both time and space are denoted \kw{\{time, space\}}. Embedment is omitted when the artifact represents a single entity.

\begin{table}[t]
\centering\footnotesize
\begin{tabular}{@{}l >{\raggedright\arraybackslash}p{0.5\columnwidth}@{}}
\toprule
\textbf{Embedment} & \textbf{Description} \\
\midrule
\kw{set}       & Unordered collection \\
\kw{sequence}  & Linearly ordered positions \\
\kw{time}      & Temporal axis \\
\kw{space}     & Spatial reference \\
\kw{relational structure} & Network, hierarchy, or similar \\
\bottomrule
\end{tabular}
\caption{Embedment types for entities.}
\label{tab:entity-embedment}
\end{table}

\textbf{Features within entities.} A declaration of an entities artifact may include one or more features. Every feature declaration specifies a \emph{value structure} (how the values are organised: \kw{atomic}, \kw{list}, \kw{vector}, \kw{matrix}, or \kw{relational configuration}) and, optionally, a \emph{value type} (the nature of the atomic components: \kw{numeric}, \kw{ordinal}, \kw{categorical}, \kw{temporal}, \kw{spatial}, \kw{text}, or \kw{reference}). When atomic components are of mixed types, set notation is used (e.g., \kw{\{numeric, temporal\}}).

\textbf{Features} artifacts describe properties of entities or relationships between them that are important for subsequent analysis. Internal features (inside an \kw{entities} artifact) provide background semantics; \kw{feature} artifacts are used when features become \emph{first-class operands} of transforms, i.e., their inputs or outputs. Feature artifacts are typically produced by transforms with intent \kw{characterise} (Section~\ref{sec:transform-ontology}).

A \kw{feature} artifact references the \kw{entities} artifact it describes and can have the same value-structure and value-type descriptors as internal features of \kw{entities} artifacts. It may additionally specify a \emph{representation form} that clarifies encoding for complex features (e.g., ``distance matrix'', ``topic-distribution vector'', ``$k$-NN graph'').

\subsubsection{Organization Layer: Arrangements}

\textbf{Arrangements} organise entities in a reference structure (context), establishing positions or placements to support analysis. It does not create new entities or values; it defines how existing entities are positioned within a reference structure, such as a calendar, geographic space, grid, or artificial projection space resulting from dimensionality reduction. The reference structure itself must be previously explicitly declared as an \kw{entities} artifact. The \emph{principle} field in the declaration of an \kw{arrangement} artifact describes the organising logic, e.g., ``calendar(year, month, weekday)'', ``2D projection based on feature similarity''.

\subsubsection{Representation Layer: Visualisations}

\textbf{Visualisations} are an external visual representation of an arrangement, designed for human perception. It references the artifact(s) it depicts and is characterised by three properties: \emph{layout} (the structure of the visual space, e.g., ``calendar grid'', ``node-link layout''), \emph{form} (the type of visual marks, e.g., ``coloured cells'', ``line segments''), and \emph{encoding} (the mapping of data attributes to visual channels, e.g., ``colour $\to$ category, size $\to$ frequency'').

\textbf{Note}: The concepts \textit{'embedment'}, \textit{'arrangement'}, and \textit{'visualisation'} capture distinct levels of organisation. \emph{Embedment} is a data-level property of entities: the environment in which they inherently reside (time, space, a relational structure). \emph{Arrangement} is an analytic-level artifact produced by a \kw{contextualise} transform: it maps entities onto positions in a context structure (calendar cells, geographic partitions, projection coordinates) and can serve as input to further computation or as a basis for visualisation. \emph{Visualisation} is a perceptual-level artifact produced by a \kw{visualise} transform: it maps artifacts into visual space, deriving layout from an arrangement when one exists or directly from entities or features otherwise. The three levels form a progression: inherent structure $\to$ analytical organisation $\to$ perceptual encoding.

\subsubsection{Interpretation Layer: Patterns and Models}

\textbf{Patterns} represent regularities, trends, or structures identified through abstraction performed by automated algorithms, human perception, or both. They reference the artifacts from which they are derived and declare a \emph{representation form}, e.g., ``ranked list of motifs'', ``textual descriptions'', ``cluster labels''.

\textbf{Models} are formal or computational representations used for prediction, simulation, or explanation. They specify a \emph{model type} (e.g., ``classifier'', ``topic model'', ``regression model'') and optionally a representation form (e.g., ``decision tree'', ``neural network weights'').

\textbf{Distinction between models and patterns:} Patterns describe \textit{observed} regularities \textit{within} the available data (descriptive and bounded by observations), while models provide mechanisms that \textit{generalise beyond} the available data. Models support interpolation between available data points and extrapolation to new contexts (future time points, unobserved spatial locations, new populations or products, etc.).

\subsubsection{Epistemic Layer: Knowledge and Specifications}

\textbf{Knowledge} artifacts represent explicitly formulated understanding. They may be \emph{derived} during analysis (insights, explanations, rules, decisions, recommendations) or \emph{injected} by the analyst or from external sources (domain assumptions, constraints, rankings, labels, feedback).  Derived knowledge is typically produced by \kw{generate-knowledge} transforms; an important sub-case is \emph{evaluative} knowledge produced by \kw{assess} transforms, which captures quality judgments or adequacy decisions about other artifacts.  Injected knowledge is marked \kw{origin:~given} and enters the workflow as input to transforms, making the role of prior expertise formally explicit. 

\textbf{Specifications} represent parameters, configurations, constraints, or method choices that control how transforms operate (e.g., distance thresholds, number of clusters, aggregation strategies, desired output properties).  They encode \emph{control directives}, i.e., decisions about \emph{how} subsequent analysis should be carried out, and are consumed by transforms whose behaviour depends on configurable choices.  Specifications may be exogenous (\kw{origin:~given}) or derived: most commonly by \kw{generate-knowledge} transforms in which the analyst formulates a methodological decision, or by \kw{assess} transforms that trigger parameter adjustments.  Within iterative loops, specifications are frequently updated as the analyst progressively refines the analytical strategy.

The distinction between the two epistemic types reflects different functional roles: \kw{knowledge} captures \emph{what is understood}, while \kw{specification} captures \emph{what is decided}. The role of specifications as mediating artifacts between human judgment and computational execution was formalised by van Wijk~\cite{vanWijk2005value}, whose model of the visualisation process represents the analyst's decisions as specifications that control image generation. ATWL generalises this concept beyond visualisation parameters to encompass all forms of machine-consumable directives produced by human analytical reasoning, including method choices, constraints, selection criteria, and model configurations.
\subsection{Transform Ontology}
\label{sec:transform-ontology}

Transforms consume the input artifacts and produce output artifacts, towards the upper layers in the ATWL architecture.
Each transform is characterised by three properties: \emph{intent}, \emph{manner}, and \emph{actor}.

\textbf{Intent} is the main property, indicating the analytical purpose of the transform. Table~\ref{tab:intents} lists and describes the eight generic intents defined in ATWL. %
\textbf{Manner} optionally specialises the intent with a free-text description of how the transform achieves its purpose, when appropriate. For example, a \kw{define-unit} transform might specify manner as ``cluster-by-similarity'' or ``time-partitioning''; a \kw{contextualise} transform might specify ``projection-based'' or ``calendar-based''. Manner provides methodological detail without expanding the intent vocabulary, preserving the ability to compare workflows at the level of intent alone.

\textbf{Actor} specifies the agency responsible for the transform: \kw{human} (analyst only), \kw{machine} (computation only), or \kw{hybrid} (human--machine collaboration). This dimension makes explicit the distribution of cognitive and computational labour across the workflow and opens up the ATWL language for human-AI interaction and collaboration analysis.

\subsection{Syntax and Control Structures}
\label{sec:syntax-overview}

ATWL uses a declarative, YAML-like syntax (summarised in Appendix A). A workflow specification begins with a \kw{workflow} declaration and consists of interleaved artifact and transform declarations. Listing~\ref{lst:example} illustrates the core syntax patterns with a minimal workflow fragment.

\begin{lstlisting}[style=atwl, caption={An illustrative ATWL fragment.}, label={lst:example}]
workflow example_workflow

artifact D_events : entities
  origin: given
  internal structure: elementary
  embedment: {set, time}
  features:
    - id: timestamp
      value structure: atomic
      value type: temporal
    - id: category
      value structure: atomic
      value type: categorical
  description: "Timestamped event records"

transform T1 :
  intent: characterise
  manner: "aggregate by time period"
  input: D_events
  output: F_daily
  actor: machine

artifact F_daily : feature(D_events)
  value structure: vector
  value type: numeric
  description: "Daily counts per category"

transform T2 :
  intent: visualise
  input: D_events, F_daily
  output: V_timeline
  actor: machine

artifact V_timeline :
    visualisation(D_events, F_daily)
  layout: "temporal axis"
  form: "stacked area chart"
  encoding: "x: date, y: count, colour: category"
\end{lstlisting}

Artifact declarations specify the artifact's type and, for exogenous artifacts, the marker \kw{origin:~given}. The type keyword may be followed by parenthesised references to related artifacts (e.g., \kw{feature(D\_events)}). Transform declarations specify intent, optional manner, input and output artifact identifiers (comma-separated), and actor.

\textbf{Control structures.} ATWL provides three constructs for expressing iterative and conditional workflow logic:

\begin{itemize}[nosep]
  \item \textbf{Loops} (\kw{loop\,\ldots\,end\,loop}) declare iterative refinement with a named identifier and a qualitative stopping condition (\kw{until}). The loop body may contain an explicit \kw{assess} transform with a conditional exit, or rely on implicit analyst-monitored termination.

  \item \textbf{Conditionals} (\kw{if\,\ldots\,then\,\ldots\,else}) express branching based on artifact properties or human judgments. The directive \kw{exit\,loop\,<ID>} in a branch terminates an enclosing loop.

  \item \textbf{Assignments} (\kw{assign}) bind an artifact identifier to a new version. They appear before loops (for initialisation) or inside loop bodies (for iterative update), providing the mechanism by which the otherwise acyclic dependency graph accommodates iteration.
\end{itemize}

\textbf{Workflow template.} The \kw{workflow} declaration begins with a \kw{template} field: a high-level summary of the main analytical stages expressed as a chain of intents (e.g., \kw{define-unit\,\textrightarrow\,characterise\,\textrightarrow\,visualise\,\textrightarrow\,abstract}). Iterative stages may be enclosed in \kw{loop(\ldots)} notation. The template serves as a compact signature of the workflow's analytical strategy.

\textbf{Validity.} A formalized ATWL workflow satisfies four structural constraints: (1)~all artifact identifiers are unique; (2)~every transform input references a declared artifact; (3)~exogenous artifacts never appear as transform outputs; and (4)~the chain of input-output dependencies among artifacts and transforms is acyclic, except through explicit \kw{assign} statements within loops. A web-based checker (in Web supplement, \url{https://geoanalytics.net/VAworkflows/}) applies 45 checks to an ATWL specification. 14 of them concern syntax, structure, and references to declared artifacts and report errors. The other 31 report warnings about unknown vocabulary, the types of artifact references and outputs, required fields, the structure of the dependency graph, loops, and the origin of artifacts. 10 checks offer an automatic correction where the correction is obvious, for example for misspelled keywords. 

\subsection{Language Design Process}
\label{sec:language:design:process}

ATWL design evolved through iterations of conceptualisation, formalisation, and testing against published workflows. A minimal set of artifact types and intents was identified that is sufficient to describe the analytical workflows observed in practice. The most significant transition was from an operation-centric formalism, in which a large fixed set of named operations were the primary constructs, to the artifact--transform duality. Workflow meaning is determined primarily by the types and roles of the artifacts produced and not by operational details, and a large set of operations was deemed too complex to motivate analysts to document their workflows. Transforms therefore became uniform connectors characterised by intent, manner, and actor, while artifacts carry the semantic differentiation through their types. Further decisions resolved ambiguities encountered during formalisation, such as separating \kw{define-unit} from \kw{abstract}, separating \kw{contextualise} from \kw{visualise}, and introducing \kw{assess}. The final eight artifact types and eight intents are a deliberate reduction from a larger initial set: intents that produce the same artifact type under the same analytical logic were merged into one intent with a free-text manner.

The language was developed with the first 18 workflows of the library. After the first 16, an audit by a human expert and a new LLM instance proposed corrections; their disagreements were discussed among the co-authors and led to refinements of the definition (Appendix~\ref{sec:library-construction}). The patterns of errors found in the audit shaped the extraction and review instructions, which two fresh LLM instances followed to produce correct representations with minimal intervention. The language did not change for the 36 workflows added afterwards. The alternatives that we considered and the reasons for our decisions are listed in Appendix~\ref{app:language-design}.

\begin{figure}[ht]
\centering
\begin{tikzpicture}[
  >=Stealth,
  art/.style={
    draw=#1, semithick,
    fill=#1!8,
    rounded corners=2pt,
    minimum width=1.2cm,
    minimum height=0.7cm,
    align=center,
    font=\tiny\sffamily,
    inner sep=1.5pt,
  },
  given/.style={
    art=#1,
    double, double distance=1pt,
  },
  tx/.style={
    ->, semithick, atwlTransform,
    shorten >=1.5pt, shorten <=1.5pt,
  },
  txl/.style={
    font=\tiny\sffamily,
    atwlTransform,
    inner sep=1pt,
    fill=white, fill opacity=0.5, text opacity=1,
    rounded corners=1pt,
  },
  loopbox/.style={
    draw=atwlTransform!50, thick, dashed,
    rounded corners=6pt,
    inner sep=5pt,
  },
]

\begin{scope}[shift={(-1.5,0)}]

\node[given=atwlEntities] (Dhour) at (2.0, 0)
  {\usebox{\boxIconEntities}\\[-1pt]D\_hour};
\node[given=atwlEntities] (Dcal) at (0, 0)
  {\usebox{\boxIconEntities}\\[-1pt]D\_calendar};
\node[given=atwlSpecification] (Sclust) at (4.50, -3.7)
  {\usebox{\boxIconSpecification}\\[-1pt]S\_clustering};

\node[art=atwlEntities] (Dday) at (2.0, -1.3)
  {\usebox{\boxIconEntities}\\[-1pt]D\_day};

\draw[tx] (Dhour) -- (Dday)
  node[txl, midway, left=2pt] {\scalebox{0.85}{\usebox{\boxIconDefineUnit}}};

\node at (0.0, -1.6) (Tcon) {\usebox{\boxIconContextualise}};

\node[art=atwlArrangement] (Acal) at (0.0, -2.6)
  {\usebox{\boxIconArrangement}\\[-1pt]A\_calendar};
\node[art=atwlFeature] (Fprof) at (2.0, -2.6)
  {\scalebox{0.9}{\usebox{\boxIconFeature}}\\[1pt]F\_profile};

\draw[tx, shorten >=-1pt] (Dday) -- (Tcon);
\draw[tx, shorten >=-1pt] ([yshift=4pt]Tcon.south) -- (Acal);
\draw[tx, shorten >=-2pt] (Dcal) -- (Tcon);
\draw[tx] (Dday) -- (Fprof)
  node[txl, midway, right=2pt] {\usebox{\boxIconCharacterise}};

\begin{scope}[on background layer]
  \node[loopbox, fit=(Sclust)
    ({-1.0,-3.9}) ({6.3,-9.4}),
    label={[font=\tiny\sffamily\itshape, atwlTransform!70, xshift=0.5cm, yshift=-0.5cm]
           above left:Loop L1}] (loop) {};
  \node[font=\small\sffamily\itshape, atwlTransform!70,
        anchor=south east]
    at (6.3, -9.6) {\scalebox{1.6}{$\circlearrowleft$}};
\end{scope}

\node at (2.0, -3.6) (Tclust) {\usebox{\boxIconDefineUnit}};

\node[art=atwlEntities] (Dclust) at (3.0, -4.5)
  {\usebox{\boxIconEntities}\\[-1pt]D\_cluster};
\node[art=atwlFeature] (Flabel) at (1.2, -4.5)
  {\usebox{\boxIconFeature}\\[-1pt]F\_clust\_label};

\draw[tx, shorten >=-1pt] (Fprof.south) -- ([yshift=-4pt]Tclust.north);
\draw[tx] ([xshift=-3pt, yshift=6pt]Tclust.south east) -- (Dclust.north);
\draw[tx] ([xshift=3pt, yshift=6pt]Tclust.south west) -- (Flabel.north);
\draw[tx] ([yshift=0.1cm] Sclust.west) -- ([xshift=-4pt]Tclust.east);

\node[art=atwlFeature] (Fcprof) at (3.0, -5.9)
  {\scalebox{0.85}{\usebox{\boxIconFeature}}\\[1pt]F\_cl\_profile};

\draw[tx] (Dclust) -- (Fcprof)
  node[txl, midway, right=2pt] {\usebox{\boxIconCharacterise}};

\node at (0.0, -6.0) (Tvis) {\usebox{\boxIconVisualise}};

\node[art=atwlVisualisation] (Vcal) at (0.0, -7.2)
  {\scalebox{0.9}{\usebox{\boxIconVisualisation}}\\[1pt]V\_calendar};
\node[art=atwlVisualisation] (Vprof) at (3.0, -7.2)
  {\scalebox{0.9}{\usebox{\boxIconVisualisation}}\\[1pt]V\_profiles};

\draw[tx, shorten >=-1pt] (Acal.south) -- ([yshift=-3pt]Tvis.north);
\draw[tx, shorten >=-1pt] (Flabel.south) -- ([xshift=0.2cm, yshift=-5pt]Tvis.north);
\draw[tx, shorten >=-1pt] ([yshift=5pt]Tvis.south) -- (Vcal.north);

\draw[tx] (Fcprof) -- (Vprof)
  node[txl, midway, right=2pt] {\usebox{\boxIconVisualise}};

\node at (1.5, -8.0) (Tabs) {\usebox{\boxIconAbstract}};

\node[art=atwlPattern] (Ppat) at (1.5, -8.9)
  {\usebox{\boxIconPattern}\\[-1pt]P\_patterns};
\draw[tx] (Tabs.south) --  (Ppat.north);

\draw[tx, shorten >=-1pt] ([xshift=0.2cm]Vcal.south) -- ([xshift=0.1cm, yshift=0.1cm]Tabs.west);
\draw[tx, shorten >=-1pt] (Vprof.south west) --  ([xshift=-0.1cm, yshift=0.1cm]Tabs.east);

\node[art=atwlKnowledge] (Kassess) at (4.2, -8.9)
  {\usebox{\boxIconKnowledge}\\[0pt]assessment};

\draw[tx] (Ppat) -- (Kassess)
  node[txl, midway, above=2pt] {\usebox{\boxIconAssess}};

\draw[tx, dashed, atwlSpecification!70!black, rounded corners=3pt]
  (Kassess.east) -- ++(0.8,0) |- (Sclust.east)
  node[txl, solid, pos=0.25, right=2pt]
    {\usebox{\boxIconGenerateKnowledge}};

\node[art=atwlKnowledge] (K1) at (1.5, -10.4)
  {\usebox{\boxIconKnowledge}\\[0pt]K1: insights};

\draw[tx] (Ppat.south) -- (K1.north)
  node[txl, midway, right=2pt] {\usebox{\boxIconGenerateKnowledge}};

\draw[tx, atwlTransform!50, rounded corners=2pt]
  (Kassess.south west) --  (K1.north east)
  node[txl, pos=0.4, right=2pt] {\tiny exit};

\end{scope}  

\begin{scope}[font=\scriptsize\sffamily, atwlTransform!70]
  \node[draw=atwlEntities, double, double distance=1pt, rounded corners=2pt,
        fill=atwlEntities!8, minimum width=0.7cm, minimum height=0.4cm, inner sep=0pt]
    (kgiven) at (-2.6, -11.2) {};
  \node[anchor=west] at ([xshift=4pt]kgiven.east) {given artifact};
  \node[draw=atwlTransform!50, thick, dashed, rounded corners=4pt,
        minimum width=0.9cm, minimum height=0.4cm, inner sep=0pt]
    (kloop) at (0.7, -11.2) {};
  \node[anchor=west] at ([xshift=4pt]kloop.east) {iterative loop};
  \draw[tx] (3.3, -11.2) -- ++(0.7, 0);
  \node[anchor=west] at (4.05, -11.2) {transform};
  \draw[tx, dashed, atwlSpecification!70!black] (-3.0, -11.85) -- ++(0.9, 0);
  \node[anchor=west] at (-2.0, -11.85) {feedback (parameter update)};
\end{scope}

\end{tikzpicture}
\caption{Visual representation of the cluster-calendar workflow
\cite{vanWijkSelow1999} using ATWL icons. Small icons on arrows indicate transform intents.}
\label{fig:cluster-calendar-icons}
\end{figure}

\subsection{Visual Representation of ATWL Workflows}
Although ATWL is a textual language, its structured artifact--transform dependencies enable translation into flow diagrams. Such diagrams complement textual representations by making the overall workflow topology immediately perceptible by a human.

We used two ways of creating the diagrams. First, we asked an LLM agent to generate a diagram from the ATWL specification of a workflow. We did this individually for each of the first 18 workflows of the library, with modest corrective feedback. Appendix~B{} shows the result for the Cluster-Calendar workflow~\cite{vanWijkSelow1999} (Fig.~\ref{fig:cluster-calendar}). A simplified visualisation of the Cluster-Calendar workflow with the use of the icons is shown in Fig.~\ref{fig:cluster-calendar-icons}. The diagrams of all 18 workflows are available in the Web supplement. Second, the library includes an interactive viewer that generates a node-link diagram automatically for any ATWL workflow loaded in it, without calling an LLM. It can be used for the other workflows of the library and for new workflows.

\section{Workflow Library}
\label{sec:library}

To evaluate the expressive power of ATWL, to study the structure of VA workflows across applications, and to support experiments on workflow design, we constructed a library of 54 workflows extracted from published VA papers. The library was built in two stages. The first 18 workflows were formalised while the language was designed and tested. We then extended the library to 54 workflows. The papers were published between 1999 and 2026. They deal with, for example, movement and spatio-temporal data, event sequences, text and topic modelling, networks, eye tracking, biomedical cohorts, simulation ensembles, and analysis of machine learning models. For each workflow, the library provides a summary, the ATWL representation, and graphical views. The Web supplement \url{https://geoanalytics.net/VAworkflows/} includes the library, a web application for viewing and analysing the library workflows by their structural features, and an ATWL syntax checker tool.

We looked for papers that describe a complete analytical workflow from data to knowledge. Papers that present a single technique with usage examples were not included. The additional papers were first recommended by the co-authors. Then we asked an LLM to estimate, from the abstracts of IEEE VIS publications (1990-2024), the probability that a paper describes a workflow. We skimmed the highest-ranked papers and selected those with sophisticated workflows that also increased the diversity of the library. Of the 54 papers, 26 have at least one co-author of this article among their authors. The co-authors have worked in different and mostly non-overlapping areas of VA, so their workflows alone cover a broad part of the field. We do not claim that the library is a representative sample of VA research.

The workflows were extracted by LLMs under human supervision, following the procedure tested on the first 18 workflows (Appendix~\ref{sec:library-construction}). An ``extractor'' LLM instance created the ATWL representation from the paper. An ``evaluator'' instance (called ``reviewer'' in the instructions and in the supplement) checked it against the language definition and the paper, and the extractor revised the representation according to the evaluator's feedback until no issues remained. Most workflows were extracted and evaluated by Claude Opus 4.6, which took both roles; GPT-5 and Claude Sonnet 5 were used only for a few of the most recently added workflows. We now recommend that the extractor and the evaluator are different models, that the syntax checker is run and its errors are corrected before the first evaluation, that evaluation and revision are repeated until the evaluator finds no issues, and that an author of the workflow or another person who knows it well checks the result. Human supervision remained necessary. In a few cases, the extractor was uncertain how to interpret parts of a paper and needed our feedback. Such uncertainty comes from ambiguities in the texts, so we do not expect that fully automatic extraction can be achieved. In the second stage, extraction of a workflow, including human supervision, took 15--20 minutes.

We checked the representations of the first 18 workflows. The checks were most detailed for the workflows that we know well, such as our own ones, Cluster-Calendar, Snapshots-to-Points, and Partition-based Regression. We did not check the other 36 workflows at the same level of detail. In addition, we asked the authors of the library papers to evaluate the representations of their workflows. The results are reported in Section~\ref{sec:expert_feedback}. To see how much a representation depends on the extraction, we extracted five workflows again with two current models (Section~\ref{sec:reextraction}). The results agree in the backbone and differ in detail, so a representation should be checked by someone who knows the workflow.

\section{Analysis of the Library Workflows} 
\label{sec:analysis}

The 54 workflows of the library are formal objects. This section illustrates how this permits comparative computational analysis of workflows. Here we take the representations as given. Since the library is not a representative sample, and the ATWL codes are not guaranteed to reproduce the papers faithfully (not all were checked by workflow authors), our analyses are exploratory; they show what becomes routinely possible once workflows are formal. All matrices, features, scripts, and further analyses are available in the Web supplement. In the ATWL Workflow Lens, readers can select workflows on an overview map of the library, compare groups of workflows (transitions, features, closed cycles), reproduce Fig.~\ref{fig:library}b interactively, and see for every feature value the transforms that realise it. Section~\ref{sec:reextraction} shows how much a representation changes when a workflow is extracted again.

\subsection{Transitions Between Analysis Steps}

\begin{figure*}[t]
\centering
\includegraphics[width=\textwidth]{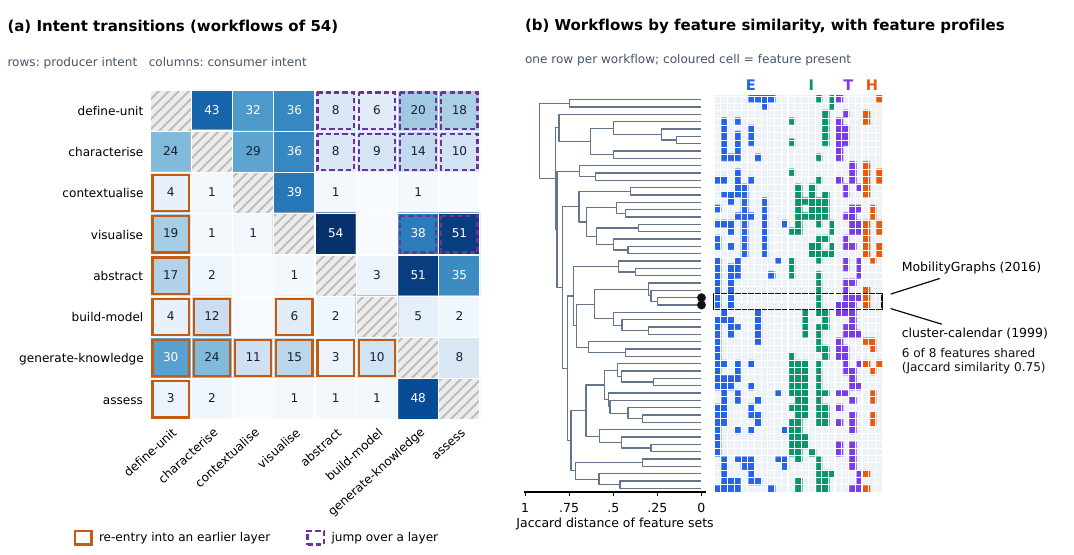}
\caption{Comparing the 54 library workflows computationally. (a)~Intent transition matrix: number of workflows in which an artifact produced by a transform of the row intent is used by a transform of the column intent (darker means more; steps between transforms of the same intent are merged and hatched). Frames mark re-entry into an earlier layer (orange; layers as in Fig.~\ref{fig:layer-cake}) and a jump over a layer (purple, dashed). (b)~Workflows ordered by the similarity of their 25 binary features (average-linkage clustering of Jaccard distances); a coloured cell means that the feature is present. Groups: E re-entries of specifications and results, I what loops re-define, T structure, H human activity. The framed workflows share 6 of 8 features although they differ in data type and are 17 years apart.}
\label{fig:library}
\end{figure*}

From each workflow we derive a graph whose nodes are transforms, labelled by their intent. An edge leads from one transform to another if an artifact produced by the first is an input of the second, either directly or through an \kw{assign} statement in a loop, which replaces an artifact by a new version for the next iteration. Consecutive transforms of the same intent are merged, which removes differences due to representation granularity. %
Counting the edges between intents over the library gives the transition matrix in Fig.~\ref{fig:library}a. Mapping intents to the layers of Fig.~\ref{fig:layer-cake} classifies every edge: it is a \emph{forward step} if it stays within a layer or leads to the next layer (the organisation layer being optional), a \emph{re-entry} if it leads to an earlier layer, and a \emph{jump} if it skips a layer upward. A jump is counted only if the consuming transform has no input from the skipped layer.

The matrix shows that nearly every workflow contains a step in which a view (\kw{visualise}) is interpreted (\kw{abstract}; 54 workflows) or judged (\kw{assess}; 51), and that an interpretation leads to \kw{generate-knowledge} (51). Together with a regular division of labour, in which humans interpret, judge, generate knowledge, and decide (through \kw{if} construct) while machines characterise, contextualise, visualise, and build models, this makes the backbone of the library. Only the definition of units (\kw{define-unit}) is mixed (human, hybrid, and machine in 19, 13, and 43 workflows). The workflows differ in where results and updated specifications re-enter earlier steps (Fig.~\ref{fig:structural-motifs}).

\subsection{Frequent Patterns}

To find recurring structure without prescribing it, we mined frequent connected subgraphs of up to four edges that occur in at least five workflows (Appendix~\ref{app:mining}). Nodes are labelled by intent and actor, and edges by the kind of artifact passed and by whether the edge is feedback. The 3\,655 patterns found were dominated by variants of one backbone, and the most frequent patterns are the least informative. We therefore ignored actors and feedback, applied explicit rules (Appendix~\ref{app:mining}), and selected the remaining candidates by hand. The procedure depends on typed artifacts and explicit input--output links, which prose does not provide.

Patterns of two edges added little, because they combine a re-entry into an earlier step with an expected forward step. Of 651 paths of three edges and 1\,440 of four edges, 26 satisfied the rules, and 25 of them extend a single re-entry by expected steps back to a view (the other leads from a view directly to knowledge or specification). The most frequent one, view $\rightarrow$ judgement $\rightarrow$ (re-)specification $\rightarrow$ redefinition of units $\rightarrow$ new view, occurs in 14 workflows. The sequence from view through interpretation to decision on specification occurs in 51 workflows.

\begin{figure}[ht]
\centering
\providecolor{atwlModel}{HTML}{8E7CC3}
\providecommand{\motifModelIcon}{\ifdefined\boxIconModel\usebox{\boxIconModel}\else{\tiny\sffamily model}\fi}
\begin{tikzpicture}[
  >=Stealth,
  art/.style={
    draw=#1, semithick,
    fill=#1!8,
    rounded corners=2pt,
    minimum width=0.85cm,
    minimum height=0.5cm,
    align=center,
    font=\tiny\sffamily,
    inner sep=1.5pt,
  },
  tx/.style={
    ->, semithick, atwlTransform,
    shorten >=1.5pt, shorten <=1.5pt,
  },
  txl/.style={
    font=\tiny\sffamily,
    atwlTransform,
    inner sep=1pt,
    fill=white, fill opacity=0.9, text opacity=1,
    rounded corners=1pt,
  },
  cnt/.style={
    font=\tiny\sffamily\bfseries,
    inner sep=0.8pt,
    fill=white, fill opacity=0.92, text opacity=1,
    rounded corners=1pt,
  },
  reent/.style={
    ->, dashed, atwlSpecification!70!black,
    shorten >=1pt, shorten <=1pt,
  },
  viewarc/.style={
    ->, atwlVisualisation!75!black,
    shorten >=1pt, shorten <=1pt,
  },
  panel/.style={
    font=\scriptsize\sffamily\bfseries,
    anchor=north west,
  },
  panelframe/.style={
    draw=black!15, thin, rounded corners=3pt,
    fill=black!1,
    inner sep=4pt,
  },
  dec/.style={
    font=\tiny\sffamily, atwlTransform,
    draw=atwlTransform, thick,
    diamond, aspect=1.8,
    minimum width=0.6cm, minimum height=0.4cm,
    inner sep=1pt, fill=white,
  },
]

\begin{scope}[shift={(0.15,0)}]
  \node[art=atwlEntities] (a-ent) at (0.45, -1.55)
    {\usebox{\boxIconEntities}};
  \node[art=atwlFeature] (a-fea) at (1.75, -1.55)
    {\usebox{\boxIconFeature}};
  \node[art=atwlArrangement, dashed, opacity=0.75] (a-arr) at (3.05, -1.55)
    {\usebox{\boxIconArrangement}};
  \node[art=atwlVisualisation] (a-vis) at (4.35, -1.55)
    {\usebox{\boxIconVisualisation}};
  \node[art=atwlPattern, dashed, opacity=0.75] (a-pat) at (5.65, -1.55)
    {\usebox{\boxIconPattern}};
  \node[art=atwlKnowledge] (a-kno) at (6.95, -1.55)
    {\scalebox{0.8}{\usebox{\boxIconKnowledge}}};

  \node[dec] (a-dec) at (6.95, -2.6) {?};
  \node[font=\tiny\sffamily\itshape, atwlTransform!80, inner sep=1pt]
    (a-exit) at (7.6, -2.6) {exit};

  \node[art=atwlSpecification] (a-spe) at (4.35, -4.05)
    {\usebox{\boxIconSpecification}};
  \node[art=atwlModel] (a-mod) at (0.45, -4.05) {\motifModelIcon};

  \draw[tx] (a-ent) -- (a-fea)
    node[txl, midway, above=1pt] {\scalebox{0.8}{\usebox{\boxIconCharacterise}}};
  \draw[tx] (a-fea) -- (a-arr)
    node[txl, midway, above=1pt] {\scalebox{0.8}{\usebox{\boxIconContextualise}}};
  \draw[tx] (a-arr) -- (a-vis)
    node[txl, midway, above=1pt] {\scalebox{0.8}{\usebox{\boxIconVisualise}}};
  \draw[tx, dashed, opacity=0.5] (a-vis) -- (a-pat)
    node[txl, midway, above=1pt, opacity=0.4] {\scalebox{0.6}{\usebox{\boxIconAbstract}}};
  \draw[tx] (a-pat) -- (a-kno)
    node[txl, midway, above=1pt] {\scalebox{0.8}{\usebox{\boxIconAssess}}};
  \draw[tx, shorten >=0pt] (a-kno) -- (a-dec);
  \draw[tx, atwlTransform!80, shorten >=0pt] (a-dec) -- (a-exit)
    node[txl, pos=0.3, above=1pt] {ok};

  \draw[tx, rounded corners=2pt] (a-dec.south) |- (a-spe.east)
    node[txl, pos=0.75, above=0pt]
      {\scalebox{0.8}{\usebox{\boxIconGenerateKnowledge}}};

  \draw[reent, line width=1.7pt] (a-spe.north west) to[out=140, in=-75]
    node[cnt, pos=0.62] {30} (a-ent.south east);
  \draw[reent, line width=1.45pt] (a-spe.north west) to[out=125, in=-80]
    node[cnt, pos=0.62] {24} (a-fea.south);
  \draw[reent, line width=0.9pt] (a-spe.north) to[out=105, in=-85]
    node[cnt, pos=0.62] {11} (a-arr.south);
  \draw[reent, line width=1.1pt] (a-spe.north) -- 
    node[cnt, pos=0.5] {15} (a-vis.south);
  \draw[reent, line width=0.9pt] (a-spe.west) --
    node[cnt, midway] {10} (a-mod.east);

  \draw[viewarc, line width=1.45pt] (a-vis.north) to[bend right=24]
    node[cnt, pos=0.5] {\scalebox{0.7}{\usebox{\boxIconDefineUnit}}\,23} (a-ent.north);
  \draw[viewarc, line width=1.25pt] (a-vis.south east) to[out=-25, in=180]
    node[cnt, pos=0.5] {19} (a-dec.west);

  \draw[atwlTransform!60, thick, dashed, rounded corners=5pt]
    (-0.2, -0.55) rectangle (7.85, -4.6);
  \node[font=\small\sffamily\itshape, atwlTransform!80, anchor=south east]
    at (7.85, -4.6) {\scalebox{1.4}{$\circlearrowleft$}};

  \begin{scope}[font=\tiny\sffamily, atwlTransform!80]
    \draw[reent, line width=0.8pt] (-0.10,-4.95) -- ++(0.50,0);
    \draw[reent, line width=1.7pt] (0.55,-4.95) -- ++(0.50,0);
    \node[anchor=west] at (1.15,-4.95) {a decision re-specifies the artifact};
    \draw[viewarc, line width=1.4pt] (4.95,-4.95) -- ++(0.50,0);
    \node[anchor=west] at (5.55,-4.95) {taken from a view};
    \node[cnt, draw=atwlTransform!60, thin, circle, inner sep=1.2pt] at (0.05,-5.30) {$n$};
    \node[anchor=west] at (0.30,-5.30) {number of workflows (of 54); the wider the line, the larger $n$};
  \end{scope}
\end{scope}

\end{tikzpicture}

\caption{The assess--refine cycle in the library (50/54 workflows): a view is interpreted, the result is judged, and a decision either ends the loop or re-specifies an earlier artifact. Dashed arcs show what a decision re-specifies (units, features, arrangement, view, model); solid arcs show decisions and units taken straight from a view. Numbers are workflows (of 54). Dashed semi-transparent nodes indicate typically arising optional artifacts.}
\label{fig:structural-motifs}
\end{figure}

What longer patterns show is closure. We call a pattern a \emph{closed cycle} if its steps are distinct, each step passes an artifact to the next, and the last passes an artifact back to the first. Since the dependency graph of a workflow is acyclic except for \kw{assign} statements, a closed cycle normally marks an iteration in which a result is fed back, through a loop, to an earlier step. Fig.~\ref{fig:structural-motifs} shows this cycle in a general form. It generalises the assess--refine loop, a loop that contains a view, a human assessment, and an explicit exit, which 50 of the 54 workflows have. Counting cycles of three to six steps, 51 workflows contain a closed cycle, 39 of them through a decision, a view, and an interpretation. A re-entry almost always lies on a closed cycle: a decision re-specifying units in 29 of the 30 workflows with it, features in 22 of 24, a view in 15 of 15, the arrangement in 11 of 11, and the model in 8 of 10. The most common cycle leads from a view through a judgement and a decision to a redefinition of units and back to the view (14 workflows); variants return through the arrangement (12) or the features (10).

\subsection{Features for Comparing Workflows}

Patterns that pass the selection become features, together with properties computed from other parts of the representation. A feature is a binary property of a workflow, defined by a rule on its graph or its fields, so that its value can be recomputed for any workflow written in ATWL; Table~\ref{tab:features} (Appendix \ref{app:mining}) gives the rules; no feature uses free text. Expected forward steps are reported as findings and not used as features, because they do not distinguish workflows. We use 25 selected features in four groups.

\emph{Re-entry} features (11) record whether a decision re-specifies units, features, visualisation, arrangement, or model, or guides what is abstracted; whether a judgement redefines the working set; whether model output serves as features or defines units; whether units are defined from a view, its interpretation, or the arrangement it is based on; and whether a decision is taken straight from a view.

\emph{Iteration} features (7) record which kind of artifact an \kw{assign} statement inside a loop updates: units, features, arrangement, model, specification, patterns, or knowledge.

\emph{Structure} features (4) refer to involving two or more given datasets, defining group/cluster units, using several views together, and having two or more sequential loops.

\emph{Human activity} features (3) record whether a workflow contains a hybrid transform, whether human configures the analysis a priori, before any view, pattern, or assessment, and whether explicit initial knowledge enters.

Because the features abstract from data and domain, workflows can be compared across domains. Fig.~\ref{fig:library}b orders the workflows by the Jaccard distance of their feature sets and shows their profiles. The most similar pair that differs in data type, cluster-calendar~\cite{vanWijkSelow1999} (time series, 1999) and MobilityGraphs~\cite{MobilityGraphs2016} (trajectories, 2016), shares six of eight features: in both, the units are clusters, a specification is revised in a loop, several views are interpreted together, a hybrid transform takes part, and decisions re-specify the units and are taken directly from views. They differ in that cluster-calendar has two given datasets and MobilityGraphs has sequential loops. The papers address different data and are 17 years apart, but the representation makes their structural similarity explicit and checkable. %

The same representation lets us test claims about the library: do the features merely reflect the data type or the authors? Among 225 comparisons of features with data and authorship attributes (modality, unit types, multi-source data, co-author involvement), 20 were significant at $p<0.05$, about what chance alone gives (11); only one survived correction for multiple comparisons, and it follows from the definitions (``two or more given datasets'' and integration of sources). Even the strongest data attributes explain only a small part of the differences between feature profiles: about 5\% for sequence data, and 5\% for multi-source workflows once the feature defined from the given data is excluded (8\% with it). Papers with a co-author of this paper do not differ from the others ($p = 0.12$, permutation test). Each analysis is a short program on the ATWL representations; it can be rerun as the library grows and extended by adding a feature.

\subsection{Sensitivity to Re-Extraction}
\label{sec:reextraction}

A paper can be encoded in several ways, but the library holds one encoding per paper. To see how much this matters, we extracted five workflows again (UTOPIAN, DPVis, EventThread, NodeTrix, and SOMFlow) with Claude Sonnet 5 and GPT-5.6-Terra. Each model extracted a workflow, the other model reviewed it, and the first model revised it once (Appendix~\ref{app:reextraction}).

\textbf{The backbone is reproduced.} All 15 versions contain a closed cycle, and 12 of them a cycle through a decision, a view, and an interpretation, similar to the library (39 of 54).

\textbf{Detail is not.} The versions of a paper differ in the number of transforms and in the intent and typing of single steps, even after consecutive transforms of the same intent are merged (Appendix~\ref{app:reextraction}). As Table~\ref{tab:reextraction} shows, a re-extraction agrees with the library version in 16--17 of the 25 features, hardly more than two library workflows agree with each other. The library version of a paper is never the most similar library workflow of its re-extraction; its rank among the 54 is between 4 and 41 (median 14).

\begin{table}[t]
\setlength{\belowcaptionskip}{0pt}
\caption{Agreement between encodings of the same workflow (means of five papers) and between two library workflows (all 1\,431 pairs).}
\label{tab:reextraction}
\footnotesize
\renewcommand{\arraystretch}{1.1}
\begin{tabular}{@{}l rr@{}}
\toprule
 & Features agreeing (of 25) & Jaccard similarity \\
\midrule
Library version, Sonnet & 17.0 & 0.28 \\
Library version, Terra & 16.0 & 0.30 \\
Sonnet, Terra & 18.0 & 0.43 \\
Two library workflows & 15.6 & 0.23 \\
\bottomrule
\end{tabular}
\end{table}

\textbf{The library statistics are stable.} Replacing the five library versions by the re-extractions of either model changes the count of any feature by at most four of the 54 workflows, and the frequencies of the features correlate with the original ones ($r=0.98$). This bounds the effect of this amount of noise, not of a systematic difference between models: Terra produced more features per workflow (10.2) than Sonnet (7.6) and the library versions of the five papers (6.8).

\textbf{Reading.} Different encodings of a paper reflect choices that the instructions leave to the extractor, such as which part of a tool paper to represent, how finely to decompose a step, or how to type an artifact. ATWL makes these choices explicit, but it does not make them unique. We therefore use the features for statements about the library and not for the similarity of single workflows, and we recommend that the authors of a workflow, or people who know it well, check its representation.

\section{Workflow Recommendation Experiments}
\label{sec:recommendation}

The library can supply sources of ideas and methods for designing analytical workflows, and ATWL can be the language in which a design is stated, so that its structure becomes explicit and checkable. We used an LLM as a workflow designer to see what each contributes, alone and together. We expected that (1)~with the library, a design names its sources and what it takes from them, whereas without it the design rests on the model's general knowledge and cannot be traced; (2)~a design in ATWL commits to who acts, where human judgement is needed, what is iterated, and when iteration ends, which prose can leave open; (3)~a library in ATWL form transfers structure but is abstract about methods, a library in prose form carries method detail, and the source papers can add it afterwards. We measure what the designs commit to, not whether they work; this would require implementing them.

\subsection{Design of the Experiments}

\textbf{Tasks, models, conditions.} Two tasks were each given to two LLMs, Claude Sonnet 5 and GPT-5.6-Terra, which makes four blocks; every design is a single run. VAST asks how a public-health team can find where a flu-like outbreak started and how it is transmitted, from geolocated microblogs, a map, population statistics, and weather data. VIS asks how to identify communities of collaborating researchers, their stable cores and leaders, research foci, overlaps, and merges and splits, in the IEEE VIS publication data (1990--2024). Notably, the library does not include any paper on detection and analysis of communities in networks. Table~\ref{tab:rec-conditions} lists the conditions. C1, C2, C4, and C3 form a 2$\times$2 of the ATWL specification (with one example) and the library, which had 53 workflows at the time; it was supplied in ATWL form (C3) or as prose summaries (C4). By prose we mean natural-language descriptions of workflows. In the experiments, the prose library consists of the workflow summaries of the library pages, which the extractor LLM wrote; manuals and system references are not included. The prompt in conditions C3 and C4 asked to cite the library workflows used. The other conditions continue a chat. Prompts, designs, and logs are in the Web supplement.

\begin{table}[t]
\setlength{\belowcaptionskip}{0pt}
\caption{Conditions. The model receives the task and, in addition:}
\label{tab:rec-conditions}
\footnotesize
\renewcommand{\arraystretch}{1.05}
\begin{tabularx}{\columnwidth}{@{}l >{\raggedright\arraybackslash}X@{}}
\toprule
C1 & nothing; prose design \\
C2 & ATWL specification and one worked example; prose and ATWL design \\
C3 & C2, and the library in ATWL form \\
C4 & the library as prose summaries; prose design \\
C5 & C3, then the source papers of the cited workflows: method-level detail, structure fixed \\
C6 & C4, then the ATWL specification and example: faithful translation (C6), what ATWL showed that could be improved (C6+), implementation of that (C6++) \\
C7 & C4, then the ATWL specification and example: fresh design in ATWL \\
\bottomrule
\end{tabularx}
\end{table}

\textbf{Measures.} The designs differ in form, so we use three instruments. For \emph{library use}, we count the library workflows that a design names. The 20 \emph{ATWL designs} (C2, C3, C6, C6++, C7) are measured with the 25 features of Section~\ref{sec:analysis}, the ATWL syntax checker, and size. The eight \emph{prose designs} (C1, C4) were coded by two LLM coders under a written protocol: does a step state who acts, what the human controls, what is iterated, which parameters are named, which evidence, criteria, and alternatives are given, which outputs; every code carries a verbatim quote enabling human check. Each design was processed in a separate session, so that every LLM instance saw one design only and nothing about its condition. The coders agree on 86--95\% of the steps on the presence of these elements. %
The coded elements are mapped to ATWL elements; the mapping is approximate except for loops (Appendix~\ref{app:recommendation}). %
A paired comparison of two conditions cannot reach $p<0.05$ with four blocks, so the results are descriptive.

\subsection{What the Library Contributes}

\textbf{Sources.} None of the eight designs without the library (C1, C2) names a library workflow. The designs with it name 4--12 (median 6) and say what they take from each (Table~\ref{tab:rec-results}). The result shows what the library adds: traceable examples from which an analyst can judge a recommendation.

\textbf{Form of the library.} In VAST scenarios, the designs with the ATWL library mostly take structural patterns, such as loops that refine extraction criteria; those with the prose library take methods, such as density-based clustering for place extraction. In VIS, the difference is not visible. Since library has no paper on community detection, the designs take such methods from general LLM's knowledge. Method detail comes mainly from the papers: after C3, using the source papers (C5) added parameter ranges, validation checks, and pointers to sections, figures, and equations for 10--27 methods per design, against 7--14 in C4, with the structure unchanged.

\textbf{Structure.} The prose library (C4 against C1) makes designs state more loops (1.4 against 0.75 per design), more annotation and logging acts (2.8 against 1.1), and provenance statements (none in C1), and name outputs in more steps (0.61 against 0.36). The direction is the same for both coders. The human-machine division of labour and the number of named parameters do not change. The ATWL library (C3 against C2) makes designs more similar to the library (Jaccard similarity of the features to the most similar library workflow 0.62 against 0.48), with more features (9.75 against 7.5 of 25), loops (2.75 against 1.75), and human judgement transforms (+4.0, in all four blocks). The similarity values are descriptive, since two encodings of the same workflow agree only moderately in their features (Section~\ref{sec:reextraction}).

\subsection{What ATWL Contributes}

\textbf{Explicit commitments.} ATWL requires what prose leaves open. All 55 loops in the ATWL designs declare an exit condition, against none of the 17 loops coded in the prose designs. Every transform names its actor, inputs, and output, whereas prose designs state an actor in 0.88--1.00 of the steps and an output in 0.36--0.61. ATWL designs contain more loops (1.75--4.25 against 0.75--1.4 per design), and the translation of a C4 design (C6) more than the design itself (2.25 against 1.4). Asked what ATWL had shown, all four models named implicit criteria that should become specifications and single-pass steps that should become loops (the prompt named such gaps as examples).

\textbf{Branching.} The translation from prose to ATWL loses conditional paths beyond loops. The C4 designs state 4.1 conditional paths per design. Most of them are not loop exits. They are conditions on data availability, rules that map an observed pattern to a hypothesis, and triggers of a step. The C6 designs have none outside loops. Translation is not the only reason, however. ATWL has conditionals, but the 20 designs contain 53 of them inside loops, mostly as exit gates, and only 2 outside. The example workflow given to the LLMs uses conditionals only in loops. No conditionals beyond loops are in the eight designs made directly in ATWL (C2, C3). The C6 prompt asked not to add structure that the design does not support, and one model said that it left out a fallback because the prose offered it only as an option. These runs cannot show whether the language or its presentation is the cause. One possible reason is that ATWL was developed from published workflows, whose descriptions are often not explicit about alternative branches. In either case, ATWL has no established way to express alternative workflow paths, which is a limitation of the current language and examples.

\textbf{Validity and methods.} ATWL designs derived from a prose design (C6, C6++, C7) have more checker errors than designs made with the ATWL materials alone (C2, C3): 7 of 12 against 0 of 8, including all six Terra designs. ATWL classifies transforms by intent, not by method, so the methods come from the papers (C5).

\textbf{Model and task.} On the 25 features, the model explains 15\% of the variance, the task 13\%, and the condition 15\% ($p=0.94$). The features do not separate the conditions clearly, and they do not measure quality.

\begin{table}[t]
\setlength{\belowcaptionskip}{0pt}
\caption{Main comparisons: the 2$\times$2 of ATWL and library (means over four blocks; prose designs: mean of the two coders; $^\ast$loop records of both coders). Human judgements in prose designs are not shown due to disagreements between the coders.}
\label{tab:rec-results}
\footnotesize
\renewcommand{\arraystretch}{1.05}
\begin{tabularx}{\columnwidth}{@{}>{\raggedright\arraybackslash}X r r r r@{}}
\toprule
 & C1 & C2 & C4 & C3 \\
 & task & ATWL & library & ATWL+lib. \\
\midrule
Library workflows named & 0 & 0 & 8.8 & 5.5 \\
Loops per design & 0.75 & 1.75 & 1.4 & 2.75 \\
Loops with a stated exit$^\ast$ & 0 of 6 & all & 0 of 11 & all \\
Steps naming an output & 0.36 & 1 & 0.61 & 1 \\
Human judgement transforms & -- & 5.8 & -- & 9.8 \\
Designs with checker errors & -- & 0 of 4 & -- & 0 of 4 \\
\bottomrule
\end{tabularx}
\end{table}

\subsection{Using the Library and ATWL Together}
\label{sec:recommendation:value}

ATWL and the library each add loops, %
and C3 has the most. The results suggest a sequence in which each element does what it is good at: a design in ATWL with the ATWL library gives a structure with sources (C3); the source papers then add methods, parameters, and pointers for implementation (C5); a prose design can be translated into ATWL to expose its implicit structure, provided the result is checked against the prose, because conditional paths are lost (C6). These are observations from single runs of two models on two tasks that the library covers to different degrees. We did not implement any design, so we cannot say which is better; the experiments show what the library and the ATWL representation contribute to a design, and what they require.

\begin{figure*}[ht]
    \centering
    \includegraphics[width=0.8\linewidth]{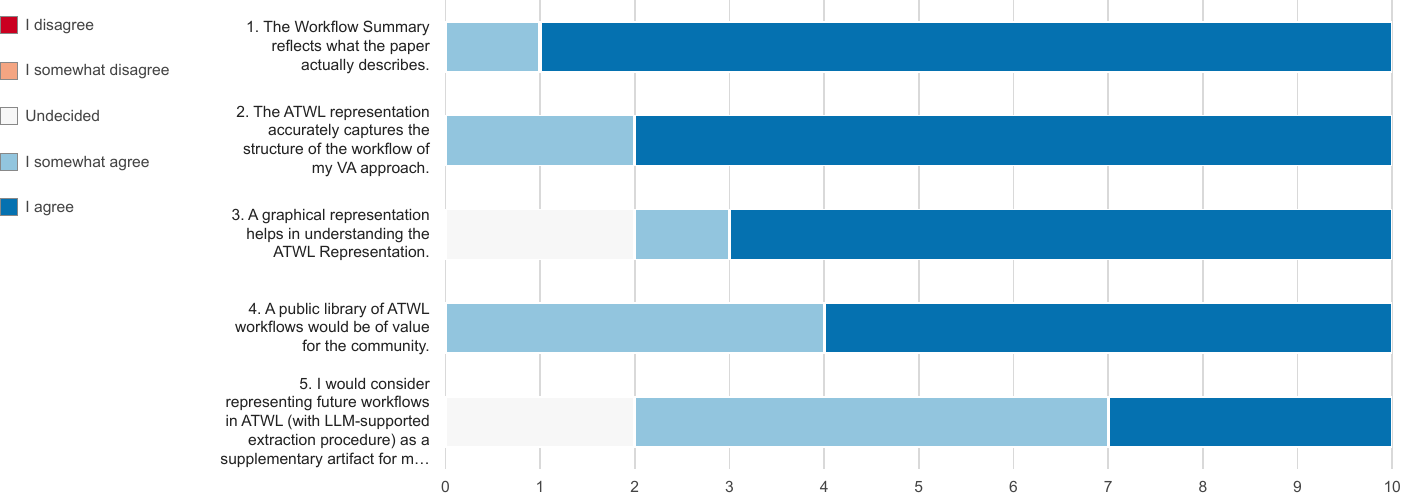}
    \caption{Quantitative results of the questionnaire based on 5-point Likert scale ratings.}\vspace{1ex}
    \label{fig:questionnaire}
\end{figure*}

\section{Expert Feedback}
\label{sec:expert_feedback}
We contacted the authors of VA publications (none was involved in ATWL development) in our workflow library to assess the correctness of generated workflows and to collect feedback on ATWL. 
The feedback was collected with a questionnaire addressing (1) the textual workflow summary and (2) the ATWL representation, both as code and with a graphical representation. The 14 questions comprised 5-point Likert-scale items, 8 free-text items, and 1 multiple-choice item. We sent the questionnaire along with tutorial slides, a link to the author's workflow, and a link to the workflow library for comparison with other workflows.
We received feedback from 10 visualization experts who investigated their ATWL workflow. Figure \ref{fig:questionnaire} depicts the results of the quantitative assessment with respect to correctness, usefulness, and further usage.

\subsection{Textual Workflow Summary}
The majority of participants agreed that the textual Workflow Summary reflects what the paper actually describes ($\text{m}=4.90$, $\text{sd}=0.30$).

All participants agreed that no parts were falsely added to their workflow. One participant mentioned that \textit{ ``... the high level description of the process does not allow the description of fine details''}, and added that it is also not necessary for this level of abstraction.

When asked for any missing parts, participants mentioned some aspects that were underrepresented: (1) Representations that \textit{``... were not part of the core workflow...''} but are mentioned \textit{``... as a possibility to show the flexibility of the approach.''}. (2) Data types and visualization types that are only mentioned once or twice in the paper were omitted. (3) One participant mentioned a missing feedback link between analysis phases. 
Overall, there were just minor gaps, often due to lower priority in the original paper (e.g., mentioned as an option), fundamental parts were not missing.

\subsection{ATWL Representation (Code/Graph)}
Overall, the experts agreed that the ATWL representation accurately captures the structure of the workflow of their VA approach ($\text{m}=4.80$, $\text{sd}=0.40$).
A graphical representation was considered helpful for understanding the ATWL representation by the majority ($\text{m}=4.50$, $\text{sd}=0.81$).

Analogously to the textual representation, we asked for interpretation errors, missing parts, and falsely added parts in the ATWL representations.

Some interpretation issues were identified. Specifically, one expert mentioned misinterpretation of the intent of the final workflow step (\kw{generate-knowledge} vs. \kw{build-model}).  %
In another workflow, refinement loop stated an exit condition that was actually an interactive filtering step, and an initial artifact state was used as transform input instead of updated one.
One expert mentioned that ATWL introduced some redundant steps, and some artifact names were deemed uninformative. A missing link was mentioned by one expert.
It was further mentioned that the graphical representation \textit{``... may help in having a general view of the process and identify the macro steps...''} while the code version \textit{``... for a trained eye, may provide a fine set of details...'' }

When looking for missing and falsely added parts, the level of granularity was mentioned. The high-level abstraction is correct, but individual parts need refinement. One participant mentioned that ``... the graph is massive and it is almost impossible to verify all edges.'' One expert stated that the step \textit{''Assess sufficiency of current insight''} is \textit{``...very implicit and something the user might not even be aware of. Different exit conditions are imaginable, like no more time, or conclusion that data does not provide anything insightful.''}
The experts further mentioned some potential improvements by aggregating/filtering large graphs, clearer marking of human decisions, annotations, and capturing user-driven visual changes and provenance.

\subsection{Future Use of ATWL}
\begin{figure*}[ht]
    \centering
    \includegraphics[width=\linewidth]{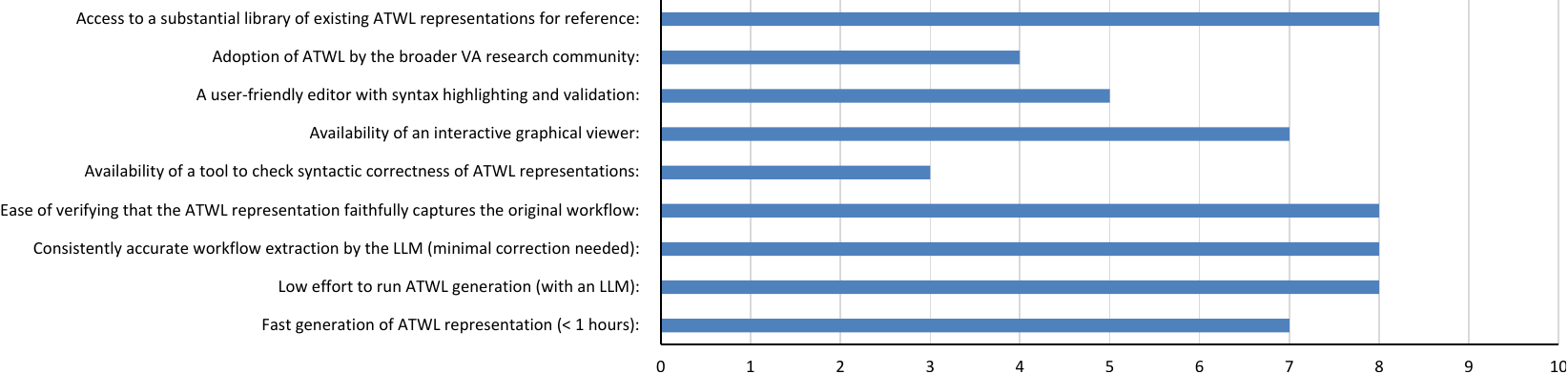}
    \caption{Conditions for the use of ATWL workflows}
    \label{fig:questionnaire_conditions}
\end{figure*}

We found strong agreement that a public library of ATWL workflows would be valuable to the community ($\text{m}=4.60$, $\text{sd}=0.49$). And experts would consider representing future workflows in ATWL ($\text{m}=4.10$, $\text{sd}=0.70$).
We further asked: "Which conditions must apply to you in order to use ATWL workflows to represent your VA approach?" Figure \ref{fig:questionnaire_conditions} depicts the count of conditions marked by the interviewees.
Interestingly, only three participants considered the availability of a tool for syntactic correctness a necessary prerequisite. This could be an indicator that the general trust in the resulting code structure is given. 
For a majority of participants, the higher priorities were the ease of verifying faithful representation of the workflow, consistency, and fast, low-effort generation of ATWL representations. Further, the availability of a library of ATWL workflows was deemed equally often as important.

For the future use of ATWL, the experts mentioned some potential improvements:

 \paragraph{Learning Curve} The experts stated that the learning curve might be a reason that the users may stay a \textit{``specific group of users within the VA community''}. Providing the introduction slides directly on the website, and a more general, domain-independent representation might facilitate the adoption of ATWL. 

 \paragraph{Complexity}
 \textit{``A usable ATWL needs to be digestible...''}, hence some mechanisms should be included to investigate workflows on different levels of abstraction.

 \paragraph{Visual Clarity}
 Experts mentioned that elements such as node names, human/machine components, layout stability, and color-coding could further improve the use of ATWL.

 \paragraph{Integration}
 Integrating ATWL into common code editors (e.g., VS Code, IntelliJ) with cross-references and natural-language summaries for a non-expert audience would further help adapt ATWL to existing research workflows.

\section{Discussion}
\label{sec:discussion}

We discuss what the evidence shows, the intended use of ATWL and its practical value, the limitations, and directions for further work.

\subsection{What the Evidence Shows}
\label{sec:discussion:contribution}

ATWL and the workflow library make visual analytics workflows comparable objects. Section~\ref{sec:analysis} compares 54 workflows by transitions between intents, closed cycles, and 25 features, and every result can be traced to the transforms that produce it. The language rests on four choices: typing artifacts by analytical role and not by computational format, classifying transforms by intent and not by method, treating human knowledge and specifications as artifacts, and restricting the language to observable, externalised products. Other choices would give a different language with different affordances.

The evidence that the language is adequate has three parts. The language did not change for the 36 workflows added after its design had stabilised. Ten authors of library papers rated the representation of their own workflow as sufficiently accurate, and the issues they raised were few and mostly minor (Section~\ref{sec:expert_feedback}). The same team selected and formalised the workflows, so this is necessary but not sufficient evidence of generality.

The experiments (Section~\ref{sec:recommendation}) show what the library and the representation add when an LLM designs a workflow. Designs made with the library refer to library workflows, and designs made without it never do. ATWL designs declare an exit for every loop and name the output of every transform, whereas none of the 17 loops in the prose designs states an exit. These are observations from single runs of two models on two tasks. We did not measure design quality and did not implement any design.

\subsection{Scope of the Language}
\label{sec:discussion:scope}

ATWL has a specific purpose: the representation of analytical workflows as they are reported or designed, so that they can be compared, reused, and checked. Three decisions follow from it.

\textbf{Designs, not executions.} ATWL describes the analytical logic of a workflow and not what happened in a particular session. Improvisation, errors and their repair, and differences between analysts appear only where a design anticipates them as loops or conditionals. Observed sessions can be related to a workflow, for example by mapping logged actions to its transforms; analytic provenance frameworks address this need (Section~\ref{sec:related_work}), and an ATWL workflow can be mapped to them.

\textbf{Abstraction from data and methods.} ATWL abstracts from the amount of data, the cost of a transform, and the scalability of a method, and it names methods only in the free-text manner. Workflows for a few hundred records and for billions of records can therefore have the same representation. This abstraction is what makes workflows from different applications comparable.

\textbf{Structure and free text.} A workflow in ATWL consists of a typed structure (artifact types, intents, actors, inputs and outputs, loops, conditionals) and of free text. Programs check and compare the structure, and all comparisons in this paper use it. Free text appears in the manner and description of transforms, in representation forms, in loop and branch conditions, and in the layout, form, and encoding of visualisations. It records what the structure deliberately leaves open, such as the method behind a transform. A clustering step and a trajectory-segmentation step are both \kw{define-unit}: the intent states what the step achieves for the analysis, and the manner states how. The syntax checker verifies structure and not meaning; the meaning is checked by the evaluator and by the authors of the workflow.

\subsection{Intended Use and Practical Value}
\label{sec:pract_value}

ATWL can be used for documentation, for design support, and for implementation, at different levels of readiness.

\textbf{Documentation and study.}
For researchers who publish VA workflows, ATWL is a documentation format that makes explicit what prose descriptions often leave implicit: the sequence of operations, the information produced and consumed, the points of human judgement, and the structure of iteration. Extracting the library workflows repeatedly exposed ambiguities in the original descriptions. The experts agreed that a public library of ATWL workflows would be valuable to the community ($\text{m}=4.60$ of 5), and they would consider representing future workflows in ATWL ($\text{m}=4.10$). The library supports this use now.

\textbf{Design support.}
An LLM can use the library when it designs a workflow for a new problem. In our experiments, the library made designs cite existing workflows, and ATWL made loops, exits, and outputs explicit. A prose design can be translated into ATWL to expose its implicit structure, provided the result is checked against the prose. The source papers remain the source of method detail (Section~\ref{sec:recommendation:value}).

\textbf{Implementation.}
We expect that problems requiring human--machine cooperation will increasingly be solved with tailor-made software that is developed iteratively with the help of an LLM, instead of existing tools. A workflow design would then be translated into a working system, and both would be updated according to feedback from execution. ATWL states what such a translation needs and prose often omits: who performs each step, which artifacts are exchanged, and when a loop ends. We have not tested this scenario. Method details and conditional paths would have to be provided in addition to the ATWL design.

\textbf{Adoption barriers.}
The experts expected the learning curve to keep ATWL with a specialised group within the community. They suggested providing the introductory slides on the website and views at different levels of abstraction. Extraction by an LLM strongly decreases the human effort but does not remove the need for human supervision. Tool support consists of the syntax checker and the interactive viewers, and there is no editor integration, which the experts asked for. The usefulness of the library grows with its size, which creates a bootstrapping problem for early adopters. Stored session logs, workflow summaries, and the modular representation (Section~\ref{sec:discussion:adoption}) lower these barriers.

\subsection{Limitations}
\label{sec:discussion:limitations}

\textbf{Sample and selection.}
The library has 54 workflows chosen for diversity and not at random, and 26 of the papers have a co-author of this article among their authors. All come from published VA research, so some regularities may reflect academic conventions and not properties of analytical reasoning. The findings may not transfer to industrial practice or to paradigms that the library does not cover. The cross-workflow analysis was done by the team that designed the language, so we cannot exclude that our perspective influenced both the formalisation and the regularities we looked for. Workflows formalised independently would be a stronger test.

\textbf{Expressiveness.}
The ontology was developed with a few workflows and then applied to the others, but the same team designed the language and selected and formalised the workflows. Workflows that need constructs outside the ontology, such as collaborative multi-analyst processes, real-time streaming, or extensively automated decisions, may exist but did not appear in the library. Our experiments show a gap in the expression of branching. ATWL has conditionals, but the 20 ATWL designs use them almost only as loop exits. The conditional paths of the prose designs were not carried over, also in the designs made directly in ATWL. One possible reason is that ATWL was developed from published workflows, whose descriptions are often not explicit about alternative paths. Workflows that are only partly defined when they start are not covered either (Section~\ref{sec:discussion:adoption}).

\textbf{Extraction and verification.}
LLMs extracted the workflows under human supervision. We checked the first 18 workflows, the authors of ten library papers checked their own workflow, and the other workflows were checked less. The authors reported few deviations (Section~\ref{sec:expert_feedback}). Details that a paper does not state cannot be recovered from its text. Formalisation also requires interpretation: where a workflow starts and ends, which intent a step has, and when a loop terminates. We extracted five of the workflows again with two current models. The backbone of the workflows was reproduced in all versions, but the transforms, the features, and the most similar library workflows varied (Section~\ref{sec:reextraction}). A workflow therefore has several defensible encodings, and its authors, or people who know it well, should check the representation. We did not compare representations written independently by several people. Writing ATWL by hand is laborious, and ATWL is meant to be produced with LLM support and checked by the authors of the workflow. The relevant question is therefore how much an LLM-assisted representation varies and whether it can be checked, which Section~\ref{sec:reextraction} and Section~\ref{sec:expert_feedback} address.

\textbf{Expert feedback.}
All ten experts assessed a paper they (co-)authored. This strengthens our confidence in the findings on fidelity, but it limits what can be said about usability and learnability for readers who do not know the workflow. No independent users have produced ATWL representations.

\textbf{Experiments.}
The experiments have four blocks (two tasks and two models) with one run per condition, so we report descriptive comparisons. We did not measure design quality. In another study \cite{softsky} we made an attempt of implementing an LLM-designed workflow and found that having the library and ATWL representation is insufficient; additional knowledge injection was necessary. We argued that the VA research field needs a typology of human knowledge injection~\cite{xing2026understandinghumansinjectknowledge} for implementing workflow designs.

\textbf{LLM dependence.}
We extracted most workflows with Claude Opus 4.6, which also evaluated them, and a few with GPT-5 and Claude Sonnet 5, and ran the experiments with Claude Sonnet 5 and GPT-5.6-Terra. The re-extractions differed between the two models, for example in the number of features per workflow. Results for other models, future versions, and smaller open-source models are unknown. The stored session logs partly mitigate this, since they keep accumulated expertise independent of a specific model.

\textbf{Size and readability.}
ATWL codes are long. The code of a library workflow has a median of about 500 lines including comments, and the longest has about 1,250. This impedes reading and checking, and one expert found the graph too large to verify all edges, while experts asked for aggregation and filtering of large graphs. We are testing a syntax extension for modular representation. A \kw{module} groups consecutive transforms and loops into a named unit with a purpose and declared inputs and outputs, which are computed from the data flow. Modules can be folded, so that a reader first sees the phases of a workflow. For each of the 54 library workflows, the Web supplement has a companion page with the modular representation (median of five modules per workflow). The viewers, the syntax checker, and the analyses do not support modules yet, and we have not evaluated whether modules improve readability.

\subsection{Further Work and Community Adoption}
\label{sec:discussion:adoption}

Several directions follow from the limitations. First, branching needs language support and examples for conditional paths, alternatives, and fallbacks beyond loop exits. Second, the viewers, the syntax checker, and the analyses need to support modules, and the effect of modules on readability needs to be evaluated. The experts further asked for clearer marking of human decisions and for integration into code editors with cross-references and natural-language summaries. Third, tests are needed of the use of ATWL by people who did not co-author the workflow.

Fourth, workflows are often not fully defined in advance. We know from our earlier analysis of the data from the VAST outbreak task that the data revealed two groups of symptoms with different modes of transmission, which a workflow designed beforehand cannot anticipate. The workflow then has to be revised during execution, and the design and the prototype need to evolve together while the expert contributes knowledge. Representing partially defined workflows, in which parts are decided only after earlier results are known, together with their revisions, requires a conceptual extension of ATWL, and we have started to work on it.

The potential of formal workflow representation can be realised only through sustained community engagement. We invite VA researchers to document the workflows in their publications with ATWL and to contribute them to a public library. For individual researchers, formalisation imposes a disciplined reflection on analytical choices. A formal representation can accompany a publication as supplementary material. For the community, a growing library would allow the hypotheses from Section~\ref{sec:analysis} to be tested and refined, and would improve recommendations by expanding the set of available building blocks.

\textbf{Longer-term directions.} Formal workflows also invite questions that are hard to ask of prose. A library of typed building blocks with declared interfaces could support compositional workflow design, in which type compatibility is a basic validity criterion. With a larger library, one could test which structural properties, for example closed cycles with an assessment, are necessary for a given analytical goal and which are habits of current practice. Distinguishing analytical strategies, such as exploratory and confirmatory ones, and the division of labour between humans and machines could show where automation can substitute for human judgement.

\textbf{Open resources.}
The public repository (\url{https://geoanalytics.net/VAworkflows/index.html}) contains the ATWL language definition, the workflow library with graphical and modular views, the ATWL Workflow Lens, the ATWL syntax checker, and stored LLM session logs for four tasks: \kw{extractor} (workflow extraction and formalisation), \kw{reviewer} (systematic review of representations), \kw{diagrammer} (flow diagram generation), and \kw{recommender} (workflow recommendations for new problems). The logs encode accumulated expertise and prime an advanced LLM with ATWL knowledge and conventions. We recommend the following procedure for a new paper. A user loads the extractor session, the language definition with 2-3 workflow examples in ATWL (to refresh the mode's knowledge), and the paper into one model to obtain an initial representation, runs the syntax checker, and passes its errors back for correction. A different model, primed with the reviewer session (and also with the ATWL knowledge refreshed), reviews the syntactically correct representation against the paper, and the first model revises it. Review and revision are repeated until the reviewer finds no issues, and an author of the workflow or another person who knows it well checks the result. In our re-extraction check we used one round of review and revision. We welcome new workflow representations, feedback on the language, and proposals for extensions motivated by workflows that the current constructs do not capture.

\section{Conclusion}
\label{sec:conclusion}

We have presented ATWL, a formal language for representing visual analytics workflows, and a library of 54 workflows extracted from published papers. The language gives the VA community a shared vocabulary in which analytical processes can be compared, decomposed, and processed by machines. Cross-workflow analysis of the library shows structural regularities, such as a common backbone, recurring motifs, and closed cycles, which are hard to see when the workflows are read as prose. The authors of ten library papers rated the representations of their workflows as accurate. Experiments with two LLMs on two tasks show that the library leads designs to cite existing workflows and that ATWL makes loop exits and outputs explicit. They also show a gap: conditional paths beyond loops are not expressed. A check with five re-extracted workflows shows that the backbone of a workflow is reproduced and the details are not, so the authors of a workflow should check its representation. The evidence is exploratory, since the library is not a representative sample and the experiments are small. Our next steps are support for branching, modular, and evolving workflows and a library that grows through contributions from the community. Such a library could support a cumulative study of human--computer analytical processes.

\balance

\bibliographystyle{IEEEtran}
\bibliography{ref}

\clearpage
\setcounter{page}{1}
\twocolumn[%
  \begin{center}
    \vspace*{1ex}
    {\large Appendix to:\par}
    \vspace{1ex}
    {\LARGE\bfseries ATWL: A Formal Language for Representing,\\
     Comparing, and Reusing Visual Analytics Workflows\par}
    \vspace{1.5ex}
    {\normalsize Natalia~Andrienko, Gennady~Andrienko,
     J\"urgen~Bernard, Kuno Kurzhals, and Michael~Sedlmair\par}
    \vspace{2ex}
  \end{center}%
]

\appendices

\newcommand{\meta}[1]{\textit{$\langle$#1$\rangle$}}

\section{ATWL Syntax Reference}
\label{sec:atwl-syntax-reference}

This section provides a concise syntax reference for the
\textbf{Artifact--Transform Workflow Language (ATWL)},
a declarative language for representing visual analytics workflows
as structured transformations of artifacts.
Angle brackets (\kw{<\,>}) denote placeholders;
the \kw{\#} character introduces comments.

\subsection{Workflow Declaration}
\label{sec:syn-workflow}

\begin{lstlisting}[style=atwl]
workflow <workflow-ID>
  template: <intent> -> <intent> -> ...       # optional
  description: "<text>"                        # optional
\end{lstlisting}
\begin{itemize}[nosep]
  \item \kw{template} summarises the main analytic stages using the
        same intent vocabulary as transforms (Section~\ref{sec:syn-intents}).
        Parenthetical qualifiers (e.g.\ \kw{(similarity-based)})
        are informal annotations, not new intents.
  \item Iteration may be indicated with
        \kw{loop(<intent> -> <intent> -> ...)}.
        Loop notation in templates is optional and used only when the iterative
        structure is a key characteristic of the analytical method.
  \item Artifacts and transforms may be declared in any order that aids
        readability.
\end{itemize}

\subsection{Artifacts}
\label{sec:syn-artifacts}

ATWL defines eight artifact types.
Every artifact is declared with a unique identifier and its type.
The field \kw{origin: given} marks \emph{exogenous} artifacts not
produced by any transform in the workflow; it is omitted for artifacts
that appear as transform outputs.
Artifact type declarations may include references to other artifacts in
parentheses, e.g.\ \kw{feature(D1)}.

\subsubsection{Entities}
\label{sec:syn-entities}

A collection of identifiable analytical objects of the same kind, each
treated as a single (possibly structured) piece of data for the purposes
of analysis.

\begin{lstlisting}[style=atwl]
artifact <ID> : entities
  origin: given                          # only for exogenous artifacts
  internal structure: <structure-type>
  embedment: <embedment-type or {...}>   # omitted for single entity
  features:
    - id: <feature_id>

      value structure: <structure>
      value type: <type>                 # optional
      description: "<text>"
  description: "<text>"
\end{lstlisting}

\textbf{Internal structure.}
Describes how components inside each entity are organised.

\begin{table}[b]\centering\footnotesize
\begin{tabularx}{\columnwidth}{@{}l >{\raggedright\arraybackslash}X l @{}}
\toprule
\textbf{Type} & \textbf{Description} & \textbf{Category} \\
\midrule
\kw{elementary}    & Indivisible unit in workflow & Atomic     \\
\kw{group/cluster} & Unordered collection & Container  \\
\kw{episode}       & Bounded time interval; temporal units & Container  \\
\kw{region}        & Bounded spatial extent; spatial units & Container  \\
\kw{sequence}      & Components in linear order & Relational \\
\kw{formation}     & General relational structure (network) & Relational \\
\bottomrule
\end{tabularx}
\caption{Types of internal structure for entities.}
\label{tab:internal-structure}
\vspace{9pt}
\end{table}

\textbf{Embedment.}
Describes the shared environment(s) in which the entities reside and how
they are related to each other.  Omitted when the artifact represents a
single entity.  Values may be single or combined in set notation
(e.g.\ \kw{\{set, time\}}).

\begin{table}[ht]\centering\footnotesize
\begin{tabularx}{\columnwidth}{@{}l >{\raggedright\arraybackslash}X @{}}
\toprule
\textbf{Embedment} & \textbf{Description} \\
\midrule
\kw{set}                 & Unordered collection \\
\kw{sequence}            & Ordered positions (no metric) \\
\kw{time}                & Temporal axis (abs./rel.) \\
\kw{space}               & Spatial reference (coords/geom) \\
\kw{relational}          & Network or hierarchical structure \\
\bottomrule
\end{tabularx}
\caption{Embedment types for entities.}
\vspace{9pt}
\label{tab:embedment}
\end{table}

\textbf{Features (inside entities).}
Each feature declares a \textbf{value structure} and, optionally, a
\textbf{value type}.  The same scheme applies to standalone feature
artifacts (Section~\ref{sec:syn-feature}).

\begin{table}[ht]\centering\footnotesize
\begin{tabularx}{\columnwidth}{@{}l >{\raggedright\arraybackslash}X @{}}
\toprule
\multicolumn{2}{@{}l}{\textbf{Value structure} --- overall organisation} \\
\midrule
\kw{atomic}   & Single value per entity \\
\kw{list}     & Enumeration of values \\
\kw{vector}   & Fixed-length array \\
\kw{matrix}   & 2D array of values \\
\kw{relational} & Irregular structure (graph/tree) \\
\midrule
\multicolumn{2}{@{}l}{\textbf{Value type} (optional) --- atomic components} \\
\midrule
\kw{numeric}     & Quantitative counts/measures \\
\kw{ordinal}     & Ordered values, no metric \\
\kw{categorical} & Unordered discrete categories \\
\kw{temporal}    & Time points or durations \\
\kw{spatial}     & Coordinates or geometries \\
\kw{text}        & Textual content \\
\kw{reference}   & ID pointing to other artifacts \\
\bottomrule
\end{tabularx}
\caption{Value structure and value type for features.}
\vspace{9pt}
\label{tab:value-structure}
\end{table}

When atomic components are of mixed types, set notation is used
(e.g.\ \kw{\{numeric, temporal\}}).  The value type may be omitted
when the mixture is complex or the types are evident from context.

\subsubsection{Feature}
\label{sec:syn-feature}

Explicit descriptors of properties of entities or relationships between
them.  Typically produced by \kw{characterise} transforms.

\begin{lstlisting}[style=atwl]
artifact <ID> : feature(<artifact-ID>)
  value structure: <structure>
  value type: <type>                   # optional
  representation form: "<text>"        # optional
  description: "<text>"
\end{lstlisting}

\kw{<artifact-ID>} refers to the artifact described by the feature.
\kw{representation form} clarifies encoding for complex features
(e.g.\ \kw{"similarity matrix"}, \kw{"k-NN graph"}).

\subsubsection{Arrangement}
\label{sec:syn-arrangement}

An arrangement organises entities in a context, defining how they are
positioned within a reference structure.  It does not create new entities
or values.

\begin{lstlisting}[style=atwl]
artifact <ID> : arrangement(<entities-ID>)
  context: <entities-ID>
  principle: "<text>"
  description: "<text>"
\end{lstlisting}

\begin{itemize}[nosep]
  \item \kw{<entities-ID>}: the entities whose members are arranged.
  \item \kw{context}: an entities artifact acting as the reference
        structure (e.g.\ calendar days, map regions, a projection layout).
  \item \kw{principle}: how positions are determined (e.g.\
        \kw{"calendar(year, month, weekday)"},
        \kw{"2D projection based on similarity"}).
\end{itemize}

\subsubsection{Visualisation}
\label{sec:syn-visualisation}

An external visual representation for human perception.

\begin{lstlisting}[style=atwl]
artifact <ID> : visualisation(<artifact-IDs>)
  layout: "<text>"
  form: "<text>"
  encoding: "<text>"                   # optional but recommended
  description: "<text>"
\end{lstlisting}

\begin{itemize}[nosep]
  \item \kw{<artifact-IDs>}: one or more artifacts of any type.
  \item \kw{layout}: structure of the visual space
        (e.g.\ \kw{"calendar grid"}, \kw{"2D projection"}).
  \item \kw{form}: visual mark type
        (e.g.\ \kw{"coloured marks"}, \kw{"node-link diagram"}).
  \item \kw{encoding}: how features and arrangements map to
        visual attributes.
\end{itemize}

\subsubsection{Pattern}
\label{sec:syn-pattern}

An abstract regularity or structure identified in data, features,
arrangements, visualisations, or models.

\begin{lstlisting}[style=atwl]
artifact <ID> : pattern(<artifact-IDs>)
  representation form: "<text>"
  description: "<text>"
\end{lstlisting}

\kw{representation form} describes how the pattern is represented
(e.g.\ \kw{"textual descriptions"}, \kw{"ranked list of motifs"},
\kw{"rules"}).

\subsubsection{Model}
\label{sec:syn-model}

A formal or computational representation used for prediction, simulation,
or explanation.

\begin{lstlisting}[style=atwl]
artifact <ID> : model(<artifact-IDs>)
  model type: "<text>"
  representation form: "<text>"        # optional
  description: "<text>"
\end{lstlisting}

\begin{itemize}[nosep]
  \item \kw{model type}: high-level category
        (e.g.\ \kw{"classifier"}, \kw{"topic model"},
         \kw{"simulation model"}).
  \item \kw{representation form}: internal parametric or structural form
        (e.g.\ \kw{"decision tree"}, \kw{"neural network weights"}).
\end{itemize}

\subsubsection{Knowledge}
\label{sec:syn-knowledge}

Explicitly formulated knowledge, either \emph{derived} in the analysis
(insights, explanations, rules) or \emph{injected} during the analysis
(domain expertise, constraints, feedback).

\begin{lstlisting}[style=atwl]
artifact <ID> : knowledge(<artifact-IDs>)
  origin: given                        # for injected; omitted for derived
  representation form: "<text>"
  description: "<text>"
\end{lstlisting}

\kw{<artifact-IDs>} reference the artifacts this knowledge is based on or
applies to.
\kw{representation form} examples: \kw{"statements"},
\kw{"if-then rules"}, \kw{"labels"}, \kw{"quality judgment"}.

Note: \kw{assess} transforms produce \emph{evaluative} knowledge
(quality judgments, adequacy decisions), while \kw{generate-knowledge}
transforms produce \emph{substantive} knowledge (domain insights,
conclusions).

\subsubsection{Specification}
\label{sec:syn-specification}

An explicit description of parameters, settings, constraints, or method
choices that determine how transforms are executed.

\begin{lstlisting}[style=atwl]
artifact <ID> : specification
  origin: given                        # for injected; omitted for derived
  representation form: "<text>"
  description: "<text>"
\end{lstlisting}

\kw{representation form} examples:
\kw{"parameter settings"},
\kw{"method choice"},
\kw{"constraints"},
\kw{"configuration schema"}.

Specifications are typically consumed by transforms whose behaviour
depends on configurable choices.  They may be given (exogenous) or
derived---most commonly by \kw{generate-knowledge} or \kw{assess}
transforms.

\subsection{Transforms}
\label{sec:syn-transforms}

Each transform consumes one or more input artifacts and produces one or
more output artifacts.

\begin{lstlisting}[style=atwl]
transform <ID> :
  intent: <generic-intent>
  manner: "<text>"                     # optional specialisation
  input: <artifact-IDs>
  output: <artifact-IDs>
  actor: human | machine | hybrid
  description: "<text>"
\end{lstlisting}

\begin{itemize}[nosep]
  \item \kw{intent}: one of the generic intents in
        Table~\ref{app:tab:intents}.
  \item \kw{manner}: optional free-text specialisation of intent
        (see Table~\ref{tab:manner}).
  \item \kw{input}/\kw{output}: comma-separated artifact IDs.
  \item \kw{actor}:
        \kw{human} (analyst only),
        \kw{machine} (computation only), or
        \kw{hybrid} (human--machine collaboration).
\end{itemize}

\subsubsection{Generic Intents}
\label{sec:syn-intents}

\begin{table}[ht]\centering\footnotesize
\begin{tabularx}{\columnwidth}{@{}l >{\raggedright\arraybackslash}X >{\raggedright\arraybackslash}p{2.2cm} @{}}
\toprule
\textbf{Intent} & \textbf{Purpose} & \textbf{Typical outputs} \\
\midrule
\kw{define-unit}    & Create/redefine units (extract, group, aggregate) & entities (+ feature) \\
\kw{characterise}   & Compute/transform features & feature \\
\kw{contextualise}  & Place entities in context & arrangement \\
\kw{visualise}      & Create visual reps. & visualisation \\
\kw{abstract}       & Derive patterns/structures & pattern (+ feature) \\
\kw{build-model}    & Construct/refine models & model \\
\kw{generate-knowledge}  & Formulate knowledge & knowl./spec. \\
\kw{assess}         & Evaluate quality/adequacy & knowl. (+ spec.) \\
\bottomrule
\end{tabularx}
\caption{Generic transform intents.}
\label{app:tab:intents}
\vspace{9pt}
\end{table}

\subsubsection{Specialisations via Manner}
\label{sec:syn-manner}

The \kw{manner} field is not part of the core type system but is
recommended for consistency and tool interoperability.
Table~\ref{tab:manner} gives illustrative (extensible) values.

\begin{table}[ht]\centering\footnotesize
\begin{tabularx}{\columnwidth}{@{}l >{\raggedright\arraybackslash}X @{}}
\toprule
\textbf{Intent} & \textbf{Example manner values} \\
\midrule
\kw{define-unit}    & extract, filter, partitioning, cluster, group, merge \\
\kw{characterise}   & summarise, aggregate, profile, project, encode, relate \\
\kw{contextualise}  & calendar-, map-, or projection-based; time-alignment \\
\kw{visualise}      & line-graph, coloured-marks, node-link, matrix-plot \\
\kw{abstract}       & pattern-mining, salient-groups, interpretation \\
\kw{build-model}    & train-classifier, fit-topic-model, calibrate \\
\kw{generate-knowledge}  & formulate-statements, derive-rules, write-summary \\
\kw{assess}         & evaluate-quality, assess-performance, judge-adequacy \\
\bottomrule
\end{tabularx}
\caption{Illustrative \kw{manner} values by intent.}
\label{tab:manner}
\vspace{9pt}
\end{table}

\subsection{Control Structures}
\label{sec:syn-control}

Control structures describe analytic logic, not executable control.

\subsubsection{Loop}
\label{sec:syn-loop}

\begin{lstlisting}[style=atwl]
loop <ID>:
  purpose: "<text>"
  until: "<qualitative stopping condition>"
  body:
    <transforms, artifact declarations, conditionals, assignments>
end loop <ID>
\end{lstlisting}

The \kw{until} field states a qualitative stopping condition.
Two styles of termination are supported:

\begin{itemize}[nosep]
  \item \textbf{Explicit assessment.}  The loop body contains an
        \kw{assess} transform followed by a conditional with
        \kw{exit loop <ID>} in one branch.
  \item \textbf{Implicit termination.}  No explicit assessment or
        conditional exit appears; the \kw{until} condition is
        monitored informally by the analyst.
\end{itemize}

\subsubsection{Conditional}
\label{sec:syn-conditional}

\begin{lstlisting}[style=atwl]
if <condition>:
  then:
    <transforms and artifacts>
  else:
    <transforms and artifacts>
\end{lstlisting}

Conditions refer to artifact properties or human judgements.
Use \kw{exit loop <ID>} in a branch to leave an enclosing loop.

\subsubsection{Assignment}
\label{sec:syn-assignment}

\begin{lstlisting}[style=atwl]
assign:
  <artifact-ID>  := <artifact-ID'>
  <artifact-ID2> := <artifact-ID2'>
\end{lstlisting}

Assignments bind an artifact identifier to a new version.
They appear in two contexts:
\begin{itemize}[nosep]
  \item \textbf{Before a loop}, to initialise an identifier that will be
        reassigned during iteration.
  \item \textbf{Inside a loop body}, to express iterative update of an
        artifact.
\end{itemize}

\subsection{Workflow Validity and Conventions}
\label{sec:syn-validity}

A well-formed ATWL workflow should satisfy:
\begin{enumerate}[nosep]
  \item All artifact identifiers are unique within the workflow.
  \item Every transform input references a declared artifact.
  \item Artifacts with \kw{origin: given} are never transform outputs.
  \item The artifact dependency graph is acyclic (except for explicit
        \kw{assign} statements in loops).
\end{enumerate}

\noindent Additional conventions:
\begin{itemize}[nosep]
  \item Exogenous artifacts (\kw{origin: given}) are typically
        declared near the top or near their first use.
  \item Lists of artifact IDs are comma-separated.
  \item Artifact references appear in parentheses after the type keyword,
        e.g.\ \kw{feature(D1)}, \kw{arrangement(D\_events)}.
  \item Each transform carries exactly one \kw{intent}, even when
        it produces multiple artifact types; the intent reflects the
        primary analytic purpose.
\end{itemize}

The ATWL syntax checker (\url{https://geoanalytics.net/VAworkflows/library/atwl_syntax_checker.html}) verifies a specification for syntax errors and cross-reference problems, such as undeclared artifacts, unclosed loops, and unknown vocabulary. The specification can be pasted, uploaded, or loaded from the URL of a library page with an embedded ATWL source. The checker offers automatic corrections where the correction is obvious, for example for misspelled keywords. For problems without an obvious correction, such as an unknown transform intent, it issues a warning. If no problems are found, the checker computes statistics and diagrams of the workflow. All workflows of the library pass the checks.

\section{Example: Cluster-Calendar Workflow Represented in ATWL}
\label{sec:cluster-calendar_ATWL}
Source \cite{vanWijkSelow1999}:
Jarke J. van Wijk and Edward R. van Selow. \\
Cluster and calendar based visualization of time series data. \\
In Proceedings of the IEEE Symposium on Information Visualization (InfoVis ’99), pages 4–9, \\
Los Alamitos, CA, USA, 1999. IEEE Computer Society.\\
doi: 10.1109/INFVIS.1999.801851

\subsection*{Concise Workflow Summary}
The cluster-calendar workflow combines hierarchical clustering with calendar-based visualization to identify and analyse recurring patterns in time series data measured at regular intervals (e.g., hourly) over extended periods. The workflow partitions continuous time series into daily episodes, characterizes each day by its temporal profile, and applies hierarchical bottom-up clustering to group days with similar patterns. Users iteratively refine the cluster structure—adjusting the number of clusters, selecting alternative distance measures, or focusing on specific time intervals—until meaningful behavioural patterns emerge. The results are presented through coordinated visualizations: a calendar view where days are colour-coded by cluster membership to reveal weekly and seasonal distributions, and line graphs showing the average temporal profile for each cluster to illustrate characteristic patterns. Through interactive exploration of these coordinated views, analysts identify both standard patterns (e.g., typical weekdays, weekends, seasonal variations) and exceptional days (e.g., holidays, special events), formulating insights about temporal regularities and anomalies that would be difficult to detect through traditional time series analysis methods.

A flow diagram representing the Cluster-Calendar workflow is shown in Fig.\ref{fig:cluster-calendar}.

\subsection*{ATWL representation:}
\begin{lstlisting}[style=atwl, basicstyle=\footnotesize\ttfamily]
workflow cluster-calendar

template: define-unit -> contextualise -> characterise -> loop(define-unit (similarity-based) ->  characterise (groups) -> visualise ->  abstract -> assess) -> generate-knowledge

description: "Identify and analyze recurring daily patterns in time series data through interactive clustering and calendar-based visualization;   detect standard patterns and exceptional days"

artifact D_hour : entities
    origin: given
    internal structure: elementary
    embedment: time
    features:
      - id: f_value
        value structure: atomic
        value type: numeric
        description: "Measured value"
    description: "Time series measurements at regular intervals over extended period"

artifact D_calendar : entities
    origin: given
    internal structure: elementary
    embedment: time
    features:
      - id: f_temporal_coords
        value structure: vector
        value type: {categorical, numeric}
        description: "Month, day of week, day number"
    description: "Calendar structure providing temporal context with month and weekday organization"        

transform T_partition :
    intent: define-unit
    manner: "time-partitioning into daily episodes"
    input: D_hour
    output: D_day
    actor: machine
    description: "Organize time series into daily episodes, each containing  measurements for one 24-hour period"

artifact D_day : entities
    internal structure: episode
    embedment: time
    features:
      - id: f_day_index
        value structure: atomic
        value type: ordinal
        description: "Sequential position of day in the year"
    description: "Daily episodes consisting of all measurements within each  24-hour period"

transform T_arrange :
    intent: contextualise
    manner: "calendar-based"
    input: D_day, D_calendar
    output: A_calendar
    actor: machine
    description: "Arrange daily episodes in calendar context
        according to their temporal position"

artifact A_calendar : arrangement(D_day)
    context: D_calendar
    principle: "calendar date mapping to grid position"
    description: "Calendar-based arrangement where each day occupies its corresponding calendar cell"

transform T_profile :
    intent: characterise
    manner: "extract temporal profile"
    input: D_day
    output: F_day_profile
    actor: machine
    description: "Represent each day by its measurement sequence"

artifact F_day_profile : feature(D_day)
    value structure: vector
    value type: numeric
    description: "Daily temporal profile: sequence of measurements within each day"

artifact S_clustering : specification
    origin: given
    representation form: "parameter settings"
    description: "Initial parameters for hierarchical clustering: number of clusters (dendrogram cut level), distance measure (geometric, normalized, shift-invariant, max-based), time interval focus"

loop L1:
    purpose: "Iteratively explore cluster structure to identify meaningful and 
        interpretable daily patterns"
    until: "Clusters provide clear, interpretable decomposition of daily patterns; 
        standard patterns and exceptional days are identified"
    body:
        transform T_cluster :
            intent: define-unit
            manner: "hierarchical clustering by similarity"
            input: D_day, F_day_profile, S_clustering
            output: D_cluster, F_cluster_label
            actor: hybrid
            description: "Apply hierarchical clustering to group days with similar profiles; user selects cut through dendrogram to determine clusters"
        
        artifact D_cluster : entities
            internal structure: group/cluster
            embedment: set
            features:
              - id: cluster_size
                value structure: atomic
                value type: numeric
                description: "Number of days in cluster"
            description: "Groups of days with similar daily profiles selected from hierarchical clustering tree"
        
        artifact F_cluster_label : feature(D_day)
            value structure: atomic
            value type: categorical
            description: "Cluster membership identifier for each day"
        
        transform T_aggregate :
            intent: characterise
            manner: "aggregate profiles per cluster"
            input: D_cluster, F_day_profile
            output: F_cluster_profile
            actor: machine
            description: "Compute average daily profile for each cluster to  represent typical pattern"
        
        artifact F_cluster_profile : feature(D_cluster)
            value structure: vector
            value type: numeric
            description: "Cluster-level average daily profiles representing typical patterns for each group"
        
        transform T_calendar_vis :
            intent: visualise
            manner: "calendar grid with color-coded clusters"
            input: A_calendar, F_cluster_label
            output: V_calendar
            actor: machine
            description: "Display days on calendar grid, colored by cluster membership"
        
        artifact V_calendar : visualisation(A_calendar, F_cluster_label)
            layout: "calendar grid (months as rows, weekdays as columns)"
            form: "colored cells"
            encoding: "position from A_calendar; color from F_cluster_label"
            description: "Calendar view showing temporal distribution of cluster patterns across year and week"
        
        transform T_profile_vis :
            intent: visualise
            manner: "line graphs of cluster profiles"
            input: F_cluster_profile, D_cluster
            output: V_profiles
            actor: machine
            description: "Display average daily profile for each cluster as line graph"
        
        artifact V_profiles : visualisation(F_cluster_profile, D_cluster)
            layout: "time axis (hour of day)"
            form: "line graphs (one per cluster)"
            encoding: "x-position: time within day; y-position: average measurement value; color: cluster identity matching calendar colors"
            description: "Line graphs showing characteristic temporal patterns for each cluster"
        
        transform T_interpret :
            intent: abstract
            manner: "interpret cluster meanings"
            input: V_calendar, V_profiles, D_cluster, F_cluster_profile
            output: P_patterns
            actor: human
            description: "Interpret cluster patterns: identify behavioral meaning of each cluster type"
        
        artifact P_patterns : pattern(D_cluster, F_cluster_profile)
            representation form: "textual labels and descriptions"
            description: "Interpreted meanings of daily patterns (e.g., 'typical weekday', 'weekend', 'holiday', 'summer Friday', 'exceptional event')"
        
        transform T_assess_clusters :
            intent: assess
            manner: "evaluate cluster quality and interpretability"
            input: V_calendar, V_profiles, P_patterns, D_cluster
            output: cluster_assessment
            actor: human
            description: "Assess whether clusters provide meaningful decomposition: patterns are interpretable, clusters well-separated, standard vs. exceptional days identified"
        
        artifact cluster_assessment : knowledge(D_cluster)
            representation form: "quality judgment"
            description: "Assessment of cluster quality: interpretability, separation, coverage of pattern types, and whether refinement with adjusted parameters is needed"
        
        if cluster_assessment indicates refinement needed:
            then:
                transform T_adjust :
                    intent: generate-knowledge
                    manner: "adjust clustering parameters based
                        on assessment"
                    input: cluster_assessment, V_calendar,
                           V_profiles, S_clustering
                    output: S_clustering'
                    actor: human
                    description: "Adjust clustering parameters: modify number of clusters, select different distance measure, or change time interval focus"

                artifact S_clustering' : specification
                    representation form: "parameter settings"
                    description: "Updated clustering parameters after analyst refinement"

                assign:
                    S_clustering := S_clustering'
            else:
                exit loop L1
end loop L1

transform T_synthesize :
    intent: generate-knowledge
    manner: "formulate statements about temporal patterns"
    input: P_patterns, V_calendar, V_profiles, cluster_assessment
    output: K1
    actor: human
    description: "Synthesize findings: document discovered patterns, their temporal distribution, correlations with external events, and exceptional occurrences"

artifact K1 : knowledge(P_patterns)
    representation form: "statements and explanations"
    description: "Understanding of temporal patterns: standard daily patterns identified, their distribution over week and year, correlation with  calendar events, exceptional patterns and their causes"
\end{lstlisting}

\section{Language Design: Alternatives and Decisions}
\label{app:language-design}

This appendix lists the main alternatives that we considered while designing ATWL (Section~\ref{sec:language}) and the reasons for our decisions. It is based on records of the design discussions, which we held with the help of an LLM assistant.

\textbf{A lightweight typed vocabulary.} We considered rigorous bases for the language, such as category theory, process algebra, typed graphs, logic-based rules, and state-transition systems, and a combination of theories from systems theory, cognitive science, and semiotics. Rigorous formalisms are heavy for VA researchers, and a combination of theories was too loose. To treat workflows as comparable empirical objects, a controlled vocabulary, typed constructs, and a compositional structure are sufficient, so we chose a declarative notation.

\textbf{One kind of transform with an intent.} We considered a large fixed set of operators (for example segment, group, compare, label, annotate, validate, refine, branch, and iterate). We judged it too complex to motivate analysts to document their workflows. A single transform with one of eight intents, an optional free-text manner, and control structures is simpler and expressive enough.

\textbf{Human contributions through artifacts and actors.} We considered operator types for cognitive acts, such as interpreting, hypothesising, and incorporating knowledge. Cognitive processes cannot be observed or reproduced. Only their externalised results, such as labels, groups, rules, and features, can be recorded and supported by tools, so a transform states who performs it (human, machine, or hybrid) and the results are artifacts.

\textbf{Patterns as configurations of artifact roles.} We considered describing recurring workflow patterns as sequences of transforms. Such sequences are tied to methods, and small differences in how a feature is derived fragment them into variants. The roles of artifacts (data, feature, pattern, arrangement, knowledge) stay the same across domains.

\textbf{Arrangement as an artifact.} We considered arrangement as a property of a transform. Arrangements are reusable, both for computation and for visualisation, and they can themselves be transformed, so they are artifacts. Embedment (an inherent property of data), arrangement (an analytical organisation produced by \kw{contextualise}), and visualisation (a perceptual encoding) are kept as three levels.

\textbf{Entities.} We considered ``dataset'', which suggests a collection and does not fit a single object such as a network; a separate ``data'' type next to ``units'', which is artificial, since a table row is already an entity; and ``units'', which is too abstract. ``Entities'' covers original and derived objects.

\textbf{Specification as an artifact type without its own intent.} We considered ``configuration'', which collides with the value structure ``relational configuration'', and ``parameters'', ``control'', and ``settings''. We also considered folding specifications into knowledge. Knowledge is what is understood, and a specification is what is decided to steer later transforms. Setting parameters is a means and not an analytical goal, so no intent produces specifications: they are given, or they result from \kw{assess} or \kw{generate-knowledge}.

\textbf{The \kw{assess} intent.} Judging the adequacy of results, such as the coherence of clusters or the fit of a model, fell between \kw{abstract} and \kw{generate-knowledge}, and it is what makes iterative refinement meaningful. We considered representing such judgements as patterns. They are interpretive statements about results and not regularities in the data, so they are knowledge.

\textbf{The internal structure \texttt{region}.} \texttt{episode} existed for bounded time intervals, but no type covered bounded spatial extents with contained objects. We introduced \texttt{region} as its spatial counterpart.

\textbf{Granularity and detail.} We considered representing every small transformation separately. Steps that do not change the artifact type or the level of abstraction are collapsed, at the analyst's discretion. Loops are used only for analytically significant iteration, such as progressive refinement, and not for implementation-level iteration, such as processing each document of a corpus. Algorithm names and parameters (``density-based clustering'', not a named algorithm) stay in the free-text manner or out of the workflow, so that workflows remain comparable across domains.

\textbf{Intended use.} We considered a dialogue-based assistant that elicits the goals of a user and recommends workflows as the main use. It needs much more than a language: procedural knowledge about analysis alternatives, conditions of applicability, and dialogue strategies. Documentation, comparison, and LLM-assisted formalisation already give value with little effort, so they are the uses that we pursue (Section~\ref{sec:pract_value}).

\textbf{Left open.} We did not define a catalogue of visual patterns, such as clusters, trends, and periodicities, but leave it to existing taxonomies. We did not model internal cognitive processes.

\section{Workflow Library Construction}
\label{sec:library-construction}

This appendix describes how the library of 54 workflows was selected, extracted, and checked. The library, the instructions for the LLM instances, and the related reports are available at \url{https://geoanalytics.net/VAworkflows/}. For each workflow, the library page shows the summary and the ATWL text and provides links to interactive graphical and structural views. The web tool ATWL Workflow Lens (\url{https://geoanalytics.net/VAworkflows/library/atwl_workflow_lens}) is a web application for exploring and comparing the library workflows by their structural features.

\subsection{Selection of Papers}
\label{sec:library-selection}

The library was built in two stages. The first stage produced 18 workflows, which we used to develop and test the language. We aimed at diversity of data types and analytical paradigms: exploratory pattern discovery, data simplification for comprehensibility, model building and validation, and model understanding and diagnostics. We did not aim at a statistical sample of VA research. A later addition, EventThread, made the number of workflows 18. In the second stage, we extended the library to 54 workflows. In both stages, we looked for papers that describe a complete analytical workflow with multiple steps from data to knowledge. This requirement is selective, because many VA papers focus on individual techniques or tools with usage examples that illustrate capabilities rather than a complete analytical process.

To select the additional papers, we first took the relevant papers recommended by the co-authors. Then we asked an LLM (Claude Opus 4.6) to analyze the abstracts of the publications in the IEEE VIS dataset that we also used in the experiments of Section~\ref{sec:recommendation}, and to estimate for each paper the probability that it contains a workflow description. The results are available at \url{https://geoanalytics.net/VAworkflows/atwl_paper_ranker_v3.html}. We skimmed the highly ranked papers and chose those that we considered most suitable in terms of workflow sophistication and that increased the diversity of the library. Papers outside the dataset entered the library through the recommendations of the co-authors.

Of the 54 papers, 26 have at least one co-author of this article among their authors: 9 by N.~Andrienko and G.~Andrienko, 7 by J.~Bernard, 6 by K.~Kurzhals, and 5 by M.~Sedlmair. One paper is shared by J.~Bernard and M.~Sedlmair. The co-authors have worked in different and mostly non-overlapping areas of VA, for example, movement and spatio-temporal data, time-oriented data and event sequences, machine learning models and simulation parameter spaces, and eye tracking. Therefore, the workflows of the co-authors' papers already cover a broad part of the field. 

\subsection{Extraction and Evaluation}
\label{sec:library-extraction}

The workflows were extracted by a collaborative process of LLM instances and human experts. For each paper, an \emph{extractor} instance extracted the analytical workflow from the prose, figures, and algorithmic descriptions of the paper and produced an ATWL specification. It also wrote a short summary of the workflow, which is shown on the library page together with the ATWL representation. An \emph{evaluator} instance (called \emph{reviewer} in the instructions and in the supplement) then verified the syntactic correctness, ontological consistency, and logical completeness of the specification and generated structured feedback. The feedback was passed back to the extractor for revision. This cycle was repeated until the specification satisfied all validity constraints (Section~\ref{sec:syntax-overview}). Most workflows were extracted and evaluated by Claude Opus 4.6, which took both roles; GPT-5 and Claude Sonnet 5 were used only for a few of the most recently added workflows. An instance processed several papers one after another in one session, so that it could use what it had learned from earlier extractions and evaluations. We now recommend that the two roles are taken by different models, that the syntax checker is run and its errors are corrected before the first evaluation, that evaluation and revision are repeated until the evaluator finds no issues, and that an author of the workflow or another person who knows it well checks the result. A detailed description of the process is available at \url{https://geoanalytics.net/VAworkflows/LLM_experiments_extract_review.pdf}.

Human supervision remained necessary. Where the iteration of the LLM instances did not resolve an issue, a human expert provided corrective prompts. In a few cases, the extractor was uncertain how to interpret parts of the paper text and needed feedback from us. These uncertainties resulted from ambiguities in the texts. We therefore do not expect that fully automatic extraction can be achieved. The role of the human expert shifts from writing ATWL specifications from scratch to resolving ambiguities and to evaluating and correcting machine-generated proposals.

\subsection{Audit and Instructions}
\label{sec:library-audit}

After the first 16 workflows had been extracted, we conducted a systematic audit of the library. It was done collaboratively by a human expert and a new LLM instance. For each workflow, the LLM identified remaining issues and proposed corrections, and the expert accepted, modified, or rejected each proposal. In several cases, the disagreements between the expert and the LLM led to discussions among the co-authors, which resulted in refinements of the language definition. The audit report is available at \url{https://geoanalytics.net/VAworkflows/review_16_workflows.pdf}.

The patterns of errors found in the audit were used to develop procedural instructions for the extractor and the evaluator (see \url{https://geoanalytics.net/VAworkflows/extraction_instructions.pdf} and \url{https://geoanalytics.net/VAworkflows/reviewing_instructions.pdf}). To check the instructions, we repeated the extraction with two new LLM instances that had not participated in any preceding work. They produced correct ATWL representations with minimal human intervention. The instructions were then used for the second stage of the library construction. Formalising a workflow in the first stage took about an hour of supervised interaction. In the second stage, it took 15--20 minutes or less. This time includes human supervision and passing the outputs from the extractor to the evaluator and back. The language definition did not need to be changed for any of the 36 workflows added in the second stage.

\subsection{Verification}
\label{sec:library-verification}

We checked the representations of the first 17 workflows and of EventThread, which was added later. The depth of the checks varied. It was highest for the workflows of our own papers and for the workflows that we have used in our research or teaching, for example, Cluster-Calendar, Snapshots-to-Points, and Partition-based Regression. The other 36 workflows were evaluated by the evaluator instance, and we answered the questions of the extractor, but we did not check these workflows at the same level of detail. 

As an additional check, we asked the authors of the library papers to evaluate the summaries and the ATWL representations of their workflows. This evaluation covers only a part of the library. Its method and results are described in Section~\ref{sec:expert_feedback}. To see how much a representation depends on the extraction, we also extracted five workflows again with Claude Sonnet 5 and GPT-5.6-Terra, each reviewed by the other model (Section~\ref{sec:reextraction} and Appendix~\ref{app:reextraction}). The backbone of the workflows was reproduced and the details varied, so the authors of a workflow should check its representation.

The ATWL checker (\url{https://geoanalytics.net/VAworkflows/library/atwl_syntax_checker.html}) applies 45 checks to a specification, which can be pasted, uploaded, or loaded from the URL of a library page with an embedded ATWL source. The page lists all checks and the sections of the language definition they refer to. There are 14 types of errors, which withhold the statistics and diagrams of the workflow until they are fixed, and 31 warnings, which never block them. The checks fall into the following groups.
\begin{itemize}
  \item \emph{Syntax and structure} (errors): malformed or incomplete declarations, missing colons, unclosed or mismatched loops, duplicate identifiers, a workflow without transforms, and inputs or outputs of transforms that refer to undeclared artifacts.
  \item \emph{Vocabulary and declarations} (warnings): unknown intents, actors, and artifact categories, a missing workflow header, and unused artifacts.
  \item \emph{Intent and output types} (warnings): none of the outputs of a transform has the main artifact type listed for its intent in Table~\ref{app:tab:intents}, or a transform outputs a type that the table does not list for the intent.
  \item \emph{Declarations of artifacts} (warnings): required fields of each artifact type are missing, and \kw{feature} and \kw{arrangement} declarations do not refer to \kw{entities} artifacts.
  \item \emph{Graph structure} (warnings): a cycle in the dependency graph other than through \kw{assign} in a loop, a transform without inputs or outputs, and disconnected parts of the workflow.
  \item \emph{Loops and assignments} (warnings): a loop without a termination condition or body, an \kw{exit loop} that does not refer to an enclosing loop, and updates of artifacts that are undeclared, not used by any transform, not produced inside the loop, or lost.
  \item \emph{Origin of artifacts} (warnings): the \kw{origin} marker does not agree with the production of the artifact.
\end{itemize}
10 of the checks offer an automatic correction where the correction is obvious, for example for misspelled keywords. For problems without an obvious correction, the checker issues a warning. The checker verifies the form of a specification. Whether a specification represents the source workflow faithfully has to be judged by the evaluator and the human experts. The library workflows produce neither errors nor warnings. The extraction had left warnings for 30 of them, which mostly concerned the typing of outputs and the use of updated artifacts. We corrected these representations until the checker reported none. The authors of the library papers evaluated their workflows before these corrections (Section~\ref{sec:expert_feedback}).

\section{Details of the Re-Extraction Check}
\label{app:reextraction}

This appendix gives the details behind Section~\ref{sec:reextraction}. The Web supplement (\url{https://geoanalytics.net/VAworkflows/reextraction_R1/index.html}) contains the extraction and review instructions, the final ATWL representations of the re-extractions, the review reports, and the script and tables of the analysis.

\subsection{Procedure}

We chose five library workflows with different data and analytical approaches: UTOPIAN, DPVis, EventThread, NodeTrix, and SOMFlow. Each was extracted by Claude Sonnet 5 and by GPT-5.6-Terra, with the extraction instructions and the example workflows of the library construction. As in the library construction, each LLM instance processed the five papers one after another in one session. The syntax checker was run on every result, and its errors were passed back to the extractor; this took between 0 and 2 corrections per workflow, most often none. The other model then reviewed the representation against the paper with the review instructions of the library construction, and the extractor revised its version once. The revisions contained no syntax errors. We compare these revised versions with each other and with the library versions, which were extracted earlier with the procedure described in Section~\ref{sec:library}; Claude Opus 4.6 took both the extractor and the evaluator role for these five.

The features and transitions were computed from the ATWL text as in Section~\ref{sec:analysis}. A re-extraction is compared with the library version of the same paper, with the other re-extraction, and, for the rank, with all 54 library workflows.

\subsection{Agreement per Paper}

Table~\ref{tab:reextraction-papers} gives the numbers per paper. The library version of a paper is never the most similar library workflow of a re-extraction. The rank is best-case; the Sonnet version of UTOPIAN shares no present feature with the library version, so its rank is between 41 and 54.

\begin{table*}[t]
\setlength{\belowcaptionskip}{0pt}
\caption{Five workflows extracted three times: library version (L), Sonnet (S), and Terra (T). Rank: position of the library version of the paper among the 54 library workflows ordered by the Jaccard similarity to the re-extraction (1 is the most similar).}
\label{tab:reextraction-papers}
\footnotesize
\renewcommand{\arraystretch}{1.1}
\begin{tabular}{@{}l rrr rrr rrr rr@{}}
\toprule
 & \multicolumn{3}{c}{Transforms} & \multicolumn{3}{c}{Features present} & \multicolumn{3}{c}{Features agreeing (of 25)} & \multicolumn{2}{c}{Rank} \\
\cmidrule(lr){2-4}\cmidrule(lr){5-7}\cmidrule(lr){8-10}\cmidrule(l){11-12}
Paper & L & S & T & L & S & T & L--S & L--T & S--T & S & T \\
\midrule
UTOPIAN & 13 & 10 & 18 & 4 & 4 & 9 & 17 & 18 & 20 & 41 & 8 \\
DPVis & 24 & 21 & 18 & 10 & 10 & 10 & 15 & 15 & 17 & 17 & 11 \\
EventThread & 9 & 13 & 19 & 5 & 7 & 11 & 21 & 15 & 19 & 4 & 33 \\
NodeTrix & 11 & 13 & 14 & 5 & 7 & 9 & 19 & 17 & 17 & 11 & 33 \\
SOMFlow & 27 & 14 & 21 & 10 & 10 & 12 & 13 & 15 & 17 & 30 & 8 \\
\midrule
Mean (rank: median) & 16.8 & 14.2 & 18.0 & 6.8 & 7.6 & 10.2 & 17.0 & 16.0 & 18.0 & 17 & 11 \\
\bottomrule
\end{tabular}
\end{table*}

Terra produced the most transforms in three of the five papers. Without a comparison of repeated runs of the same model, we cannot separate the effect of the model from the variation between runs.

\subsection{Sources of Variation}

The extraction instructions leave several decisions to the extractor. The explanations that Sonnet gave for its representations and the reviews show five kinds of decisions.

\textbf{Scope.} For a tool paper, the instructions ask for an assumed composite workflow built from the usage scenarios, and they leave the choice of the scenarios and of the linear order to the extractor. Nine of the ten re-extractions state in their description that they are assumed composite workflows; none of the 54 library descriptions does.

\textbf{Abstraction.} Sonnet folded the model selection of DPVis into the manner of one \kw{build-model} transform as an implementation detail. The Terra review of this version flagged this as an issue.

\textbf{Decomposition.} Because ATWL allows one intent per transform, Sonnet split the self-organising map of SOMFlow into three transforms (model, cells, and layout).

\textbf{Typing.} Sonnet typed the researchers of NodeTrix as separate entities with a relational embedment, and the cells of SOMFlow as groups that are related to each other. The Sonnet review of the Terra version of NodeTrix flagged the embedment of the networked entities.

\textbf{Control structure and bundling.} The choice between the two loop patterns, and whether interactions that the paper separates are bundled into one specification, differ between versions. Sonnet used one specification for both the adapting and the partitioning interactions of SOMFlow.

\subsection{Reviews}

The ten review reports contain 47 critical issues and recommended improvements. We assigned each to a phase of the review instructions; the assignment is ours and approximate. Structural validation, such as missing inputs of visualisations, provenance of the final knowledge, and declarations, accounts for 23 items. Paper alignment accounts for 18, mostly content of the paper that is not represented and statements that the paper does not support. Abstraction level accounts for five, and semantic correctness for one.

The reviewers differed in their strictness. The Terra reviews of the five Sonnet versions all recommended a revision and contained 15 critical issues. The Sonnet reviews of the Terra versions recommended minor revisions for four of the five and contained two critical issues, both for DPVis. The strictness is confounded with the quality of the extraction that was reviewed, and the models reviewed different versions. One issue, the provenance of the final knowledge artifact, was flagged in four of the five Sonnet versions, in three of them after an earlier review in the same session had raised it.

\subsection{Library Statistics under Substitution}

We replaced the five library versions by the re-extractions of one model and recomputed the feature frequencies of the library. The count of any feature changes by at most four of the 54 workflows, and the mean number of features per workflow changes from 7.5 to 7.6 (Sonnet) and 7.8 (Terra). The frequencies correlate with the original ones with $r=0.98$ for both models. The Jaccard similarity of the sets of intent transitions, after merging consecutive transforms of the same intent, is 0.44 between a re-extraction and the library version and 0.59 between the two re-extractions.

\section{Frequent Subgraph Mining, Path Enumeration, Features and Tests}
\label{app:mining}

\textbf{Setting.}
Each workflow $w$ is a directed graph $G_w=(V_w,E_w)$. Nodes are transforms with a label $\lambda(v)$ (the intent, and in the first run also the actor). An edge $e=(u,v,\ell)$ carries a label $\ell$ (the category of the artifact passed and, in the first run, whether the edge is feedback); parallel edges with different labels are distinct edges.
A \emph{pattern} is a connected, labelled, directed graph. It \emph{occurs} in $G_w$ if it is isomorphic, respecting directions and labels, to the subgraph formed by a subset of edges of $G_w$ (the subgraph need not be induced). The \emph{support} of a pattern is the number of workflows in which it occurs (graph-transaction setting); repeated occurrences within one workflow are not counted.

\textbf{Mining.}
Algorithm~\ref{alg:mining} is a level-wise (Apriori-style \cite{AgrawalSrikant1994}) pattern-growth miner, similar in spirit to FSG~\cite{KuramochiKarypis2001}. For every workflow it keeps the embeddings (edge sets) of the frequent patterns of size $k-1$ and extends each by one adjacent edge, which either adds a node or closes a cycle. The resulting subgraphs are grouped by a canonical form, and the patterns whose support reaches the threshold $\sigma$ are kept for the next level. Since support cannot increase when a pattern grows, and every connected graph with $k$ edges has an edge whose removal leaves a connected graph, the procedure finds every frequent pattern of up to $K$ edges. Because patterns have at most five nodes, the canonical form is computed by trying all orderings that permute only nodes with equal labels. The cost grows exponentially with $K$; for graphs of up to about 58 nodes and $K\le4$ the runs took minutes. The implementation (the package \texttt{atwl\_tools}) is part of the Web supplement. It is our own simple implementation and was not benchmarked against other miners, such as gSpan~\cite{YanHan2002}; the supports of patterns of one and two edges were verified by recomputing them independently for every workflow.
The first run used $\sigma=5$ and $K=4$ with actor and feedback labels and found 3\,655 patterns. The later analyses used intent-level labels (the artifact category is kept only on edges leaving generate-knowledge or define-unit), $\sigma=3$, and $K=2$.

\textbf{Paths and closed cycles.}
Paths of three and four edges and closed cycles were not obtained by the miner but enumerated exhaustively (Algorithm~\ref{alg:paths}) in the graphs with intent-labelled nodes, which keeps their definition exact. A path visits distinct nodes; a closed cycle is a path of distinct nodes whose last node has an edge back to the first. Since the dependency graph of the transforms of an ATWL workflow is acyclic except for the edges created by \kw{assign} statements, every closed cycle of transforms contains such an edge; merging consecutive transforms of the same intent can also create a cycle of steps without one, which happens in two of the 54 workflows. Each cycle is recorded once, from its smallest node, as its label sequence rotated to the lexicographically smallest rotation.

\begin{figure*}[!t]
\centering
\footnotesize
\algrenewcommand\algorithmicindent{0.9em}
\begin{minipage}[t]{0.485\textwidth}
\hrule height 0.8pt\vspace{2pt}
\refstepcounter{algorithm}\label{alg:mining}
\textbf{Algorithm~\thealgorithm}\enspace Level-wise frequent subgraph mining
\vspace{2pt}\hrule height 0.4pt\vspace{1pt}
\begin{algorithmic}[1]
\Require graphs $\{G_w\}$, minimum support $\sigma$, maximum size $K$ (edges)
\Ensure frequent patterns $\mathcal{F}_k$ with supports, $k=1,\dots,K$
\Function{Canon}{$S$}\Comment{canonical form of edge set $S$}
  \State $\pi$: orderings of the nodes of $S$ by label that permute only equal labels
  \State \Return node labels, $\min_\pi\,\mathrm{sort}\,[(\pi(u),\pi(v),\ell)]_{(u,v,\ell)\in S}$
\EndFunction
\ForAll{$w$ and $e\in E_w$}
  \State add $w$ to $\mathit{sup}[\textsc{Canon}(\{e\})]$
\EndFor
\State $\mathcal{F}_1\gets\{p : |\mathit{sup}[p]|\ge\sigma\}$
\State $\mathit{Emb}^1_w\gets\{\{e\} : \textsc{Canon}(\{e\})\in\mathcal{F}_1\}$ for all $w$
\For{$k=2,\dots,K$}
  \State $\mathit{sup}\gets\emptyset$; $\mathit{New}_w\gets\emptyset$ for all $w$
  \ForAll{$w$ and $S\in\mathit{Emb}^{k-1}_w$}
    \ForAll{$e\in E_w\setminus S$ incident to a node of $S$}
      \State $T\gets S\cup\{e\}$
      \If{$T\notin\mathit{New}_w$}
        \State add $T$ to $\mathit{New}_w$; add $w$ to $\mathit{sup}[\textsc{Canon}(T)]$
      \EndIf
    \EndFor
  \EndFor
  \State $\mathcal{F}_k\gets\{p : |\mathit{sup}[p]|\ge\sigma\}$
  \State $\mathit{Emb}^k_w\gets\{T\in\mathit{New}_w : \textsc{Canon}(T)\in\mathcal{F}_k\}$ for all $w$
\EndFor
\State \Return $\mathcal{F}_1,\dots,\mathcal{F}_K$ with their supports
\end{algorithmic}
\vspace{1pt}\hrule height 0.8pt
\end{minipage}\hfill
\begin{minipage}[t]{0.485\textwidth}
\hrule height 0.8pt\vspace{2pt}
\refstepcounter{algorithm}\label{alg:paths}
\textbf{Algorithm~\thealgorithm}\enspace Enumeration of simple paths and closed cycles
\vspace{2pt}\hrule height 0.4pt\vspace{1pt}
\begin{algorithmic}[1]
\Require graphs $\{G_w\}$ with intent-labelled nodes; path length $L=4$ edges, cycle length $C=6$ nodes
\Ensure supports of label sequences of paths and of cycles
\ForAll{$w$ and $s\in V_w$}
  \State \Call{Path}{$w,[s]$}; \Call{Cycle}{$w,s,[s]$}
\EndFor
\Procedure{Path}{$w,P$}\Comment{$P$: distinct nodes, $\lambda(P)$: their labels}
  \If{$|P|\ge2$}
    \State add $w$ to $\mathit{sup}[\lambda(P)]$
  \EndIf
  \If{$|P|-1<L$}
    \ForAll{$v\notin P$ with an edge $(\mathrm{last}(P),v)$}
      \State \Call{Path}{$w,P+[v]$}
    \EndFor
  \EndIf
\EndProcedure
\Procedure{Cycle}{$w,s,P$}
  \ForAll{$v$ with an edge $(\mathrm{last}(P),v)$}
    \If{$v=s$ and $|P|\ge3$}
      \State record $\lambda(P)$ in its smallest rotation for $w$
    \ElsIf{$v>s$, $v\notin P$ and $|P|<C$}
      \State \Call{Cycle}{$w,s,P+[v]$}
    \EndIf
  \EndFor
\EndProcedure
\end{algorithmic}
\vspace{1pt}\hrule height 0.8pt
\end{minipage}
\end{figure*}

\textbf{Selection.}
Candidate patterns were reduced by explicit rules (Section~\ref{sec:analysis}). For a path or chain $p$ let $p'$ and $p''$ be $p$ without its last and without its first edge, $s(\cdot)$ the support, and $c(p)$ the number of workflows containing both $p'$ and $p''$. A pattern is kept if (i) $3\le s(p)\le 48$; (ii) $s(p)<\min\{s(p'),s(p'')\}$, so that it is not implied by a part; (iii) $c(p)-s(p)\ge3$, so that it differs from the mere co-occurrence of its parts in at least three workflows; and (iv) $s(p)/c(p)\ge0.5$, so that its parts are coupled. For chains of two edges, the two edges are the parts.

\textbf{Features.}
The 25 features of Section~\ref{sec:analysis} are computed by program from the ATWL text; Table~\ref{tab:features} gives their rules. No feature uses free text. In the ATWL Workflow Lens, every feature value is linked to the transforms or assignments that realise it; a test over all 54 workflows confirms that a feature is present exactly when such evidence exists.

\textbf{Statistical tests.}
The tests behind the last paragraph of Section~\ref{sec:analysis} compare each of the 25 features with each of nine binary attributes of the workflows (five kinds of given data, two kinds of units, integration of several datasets, and co-author involvement), 225 comparisons in total. The attributes of data and units are read from the ATWL declarations, co-author involvement from the author lists of the library pages. We used Fisher's exact test (two-sided) and controlled the false discovery rate over the 225 comparisons with the Benjamini--Hochberg procedure. To ask whether whole feature profiles differ between workflows with and without an attribute, we applied a permutation test (PERMANOVA, 9\,999 permutations) to the Jaccard distances between feature sets; for two groups, the share of explained variance is $F/(F+N-2)$ with $N=54$. With all 25 features, sequence data explain 5.1\% and integration of several datasets 8.3\% of the variance; without the feature ``two or more given datasets'', which is defined from the given data, the values are 5.5\% and 5.1\%. Co-author involvement explains 2.8\% ($p=0.12$). Tests within three strata of workflow size (number of transforms; Cochran--Mantel--Haenszel test with the continuity correction of R) found no further associations: 6 comparisons with $p<0.05$ (20 without the adjustment) and one after correction.

\textbf{Reproducibility.}
The numbers of Section~\ref{sec:analysis} refer to the library of 13 October 2026. The features and all analyses can be recomputed with the scripts of the Web supplement, which read the ATWL files of the library; the data file of the ATWL Workflow Lens for this version of the library is archived there.

\begin{table*}[t]
\setlength{\belowcaptionskip}{0pt}
\caption{The 25 features. An edge links the transform that produces an artifact to a transform that uses it, directly or through an \texttt{assign} in a loop; a machine-performed abstraction counts as define-unit. $n$: number of the 54 workflows that have the feature.}
\label{tab:features}
\footnotesize
\renewcommand{\arraystretch}{1.12}
\newcolumntype{Y}{>{\raggedright\arraybackslash}X}
\begin{tabularx}{\textwidth}{@{}l >{\raggedright\arraybackslash}p{0.31\textwidth} Y r@{}}
\toprule
ID & Feature & Rule & $n$ \\
\midrule
E01 & decision re-specifies units & edge generate-knowledge $\to$ define-unit & 30 \\
E02 & decision re-specifies features & edge generate-knowledge $\to$ characterise & 24 \\
E03 & decision straight from a view & edge visualise $\to$ generate-knowledge, and the consuming transform has no pattern or model among its inputs & 19 \\
E04 & units from view, interpretation or arrangement & edge from visualise, abstract or contextualise $\to$ define-unit & 23 \\
E05 & model output used as features & edge build-model $\to$ characterise & 12 \\
E06 & decision re-specifies a view & edge generate-knowledge $\to$ visualise & 15 \\
E07 & decision re-specifies the arrangement & edge generate-knowledge $\to$ contextualise & 11 \\
E08 & decision re-specifies the model & edge generate-knowledge $\to$ build-model & 10 \\
E09 & decision guides abstraction & edge generate-knowledge $\to$ abstract & 3 \\
E10 & judgement redefines the working set & edge assess $\to$ define-unit & 3 \\
E11 & model output defines units & edge build-model $\to$ define-unit & 4 \\
\midrule
I01 & units redefined in a loop & an \texttt{assign} inside a loop updates an entities artifact & 19 \\
I02 & features redefined in a loop & an \texttt{assign} inside a loop updates a feature artifact & 23 \\
I03 & arrangement redefined in a loop & an \texttt{assign} inside a loop updates an arrangement artifact & 16 \\
I04 & model refined in a loop & an \texttt{assign} inside a loop updates a model artifact & 8 \\
I05 & specification revised in a loop & an \texttt{assign} inside a loop updates a specification artifact & 35 \\
I06 & patterns revised in a loop & an \texttt{assign} inside a loop updates a pattern artifact & 23 \\
I07 & knowledge accumulated in a loop & an \texttt{assign} inside a loop updates a knowledge artifact & 7 \\
\midrule
T01 & two or more given datasets & two or more given entities artifacts & 18 \\
T02 & units that are groups or clusters & a derived entities artifact whose internal structure is a group & 23 \\
T03 & several views used together & a human or hybrid abstract or assess transform with two or more visualisation inputs & 27 \\
T04 & two or more sequential loops & two or more top-level (not nested) loops & 14 \\
\midrule
H01 & hybrid transform & at least one transform with actor hybrid & 19 \\
H02 & a priori configuration & a human or hybrid generate-knowledge outside any loop produces a specification and has no assess, visualise or abstract predecessor & 10 \\
H03 & initial knowledge & a given knowledge artifact, or a human or hybrid generate-knowledge that produces knowledge and has no such predecessor & 8 \\
\bottomrule
\end{tabularx}
\end{table*}

\section{Details of the Workflow Recommendation Experiments}
\label{app:recommendation}

This appendix gives the details behind Section~\ref{sec:recommendation}. The Web supplement (\url{https://geoanalytics.net/VAworkflows/experiments_R1/index.html}) contains the prompts, all designs in ATWL and prose, the chat logs, the coder instructions and outputs, the data of all tables, and the scripts that recompute them.

\subsection{Conditions and Prompts}

Each of the two tasks and two models gave 8 designs (C1--C7 and C6++), 32 in total, and 4 responses to the C6++ prompt. The designs of C2 and C3 contain a prose explanation and an ATWL representation; we treat the ATWL as the design. The prompts differed in the following points. C2 and C3 asked for the recommendation in natural language and in ATWL; the example workflow was marked as showing how the language is used, not the structure that a task requires. C3 and C4 asked the model to cite the library workflows it used by name, and said that the library need not be used. The C5 prompt told the model that the structure remains fixed, asked for method-level detail from the source papers in a table, and asked to list possible structural refinements separately. The C6 prompt asked for the ATWL representation that most faithfully captures the design, not to fabricate structure that the design does not support, and to list the structural elements that were implicit. The C7 prompt asked for a fresh design in ATWL and said that fidelity to the earlier design is not a goal. The C6++ prompt asked whether ATWL thinking revealed gaps, and named three kinds as examples (loops without specifications, ambiguous actor roles, implicit criteria). The C6++ prompt asked to implement the modifications only where they are real improvements and allowed restructuring. The C6++ designs keep 51--100\% of the artifact and transform identifiers of the C6 design of the same block, whereas the C7 designs share only 9--25 identifiers with it (of 59--92 in C7).

\subsection{Library Use}

Library workflows named by a design were found by matching the titles and tool names of the 54 workflows in the text of the design (for C2 and C3, prose and ATWL). Table~\ref{tab:rec-library} gives the counts. The citation lists are the lists of papers that the models returned when asked for the papers from which they had taken approaches (C3, C4). The methods are those listed in the comparison tables of the supplement, where for each paper and each method the cells of the four designs show what was taken. The method rows were compiled by us, so only the comparison between C4 and C5, not the absolute numbers, should be read.

\begin{table}[H]
\setlength{\belowcaptionskip}{0pt}
\caption{Library use per block (task and model). Named: library workflows named in the design. Cited: papers in the citation list. Methods: methods of the comparison tables that the design includes. C1 and C2 name no library workflow in any block.}
\label{tab:rec-library}
\footnotesize
\renewcommand{\arraystretch}{1.1}
\begin{tabular}{@{}l rr rr rr@{}}
\toprule
 & \multicolumn{2}{c}{Named} & \multicolumn{2}{c}{Cited} & \multicolumn{2}{c}{Methods} \\
\cmidrule(lr){2-3}\cmidrule(lr){4-5}\cmidrule(lr){6-7}
Block & C3 & C4 & C3 & C4 & C4 & C5 \\
\midrule
VAST, Sonnet & 5 & 11 & 6 & 6 & 7 & 15 \\
VAST, Terra & 4 & 7 & 4 & 7 & 7 & 10 \\
VIS, Sonnet & 9 & 12 & 10 & 12 & 14 & 27 \\
VIS, Terra & 4 & 5 & 5 & 8 & 10 & 11 \\
\bottomrule
\end{tabular}
\end{table}

\subsection{Coding of the Prose Designs}

The eight prose designs of C1 and C4 were cleaned (final answer only, closing offers removed), divided into units at the author's own steps, and given neutral identifiers. Two LLM coders, Claude Sonnet 5 and GPT-5.6-Terra, coded every design, each design in its own fresh, independent session, so that each instance saw one design only. A session received the written protocol (version 1.5) and the design, and no information about condition, model, or task; the sessions were run in a different random order for each coder. For each unit the output states whether a human and a machine action is stated, the human control acts (set, select, judge, decide, annotate) with the evidence and criterion of each judgement, and whether an output is named. For the whole design it lists loops (what is redefined, trigger, exit), named parameters, branches, and provenance statements. Every code carries a verbatim quote. The protocol was developed on two designs (D01 and D02), which were then coded again under the final version; the other six were coded once.

All 16 outputs parse and cover every unit. Of 925 quotes, 910 are exact, 15 splice a sentence to a non-adjacent list item (each part verbatim), and none is missing. Over the 65 units the coders agree on a stated human action in 92\%, a stated machine action in 95\%, a named output in 89\%, and the presence of a control act in 86\%. Item-level agreement is lower (F1 for control acts 0.71, for parameters 0.86, for branches 0.71) because the coders split lists and conditionals differently. On whole-design measures the coders agree well on named parameters ($r=0.98$), the share of units naming an output (0.95), a machine action (0.91), a control act (0.86), and loops (0.83) and branches (0.82), and poorly on the number of judgement acts (rank correlation 0.29), which we do not use. Sonnet reports more units with a control act and more named outputs than Terra. Table~\ref{tab:rec-coding} gives the comparison of C4 with C1 for each coder, and Table~\ref{tab:rec-mapping} the mapping to ATWL elements.

\begin{table}[H]
\setlength{\belowcaptionskip}{0pt}
\caption{Library in prose form: C4 compared with C1 in the coding (means of four blocks; Sonnet / Terra coder). $\Delta$: mean difference C4 minus C1; blocks: number of blocks with higher / lower value in C4.}
\label{tab:rec-coding}
\footnotesize
\renewcommand{\arraystretch}{1.1}
\begin{tabularx}{\columnwidth}{@{}>{\raggedright\arraybackslash}X c c c c@{}}
\toprule
 & C1 & C4 & $\Delta$ & Blocks \\
\midrule
Loops stated & 0.75 / 0.75 & 1.50 / 1.25 & +0.75 / +0.50 & 3/0 ; 2/0 \\
Annotation and logging acts & 1.50 / 0.75 & 3.00 / 2.50 & +1.50 / +1.75 & 3/0 ; 4/0 \\
Provenance statements & 0 / 0 & 0.50 / 0.75 & +0.50 / +0.75 & 2/0 ; 3/0 \\
Units naming an output & 0.37 / 0.35 & 0.71 / 0.51 & +0.34 / +0.16 & 3/1 ; 2/1 \\
Branches stated & 3.75 / 5.00 & 6.50 / 6.00 & +2.75 / +1.00 & 2/1 ; 3/1 \\
Named parameters & 5.00 / 4.50 & 4.75 / 5.00 & $-$0.25 / +0.50 & 2/2 ; 2/2 \\
Units with a control act & 0.83 / 0.73 & 0.84 / 0.66 & +0.01 / $-$0.06 & 2/1 ; 2/2 \\
Units with a human action & 0.85 / 0.78 & 0.84 / 0.76 & $-$0.01 / $-$0.02 & 2/1 ; 2/1 \\
Judgements with criterion & 0.44 / 0.48 & 0.40 / 0.53 & $-$0.04 / +0.05 & 2/2 ; 2/2 \\
Loops stating an exit & 0 / 0 & 0 / 0 & 0 / 0 & -- \\
\bottomrule
\end{tabularx}
\end{table}

\begin{table}[H]
\setlength{\belowcaptionskip}{0pt}
\caption{Mapping between the coded elements and ATWL.}
\label{tab:rec-mapping}
\footnotesize
\renewcommand{\arraystretch}{1.1}
\begin{tabularx}{\columnwidth}{@{}>{\raggedright\arraybackslash}p{0.25\columnwidth} >{\raggedright\arraybackslash}X >{\raggedright\arraybackslash}X@{}}
\toprule
 & Prose (coded) & ATWL (computed) \\
\midrule
Loops & loops stated & loop constructs \\
Judgement steps & judge and decide acts & assess, abstract, or generate-knowledge transforms by human or hybrid actors \\
Tunable parameters & parameters stated as tunable & specification artifacts revised by an \kw{assign} in a loop \\
Conditional paths & conditional branches & \kw{if} blocks outside loops \\
Actor, output, exit & coded per step or loop & declared by syntax \\
\bottomrule
\end{tabularx}
\end{table}

\subsection{ATWL Designs}

Table~\ref{tab:rec-atwl} gives the 25 features and descriptors of the 20 ATWL designs by condition. The designs have on average 10.4 of the 25 features, the library 7.5, and ten features differ at $q<0.05$ (Fisher exact test, Benjamini--Hochberg): E05, E10, I05, T01, T02, T04, H01, and H03 are more frequent in the designs, I03 and I06 less. The median Jaccard similarity of a design to its most similar library workflow is 0.55, against 0.60 between library workflows. Two encodings of the same workflow agree only moderately in their features (Section~\ref{sec:reextraction}), so we treat these similarities as descriptive and not as evidence of a difference between conditions. A randomisation of the condition labels within blocks gives $p=0.11$ for the number of features, and no single feature differs between conditions after correction. A PERMANOVA on the Jaccard distance of the feature profiles attributes 15\% of the variance to the model ($p=0.0015$), 13\% to the task ($p=0.006$), and 15\% to the condition ($p=0.94$). The designs made directly with the ATWL materials (C2, C3, $n=8$) have fewer transforms (21.5) than the designs derived from a prose design (C6, C6++, C7, $n=12$; 36.9), fewer features (8.6 against 11.6) and no checker errors (0 of 8 against 7 of 12); the direction is the same in all four blocks, and the contrast was defined after seeing the data. The errors are dangling references (98) and duplicate artifacts (2); six of the six Terra designs and one of the six Sonnet designs of the derived group have them.

\begin{table}[H]
\setlength{\belowcaptionskip}{0pt}
\caption{ATWL designs by condition (means of four blocks). Similarity: Jaccard similarity of the 25 features to the most similar library workflow.}
\label{tab:rec-atwl}
\footnotesize
\renewcommand{\arraystretch}{1.1}
\begin{tabular}{@{}l rrrrr@{}}
\toprule
 & C2 & C3 & C6 & C6++ & C7 \\
\midrule
Features present (of 25) & 7.50 & 9.75 & 10.50 & 12.75 & 11.50 \\
Transforms & 18.8 & 24.2 & 35.2 & 43.8 & 31.8 \\
Loops & 1.75 & 2.75 & 2.25 & 4.25 & 2.75 \\
Hybrid share of transforms & 0.13 & 0.05 & 0.09 & 0.07 & 0.17 \\
Designs with checker errors (of 4) & 0 & 0 & 2 & 2 & 3 \\
Similarity to the library & 0.48 & 0.62 & 0.53 & 0.58 & 0.51 \\
\bottomrule
\end{tabular}
\end{table}

\subsection{From the Prose Design to ATWL}

For each block, the C4 design (coded) was followed by its translation (C6), its revision (C6++), and its redesign (C7). On average over the four blocks, the C4 designs have 1.4 loops, the C6 designs 2.25, C6++ 4.25, and C7 2.75. Human judgement and decision transforms are 11.0, 15.3, and 10.5 in C6, C6++, and C7; they are not compared with the prose designs, because the coders disagree on judgement acts. Conditional paths are 4.1 in C4 and 0, 0.25, and 0.25 in C6, C6++, and C7 (\kw{if} blocks outside loops). Of the 17 loop records of the prose designs (both coders), none states an exit, and 3 of 6 in C1 and 4 of 11 in C4 state a trigger. Of the 55 loops in the ATWL designs all have an \kw{until} clause, whose content was not evaluated. When asked, the models explained what they had added in the ATWL: faithful translations of stated iteration, formalisations that the syntax requires (control flow, specification artifacts, actors), and genuine additions. Conditionals are rare outside loops in all ATWL designs: the 20 designs have 53 \kw{if} blocks inside loops and 2 outside, none in C2 or C3, and the two \kw{if} blocks of the example workflow are loop exit gates. Most branches coded in the prose designs are conditional paths (4.1 of 6.25 per C4 design; the others are alternative hypotheses and fallbacks). By our reading of the quotes, about 4 of the 33 conditional-path records of the C4 designs (both coders) describe repeating a step until a result is adequate, which ATWL expresses as a loop exit; the others are conditions on data availability, rules that map an observed pattern to a hypothesis, and triggers of a step. We did not check whether the ATWL designs capture such rules in another form, for example in specification artifacts. One model (VIS, Sonnet) stated in its C6 explanation that it did not represent a fallback to another method, because the prose had offered it only as an option. In the C7 explanations, one model (VAST, Sonnet) states that naming a specification artifact for every loop forced it to separate steps with a parameter that can be calibrated from steps of pure interpretation, and that it left loops out of steps that had no such parameter.

\begin{figure*}[t]
\centering
\resizebox{!}{0.98\textheight}{%
\begin{tikzpicture}[
    scale=1.0, transform shape,
    every node/.style={align=center},
    decision/.style={diamond, draw=black, thick, fill=yellow!15,
        aspect=2, minimum width=1.8cm, text centered,
        font=\small, inner sep=1pt}
]

\node[artifact] at (0, 0)       (Dhour)  {D\_hour\\[-2pt]{\scriptsize Time series}};
\node[artifact] at (5.5, 0)     (Dcal)   {D\_calendar\\[-2pt]{\scriptsize Calendar}};

\node[transform] at (0, -1.2)   (Tpart)  {T\_partition\\[-2pt]{\scriptsize define-unit}};
\node[artifact]  at (0, -2.4)   (Dday)   {D\_day\\[-2pt]{\scriptsize Daily episodes}};

\node[transform] at (-2.5, -3.6)(Tprof)  {T\_profile\\[-2pt]{\scriptsize characterise}};
\node[transform] at (3, -3.6)   (Tarr)   {T\_arrange\\[-2pt]{\scriptsize contextualise}};

\node[artifact] at (-2.5, -4.8) (Fprof)  {F\_day\_profile\\[-2pt]{\scriptsize Daily profiles}};
\node[artifact] at (5.5, -4.8)  (Acal)   {A\_calendar\\[-2pt]{\scriptsize Calendar arr.}};
\node[artifact] at (-6.5, -4.8) (Sclust) {S\_clustering\\[-2pt]{\scriptsize Parameters}};

\draw[arrow] (Dhour) -- (Tpart);
\draw[arrow] (Tpart) -- (Dday);
\draw[arrow] (Dday)  -- (Tprof);
\draw[arrow] (Dday)  -- (Tarr);
\draw[arrow] (Dcal)  -- (Tarr);
\draw[arrow] (Tprof) -- (Fprof);
\draw[arrow] (Tarr)  -- (Acal);

\node[transform] at (0, -7.5)    (Tclust) {T\_cluster\\[-2pt]{\scriptsize define-unit}};
\node[artifact]  at (-2.5, -8.7) (Dclust) {D\_cluster\\[-2pt]{\scriptsize Day groups}};
\node[artifact]  at (2.5, -8.7)  (Flabel) {F\_cluster\_label\\[-2pt]{\scriptsize Membership}};

\node[transform] at (-2.5, -9.9) (Tagg)    {T\_aggregate\\[-2pt]{\scriptsize characterise}};
\node[transform] at (5.5, -9.9)  (Tcalvis) {T\_calendar\_vis\\[-2pt]{\scriptsize visualise}};
\node[artifact]  at (-2.5,-11.1) (Fcprof)  {F\_cluster\_profile\\[-2pt]{\scriptsize Cluster profiles}};
\node[artifact]  at (5.5, -11.1) (Vcal)    {V\_calendar\\[-2pt]{\scriptsize Calendar view}};

\node[transform] at (-2.5,-12.3) (Tprofvis){T\_profile\_vis\\[-2pt]{\scriptsize visualise}};
\node[artifact]  at (-2.5,-13.5) (Vprof)   {V\_profiles\\[-2pt]{\scriptsize Line graphs}};

\node[human]    at (1.5, -14.9) (Tinterp) {T\_interpret\\[-2pt]{\scriptsize abstract}};
\node[artifact] at (1.5, -16.0) (Ppat)    {P\_patterns\\[-2pt]{\scriptsize Interpreted patterns}};
\node[human]    at (1.5, -17.1) (Tassess) {T\_assess\\[-2pt]{\scriptsize assess}};
\node[artifact] at (1.5, -18.2) (Kassess) {cluster\_assessment\\[-2pt]{\scriptsize Quality judgment}};

\node[decision] at (1.5, -19.4)  (dec)     {Refine?};
\node[human]    at (-6.5, -19.4) (Tadjust) {T\_adjust\\[-2pt]{\scriptsize refine params}};

\draw[arrow] (Fprof)  -- (Tclust);
\draw[arrow] (Sclust) -- (Tclust);
\draw[arrow] (Acal)   -- (Tcalvis);

\draw[arrow] (Tclust)   -- (Dclust);
\draw[arrow] (Tclust)   -- (Flabel);
\draw[arrow] (Dclust)   -- (Tagg);
\draw[arrow] (Tagg)     -- (Fcprof);
\draw[arrow] (Flabel)   -- (Tcalvis);
\draw[arrow] (Tcalvis)  -- (Vcal);
\draw[arrow] (Fcprof)   -- (Tprofvis);
\draw[arrow] (Tprofvis) -- (Vprof);
\draw[arrow] (Vprof)    -- (Tinterp);
\draw[arrow] (Vcal)     -- (Tinterp);
\draw[arrow] (Tinterp)  -- (Ppat);
\draw[arrow] (Ppat)     -- (Tassess);
\draw[arrow] (Tassess)  -- (Kassess);
\draw[arrow] (Kassess)  -- (dec);

\draw[arrow] (dec)     -- node[above] {\scriptsize yes} (Tadjust);
\draw[arrow] (Tadjust) -- node[left]  {\scriptsize update} (Sclust);

\begin{scope}[on background layer]
    \coordinate (loopNW) at
        ($(Tadjust.west |- Tclust.north) + (-0.5, 0.8)$);
    \coordinate (loopSE) at
        ($(Vcal.east |- Tadjust.south) + (0.5, -0.5)$);
    \node[draw, dashed, inner sep=0pt,
          label={[anchor=north west]north west:%
              \textbf{Loop L1: Refine cluster structure}},
          fit=(loopNW)(loopSE)] (loopbox) {};
\end{scope}

\node[human]    at (1.5, -21.2) (Tsynth) {T\_synthesize\\[-2pt]{\scriptsize generate-knowledge}};
\node[artifact] at (1.5, -22.5) (K1)     {K1\\[-2pt]{\scriptsize Temporal pattern knowledge}};

\draw[arrow] (dec)    -- node[right, pos=0.75] {\scriptsize no} (Tsynth);
\draw[arrow] (Tsynth) -- (K1);

\end{tikzpicture}

}%

\caption{Diagrammatic representation of the Cluster-Calendar workflow}
\label{fig:cluster-calendar}
\end{figure*}

\end{document}